\pdfoutput=1 
\documentclass[12pt]{article}                     

\usepackage{graphicx,natbib}
\usepackage{amsmath}
\usepackage{amssymb}
\usepackage{verbatim}

\usepackage{setspace} 

\newcommand{\bw}{{\mathbf{w}}}

\newcommand{\bz}{{\mathbf{z}}}

\newcommand{\cM}{{\mathcal{M}}}

\newcommand{\R}{{\mathbb{R}}}

\newcommand{\E}  {\mbox{E}}

\begin{document}

\doublespacing

\title{Revisiting Complex Moments
For 2D Shape Representation and Image Normalization}


\author{Jo\~{a}o B. F. P. Crespo \and Pedro M. Q. Aguiar 
}

%

\date{May, 2010}

\sloppy

\maketitle

\begin{abstract}
When comparing 2D shapes, a key issue is their normalization. Translation and
scale are easily taken care of by removing the mean and normalizing the energy.
However, defining and computing the orientation of a 2D shape is not so simple.
In fact, although for elongated shapes the principal axis can be used to define
one of two possible orientations, there is no such tool for general shapes.
As we show in the paper, previous approaches fail to compute the orientation of
even noiseless observations of simple shapes. We address this problem. In the
paper, we show how to uniquely define the orientation of an arbitrary 2D shape,
in terms of what we call its Principal Moments.
We show that a small subset of these moments suffice to represent the underlying
2D~shape and propose a new method
to efficiently compute the shape orientation: Principal Moment Analysis.
Finally, we discuss how this method can further be applied to normalize grey-level images.
Besides the theoretical proof of correctness, we describe experiments
demonstrating robustness to noise and illustrating the method with real images.
\end{abstract}

\section{Introduction}
\label{sec:int}

Representing shape is a challenging task. In fact, unlike local characteristics like color, which can be uniquely determined by a small set of parameters, or texture, which has been successfully captured by using statistical descriptors, the visual information conveyed by a shape, of more global nature and easily perceived by humans, is hard to represent in an appropriate way. This paper deals with two-dimensional~(2D) shape representation.

When the 2D~shapes to describe are simply connected regions, researchers have used contour-based descriptions \citep[{\it e.g.},][]{chauang96,bartolini05}. Naturally, for more general shapes, usually consisting in arbitrary sets of points, or landmarks, these descriptors are not adequate. If the points describing the shape are labeled, {\it i.e.}, if the correspondences between the landmarks of two shapes to compare are known, the problem reduces to the impact of geometrical transformations and disturbances, elegantly addressed through the statistical theory of shape of \cite{kendall99}. However, in many practical scenarios, the shape points are obtained from an automatic process, {\it e.g.}, edge or corner detection, thus come without labels or natural ordering.

Estimating the correspondences between points of two shapes leads to a combinatorial problem, which requires prohibitively time-consuming algorithms, even for shapes described by a moderate number of landmarks. To circumvent this problem, researchers have recently attempted to come up with permutation-invariant representations for sets of points. For example, \cite{jebara03} shows how to factor out unknown labels through the solution of a convex optimization problem over the set of permutation matrices, and \cite{rodrigues08a} proposes a permutation-invariant representation obtained by densely sampling an analytic function.

Naturally, the study of shape representation if often motivated by the challenge of comparing two arbitrary shapes. Indeed, if an efficient and universal way of representing shapes is found, the comparison of two shapes can be brought off through the direct comparison of its representations. Regarding universality, an issue that arises when representing and comparing two shapes is their normalization with respect to geometrical transformations, such as translation, scale and rotation. In fact, a desirable requirement of shape representations is that they should form a \emph{complete set of invariants} with respect to these transformations, {\it i.e.}, two shapes should be equal up to a combination of these transformations {\it if and only if} their representations are equal. Translation and scale are easily taken care of by removing the mean and normalizing the energy of the shape. Rotation is correspondingly factored out by normalizing according to an \emph{orientation} of the shape, an angle, intrinsic to the shape, that varies coherently when the shape is rotated. However, as we discuss in the following paragraphs, defining and computing the orientation of an arbitrary 2D shape is not so simple.

Additionally, as motivated earlier, complete permutation invariance is a key requirement of unlabled shape representations as well. Nevertheless, the comparison of two sets of points that are related by an unknown transformation that includes, simultaneously, a 2D rotation, due to different orientation, and a permutation, due to the absence of labels for the points, results highly complex, due to the fact that estimating the transformation leads to a non-convex problem. Iterative methods have been used to compute, in alternate steps, rotation and permutation: the Iterative Closest Point~(ICP) algorithm of \cite{besl92}, or its probabilistic versions based on Expectation-Maximization~(EM) \citep[{\it e.g.},][]{mcneil06cvpr}. However, these approaches suffer from the usual sensitivity to the initialization, exhibiting uncertain convergence. Other proposed approaches exhibit drawbacks as well: the convex optimization approach of \cite{jebara03} does not deal with rotation and the analytic representation of \cite{rodrigues08a} is not rotation invariant, requiring pairwise alignment.

\begin{figure}[hbt]
\centerline{\includegraphics[width=4.5cm]{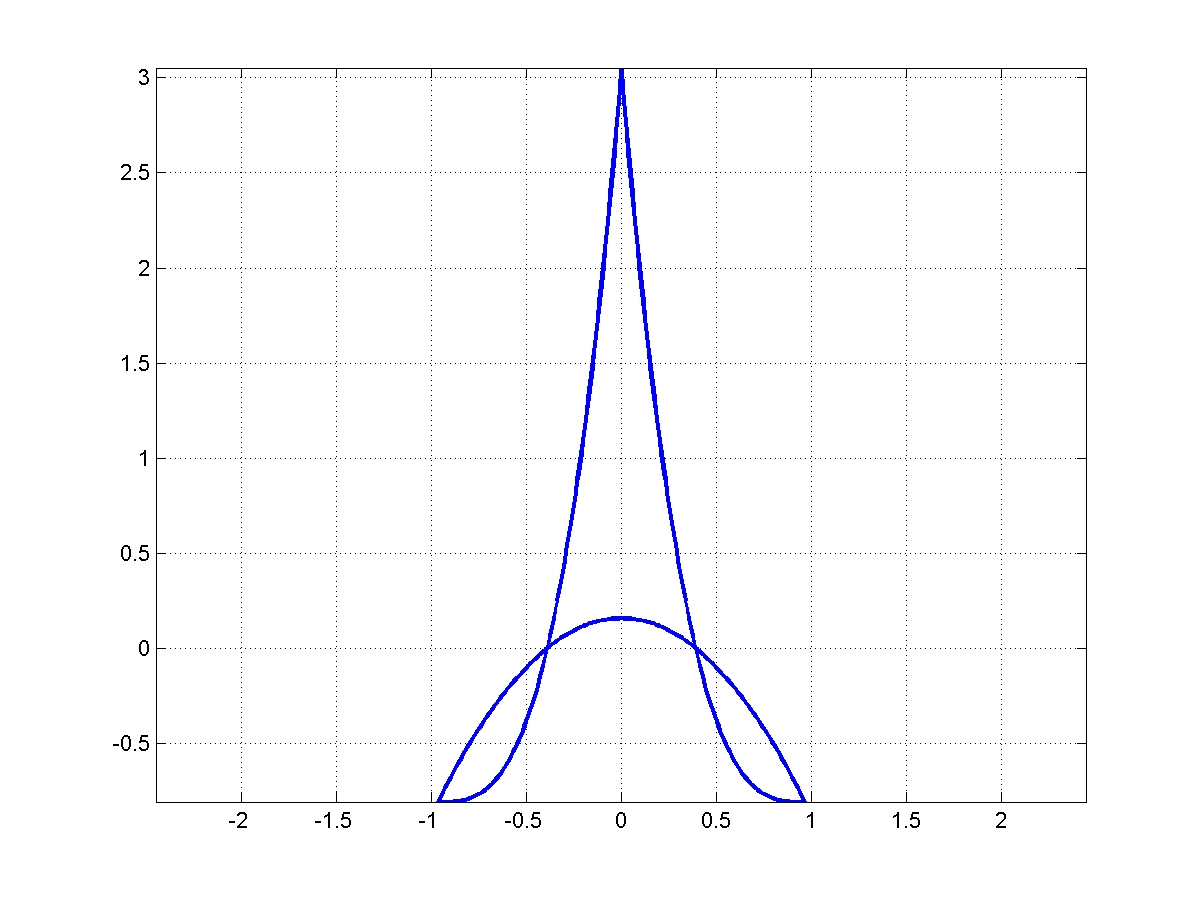}\includegraphics[width=4.5cm]{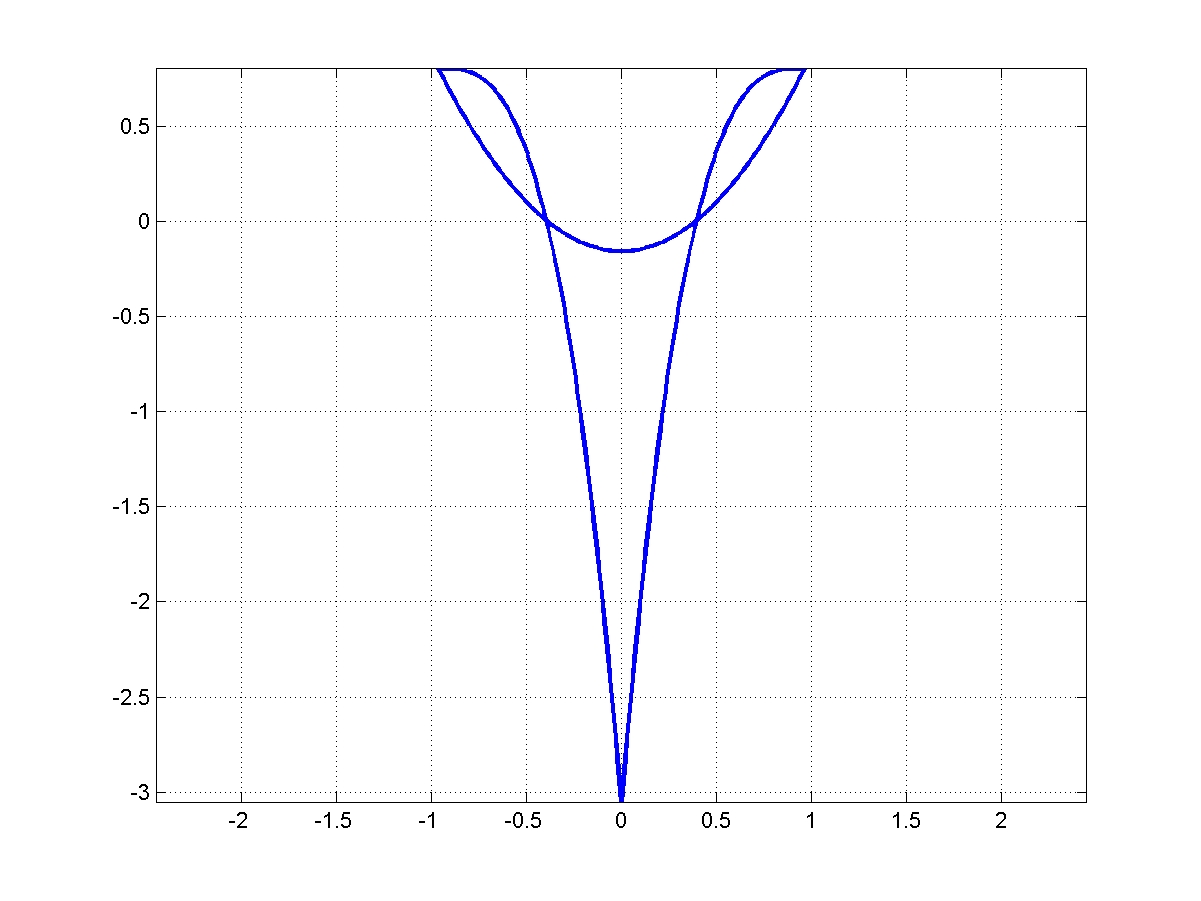}}
\caption{Ambiguity in PCA-based orientation of general shapes: only the principal axis is determined, not its \emph{ direction}.\label{fig:ambill}}
\end{figure}

\begin{figure}[hbt]
\centerline{\includegraphics[width=4.5cm]{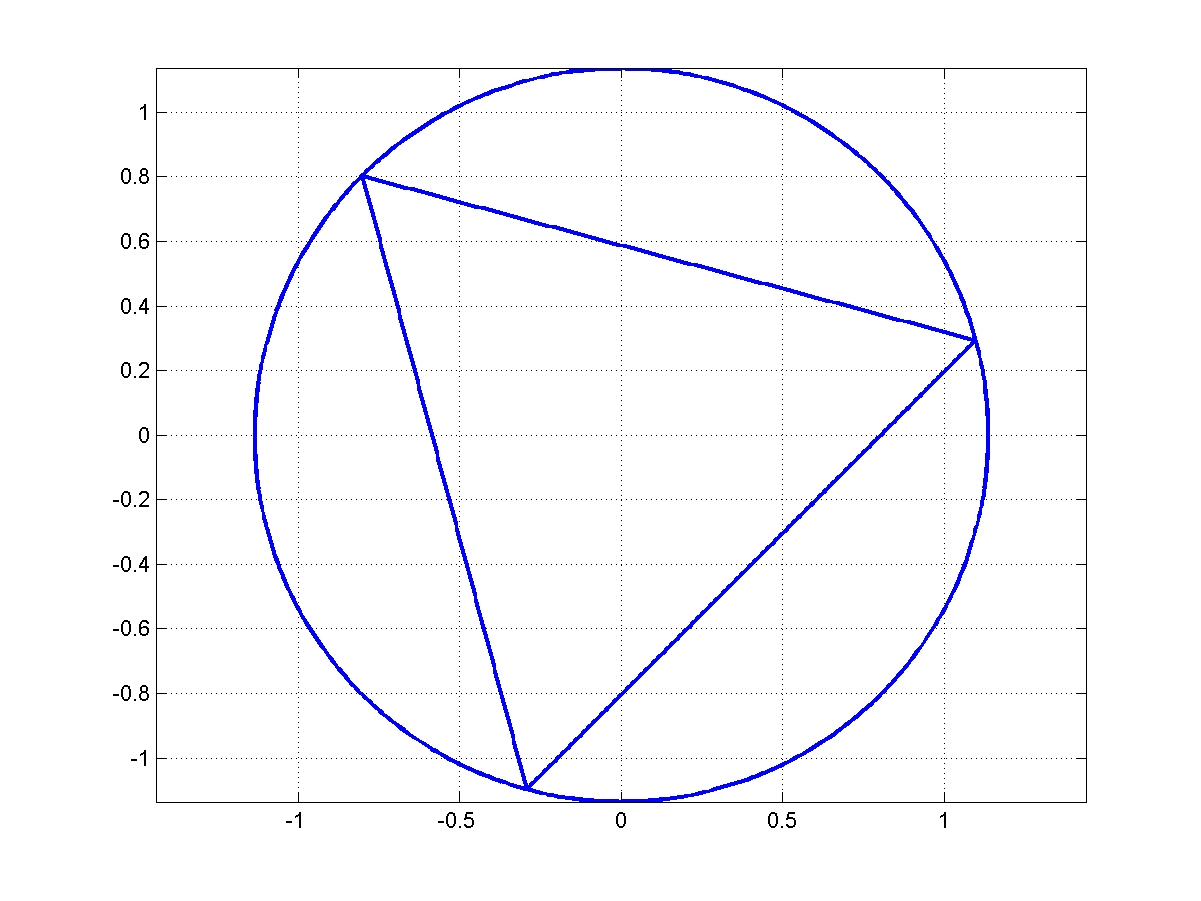}\includegraphics[width=4.5cm]{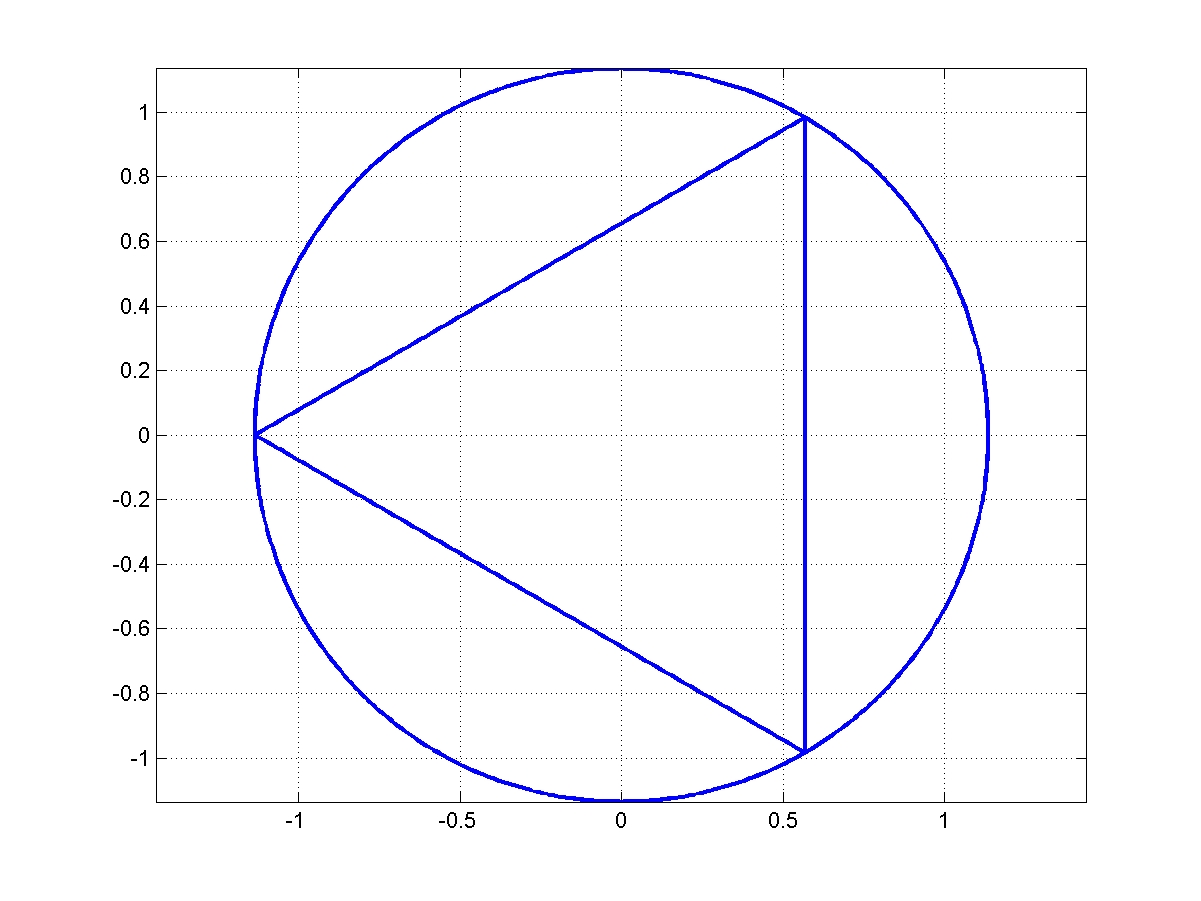}}
\centerline{\includegraphics[width=4.5cm]{equalill3}\includegraphics[width=4.5cm]{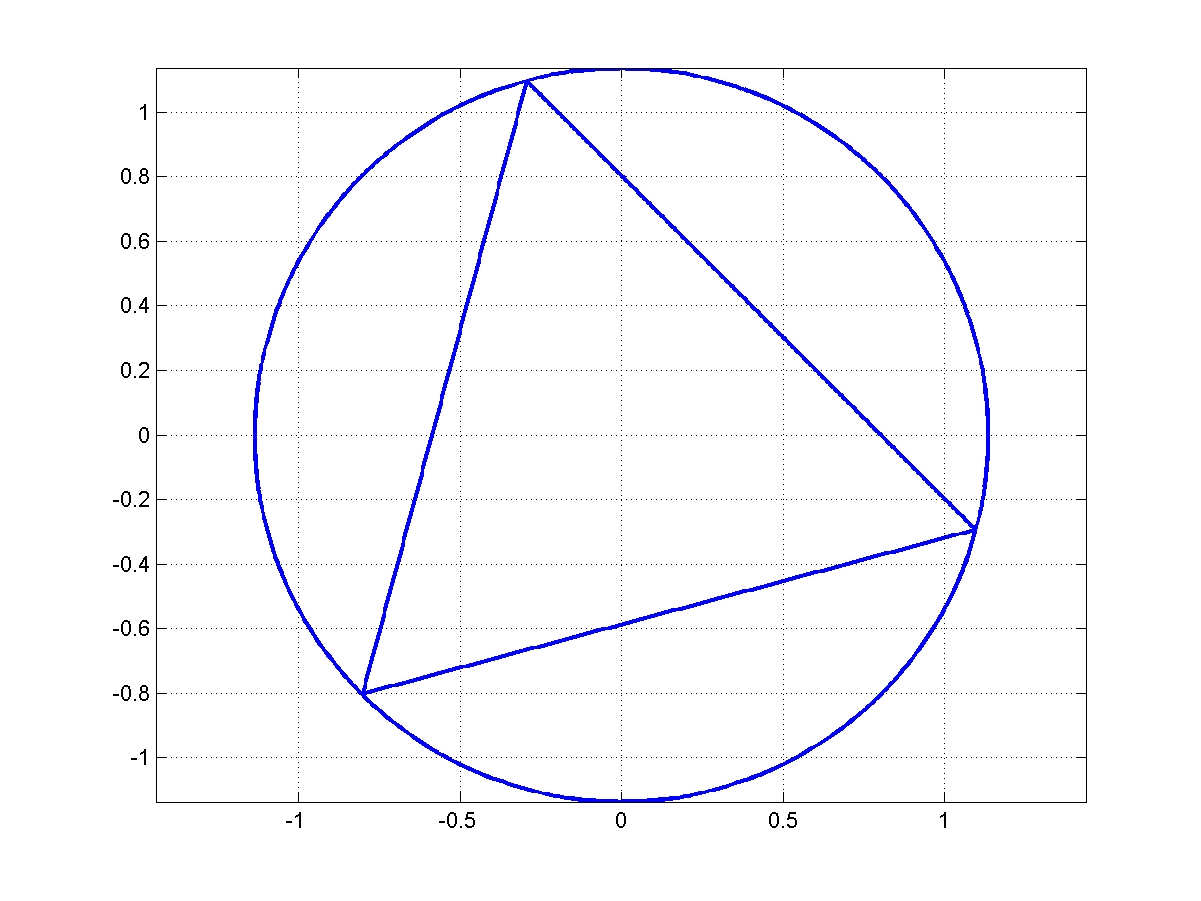}}
\caption{A rotationally symmetric shape. Since for this kind of shapes the {\it principal axis is not defined}, PCA can not be used to compute their orientation.\label{fig:equalill}}
\end{figure}

The most straightforward method to define orientation uses the principal axis of the shape, obtained, {\it e.g.}, through Principal Component Analysis (PCA). Although unable to provide a unique orientation, this method defines two possible orientations for elongated shapes, see Fig.~\ref{fig:ambill} (the directional ambiguity happens in general, not only for mirror-symmetric shapes such as the one used for illustration). For shapes that do not have a well defined principal axis, {\it e.g.}, rotationally symmetric shapes, PCA-based orientation is completely ambiguous, see Fig.~\ref{fig:equalill}.

To deal with these ambiguities, researchers attempted to work with concepts like mirror-symmetry axes \citep{atallah85,marola89}, universal principal axis \citep{lin93}, and generalized principal axis \citep{tsai91,zunic06}. In general, the motivation for these works is more on the definition of a ``reasonable geometric orientation'' than on the robust computation of a unique orientation angle for arbitrary shapes. Since rotationally symmetric shapes are particularly challenging, the automatic detection of symmetry and fold number, by itself a relevant problem, has also received attention \citep{lin94,shen99,derrode04,prasad04}.

The more theoretically sustained methods to compute orientation are based on the geometric moments of the points defining the shape. In particular, the so-called Complex Moments (CMs) were introduced in the eighties \citep{mostafa85,teh88}. The elegance of these approaches comes from defining the orientation through the phase of a single CM of a particular order. In the nineties, more general moments were proposed to deal with degenerate shapes \citep{shen97,shen99}, at a cost of dealing with several moments, chosen by tuning a free parameter index through search, and detecting rotational symmetry as an intermediate step. However, as we detail in Section~\ref{sec:mom}, these methods do not cope with several shapes that lead to singular moments. In practice, this means that the phase of these moments is sensitive to noise, leading to unstable estimates of orientation. Other approaches require the exhaustive search for the angle maximizing a given orientation measure \citep{ha03,ha05}, without any guarantee of uniqueness of the solution.

In this paper, we address the need to combine compact descriptions of 2D shapes, for computational efficiency, with the invariance and discriminative power of complete sets of invariants. We propose to represent 2D shapes in terms of particular complex moments, which we call the {\it Principal Moments}~(PMs). Although moments of image patterns have been extensively used due to their invariance properties, since at least the early sixties \citep{hu62}, their discriminative properties have not been studied even in more recent related work \citep[{\it e.g.},][]{mostafa85,khotanzad90,shen99,mukundan01}. In opposition, our representation uniquely defines the shape. In fact, using the same number of PMs as the number of shape landmarks, our representation forms a complete set of invariants with respect to permutation.
We further show that the PMs coincide with the coefficients of the Fourier series of the representation of \cite{rodrigues08a}, a result that guarantees that our representation inherits the discriminative power demonstrated by the experiments reported in that paper. Subsequently, we derive an upper bound for the magnitude of these coefficients in terms of the shape complexity, {\it i.e.}, of the number of landmarks. Using these results, we show that our representation is compact, in the sense that a small number of PMs (much smaller than the number of landmarks) suffices to represent the shape. This compactness contrasts with the usual large dimension of other complete representations ({\it e.g.}, those based on the bispectrum \citep{kondorthesis08} or the densely sampled functions of \cite{rodrigues08a}), an issue of outmost relevance when working with large databases.

In what respects to geometric transformations, besides trivially extending the representation to deal with translation and scale transformations, a major contribution of this paper is the extension to also include rotations.
This extension consists in previously rotating each shape instance of a normalization angle which defines the shape orientation. Overcoming the limitations discussed above, we present a new method to define and compute a unique orientation of any 2D~shape, based on its PMs. More specifically, we show that the phases of two of these moments unambiguously define the orientation of an arbitrary 2D~shape (including rotationally symmetric ones) and propose an algorithm, {\it Principal Moment Analysis} (PMA), that computes the orientation angle by integrating the contributions of all pairs of moments.

Naturally, PMA can be used to normalize arbitrary 2D shapes, {\it e.g.}, binary images, with respect to orientation, before any other processing takes place. In the paper we also discuss the straightforward extension of PMA to the normalization of general gray-level images.

Besides theoretically sound, PMA results are robust to noise, as the experiments in the paper illustrate.

The remaining of the paper is organized as follows. Section~\ref{sec:pms} introduces the PMs and relates them to previously proposed moments. In Section~\ref{sec:ansig}, we relate the PMs with the representation of \cite{rodrigues08a} and derive an upper bound for the length containing most of its energy, which enables ending up with a compact shape representation. Section~\ref{sec:mri} describes our approach to extend the representation towards obtaining a complete set of invariants with respect to 2D~rotation. In Section~\ref{sec:mom}, we detail the limitations of current methods when estimating 2D~orientation.
Section~\ref{sec:pma} presents PMA, our algorithm for computing a unique orientation of an arbitrary 2D~shape
from its PMs. In Section~\ref{sec:grey}, we extend PMA to the orientation normalization of gray-level images. Section~\ref{sec:exp} contains experiments and Section~\ref{sec:conc} concludes the paper.

\section{Principal Moments for 2D Shape Representation}
\label{sec:pms}

Consider an arbitrary 2D shape described by a set of $N$ points in the plain, thus by an $N$-dimensional complex vector $\bz\in {\mathbb C}^N$, containing their coordinates:
\begin{equation}\label{eq:shapevector}
\bz=\left[
      \begin{array}{c}
        x_1+jy_1 \\
        x_2+jy_2 \\
        \vdots \\
        x_N+jy_N \\
      \end{array}
    \right]=\left[
\begin{array}{c}
        z_1 \\
        z_2 \\
        \vdots \\
        z_N \\
      \end{array} \right]\,.
\end{equation}
Naturally, since the points do not have labels, the same shape can be described by any vector obtained from $\bz$ by re-ordering its entries.

We define the $k^{\mbox{\scriptsize th}}$-order \emph{Principal Moment} (PM), $k \in \{1, 2, 3, \ldots\}$, by
\begin{equation}
M_k(\bz)=\frac{1}{N k!} (z_1^k+z_2^k+\cdots+z_N^k) = \frac{1}{N k!}\sum_{n=1}^{N}z_n^k\,.\label{eq:pm}
\end{equation}
The $k^{\mbox{\scriptsize th}}$~PM, after stripping off the scaling factor $1/(N k!)$, is also known as the $k^{\mbox{\scriptsize th}}$~\emph{power sum} of the landmarks $\{z_1, z_2, \ldots, z_N\}$. The inclusion of this particular scaling factor in the definition \eqref{eq:pm} is motivated in the next section. Through the Newton's identities, the first $N$ power sums can be unambiguously converted into the so-called \emph{fundamental symmetric polynomials} of the landmarks. As these polynomials can be seen as the coefficients (up to sign changes) of the $N^{\mbox{\scriptsize th}}$-order univariate polynomial containing the landmarks as roots, there is a bijection between this univariate polynomial and the first $N$ power sums. The reader is referred to the enlightening book of \cite{kanatani} for what respects to these equivalences. Since an $N^{\mbox{\scriptsize th}}$-order univariate polynomial unequivocally defines its $N$ roots, although orderless, the first $N$ power sums unambiguously define the shape $\bz$, up to a permutation, and so do the first $N$ PMs. This $N$-sized representation is thus said to form a \emph{complete set of invariants} over the permutation group, or, equivalently, to be \emph{maximally invariant} to permutations. In other words, (i) any vector obtained from $\bz$ by re-ordering its entries leads to the same PMs (permutation invariance); and (ii) any vector with at least one landmark lying at a different position than the ones in $\bz$, leads to distinct PMs (discrimination).

The defined representation can be trivially extended to be maximally invariant with respect to simple geometric transformations. Indeed, it is easily shown that the simple pre-processing step of working with
\begin{equation} \label{eq:translation_scale_norm}
    \sqrt{N}\frac{\bz-\overline{\bz}}{\left\|\bz-\overline{\bz}\right\|} \,,
\end{equation}
where $\overline{\bz} = \frac{1}{N} \sum_{n=1}^N z_n$ denotes the center of mass of $\bz$ and $\|\cdot\|$ denotes the $l_2$ norm, rather than directly with $\bz$, extends the complete set of invariants to include translation and scale transformations. The extension with respect to rotation operations, though, is not so trivial and it is, along with the subject of shape representation using PMs, the central subject of this paper.

Before proceeding, we relate the PMs in \eqref{eq:pm} with the more general \emph{Complex Moments}~(CMs) of \cite{mostafa85} and \emph{Generalized Complex}~(GC) moments of \cite{shen97,shen99}. The CM of order $(p,q)$ of an image $g(x,y)$ is defined by
\begin{equation}
C_{pq}(g)=\int\!\int_{-\infty}^{+\infty} \left(x+jy\right)^p \left(x-jy\right)^q
g(x,y) \,dx\,dy\,,\label{eq:cms}
\end{equation}
where $p\geq 0$ and $0\leq q\leq p$ \citep{mostafa85}. Considering an image composed by a set of $N$ mass points located at the shape landmarks $\{z_1, z_2, \ldots, z_N\}$, the integral in \eqref{eq:cms} becomes a sum:
\begin{equation}\label{eq:cm1}
C_{pq}(\bz)=\sum_{n=1}^{N}z_n^p (z_n^*)^q \,,
\end{equation}
where ${z_n^*}$ denotes the complex conjugate of $z_n$. In turn, the GC moment of order $(p,q)$ is given by the integral in polar coordinates
\begin{equation} \label{eq:gci}
GC_{pq}(g) =\int_{-\pi}^{\pi}\int_{0}^{\infty}r^p
e^{jq\theta}g(r\cos \theta , r\sin \theta)\,r\,dr\,d\theta\,,
\end{equation}
where $p\in\left\{0,1,2,\ldots\right\}$ and $q\in\left\{1,2,3,\ldots\right\}$ \citep{shen97}. For a shape $\bz$, the GC moments collapse into the sums
\begin{equation}\label{eq:gcz}
GC_{pq}(\bz)=\sum_{n=1}^{N}\left|z_n\right|^{p}e^{jq\arg z_n}\,.
\end{equation}

It is now clear that the PMs are CMs and GC moments of particular orders (up to a scaling factor): from (\ref{eq:pm},\ref{eq:cm1},\ref{eq:gcz}), $M_k(\bz)=C_{k0}(\bz)/(Nk!)=GC_{kk}(\bz)/(Nk!)$. Finally, note also that, for any shape, $M_0(\bz)=\frac{1}{N}\sum_{i=1}^{N}1=1$, and, assuming the shapes were pre-processed as in \eqref{eq:translation_scale_norm}, $M_1(\bz)=\frac{1}{N}\sum_{n=1}^{N}z_n=0$.

\section{A Compact Shape Representation} \label{sec:ansig}

We now show why a representation that uses a small subset of PMs suffices in practice. To do this, we start by relating the PMs with the {\it analytic signature} (ANSIG), the representation introduced by \cite{rodrigues08a}. The ANSIG is an analytic function on the complex plane, obtained from $\bz$ through
\begin{equation}
a({\boldsymbol z},\xi) = \frac{1}{N} \sum_{n=1}^N e^{z_n \xi}\,. \label{eq:ansig}
\end{equation}
\cite{rodrigues08a} show that this representation is complete and permutation-invariant and thoughtfully illustrate its capabilities with several shape-based image classification experiments. In these experiments, the analytic function \eqref{eq:ansig} is described in the computer by its 512 samples uniformly taken on the unit-circle of the complex plane. Working with these high dimensional vectors may be adequate for tasks requiring the comparison of a small number of shapes but certainly not for applications that deal with very large databases, {\it e.g.}, the internet.

In the sequel, we derive that our PMs \eqref{eq:pm} are intimately related with the ANSIG \eqref{eq:ansig} and show that a small number of either PMs or of ANSIG samples suffices to represent the shape with similar performance.

\subsection{The Principal Moments and the ANSIG Spectrum}

A direct consequence of Cauchy's integral formula \citep[see, {\it e.g.},][]{ahlfors} is that any analytic function is fully specified by the values it takes on a closed contour
on the complex plane. Thus, the ANSIG in \eqref{eq:ansig} is equivalently described by its restriction to the unit-circle,
\begin{equation}
h(\bz,\theta)=a(\bz,e^{j\theta})=\frac{1}{N} \sum_{n=1}^N \exp\left(z_n e^{j\theta}\right)\,.\label{eq:cauchy}
\end{equation}

To approximate the unit-circle restriction of the ANSIG by a finite-dimensional computer representation, we study its frequency spectrum. Note that $h(\bz,\theta)$ in \eqref{eq:cauchy} can be seen as a real-argument complex-valued periodic function, with fundamental period $T=2\pi$ and fundamental frequency $\omega_0={2\pi}/{T}=1$. Thus, it can be written in terms of its Fourier series,
\begin{equation}
h(\bz,\theta)=\sum_{k=-\infty}^{+\infty} H_k(\bz)\, e^{jk\theta} \,, \label{eq:fourier1}
\end{equation}
where each coefficient $H_k$ is given by \citep[see, {\it e.g.},][]{oppss}:
\begin{equation}
H_k(\bz)=\frac{1}{2\pi}\int_{-\pi}^{\pi}h(\bz,\theta)\,e^{-jk\theta}\,d\theta\,.\label{eq:fourier2}
\end{equation}

The analysis expression \eqref{eq:fourier2} is hard to carry out in the case of $h(\bz,\theta)$ given by \eqref{eq:cauchy}. However, the coefficients of the
Fourier series easily follow from the comparison of the synthesis expression \eqref{eq:fourier1} with the definition of $h(\bz,\theta)$ in \eqref{eq:cauchy}, after some manipulations. In fact, expressing the exponential in \eqref{eq:cauchy} by its Maclaurin series ({\it i.e.}, its Taylor series at the origin), we get
\begin{align}
h(\bz,\theta)&=\frac{1}{N} \sum_{n=1}^N \sum_{k=0}^\infty \frac{\left(z_n \,e^{j\theta}\right)^k}{k!}\nonumber \\
&= \sum_{k=0}^\infty \left(\frac{1}{Nk!} \sum_{n=1}^{N} z_n^k \right) e^{j k\theta}\,.\label{eq:mclaurin}
\end{align}
Comparing \eqref{eq:mclaurin} with \eqref{eq:fourier1}, and since the coefficients of the Fourier series representation are unique \citep{oppss}, we conclude that
\begin{equation}
    H_k(\bz) =
    \begin{cases}
        \frac{1}{Nk!} \sum_{n=1}^{N} z_n^k & \mbox{for}\;\;\; 0\leq k<+\infty\,, \\
        0 & \mbox{for}\;\; -\infty<k\leq-1\,.\label{eq:hk}
    \end{cases}
\end{equation}

Thus, given a 2D shape described by a set of $N$ landmarks, expression \eqref{eq:hk} relates the values of the coefficients of the Fourier series of the unit-circle restriction of its ANSIG, $\left\{H_k,-\infty<k<+\infty\right\}$, to the positions of the landmarks in the plane, $\left\{z_1, z_2, \ldots, z_N\right\}$. Comparing with \eqref{eq:pm}, we conclude that these coefficients for $k\geq 0$ coincide with the PMs:
\[
    H_k(\bz) = M_k(\bz), \quad k = 0, 1, 2, \ldots \,.
\]
Due to the one-to-one correspondence just proved between $M_k(\bz)$, $h(\bz,\theta)$ and $a(\bz, \xi)$, the PM shape representation inherits all the properties of the ANSIG. Although the fact that the PMs constitute a complete representation was already referred in the previous section (and the fact that they enjoy permutation invariance is immediate from definition \eqref{eq:pm}), we now know they are equivalent to the ANSIG also in what respects to discrimination capabilities.

\subsection{Compactness of the Principal Moments}

We now show that the proposed representation is compact, in the sense that only a small number of PMs is enough to represent the shape with neglectable loss of discrimination power. For that purpose, we derive an upper bound $b(k)$ for the magnitude of the coefficient $H_k$ (or, equivalently, the PM $M_k$), for $k\geq 2$, in terms of the number of landmarks describing the shape. Using this result, we deliver an upper limit for the (wide-sense) bandwidth of the unit-circle restriction $h(\bz,\theta)$ of the ANSIG. The first step is done through the following chain of equalities and inequalities:
\begin{eqnarray} |M_k| &=& \frac{1}{Nk!} \left| \sum_{n=1}^{N}
z_n^k \right| \;\leq\; \frac{1}{Nk!} \sum_{n=1}^{N}
\left|z_n\right|^k \label{eq:ineq1}\\
&=& \frac{1}{Nk!} \sum_{n=1}^{N} \left(|z_n|^2\right)^{\frac{k}{2}} \;\le\;
\frac{1}{Nk!} \left( \sum_{n=1}^{N} |z_n|^2
\right)^{\frac{k}{2}} \label{eq:ineq2} \\
&=&\frac{N^{\frac{k}{2}-1}}{k!}\;\stackrel{\mathrm{def}}{=}\;b(k)\,, \label{eq:final}
\end{eqnarray}
where \eqref{eq:ineq1} uses the triangle inequality, \eqref{eq:ineq2} uses the fact that the function $(a+b)^{n}$ is convex for $a,b$ real positive and $n\geq
1$, and \eqref{eq:final} is due to the fact that $\sum_{n=1}^{N}\left|z_n\right|^2=N$, for shape vectors $\bz$ normalized according to \eqref{eq:translation_scale_norm}.

To estimate the bandwidth of $h(\bz,\theta)$ in terms of the number of landmarks, usually, we would seek the smallest $k$ such that the ratio $|M_k|/|M_0|$ is below a given threshold~$p$. In our case, since $M_0=1$, this would lead to $|M_k|<p$. Nevertheless, as we do not know the exact value of $|M_k|$, we will use the upper bound $b(k)$ as a proxy. We are not sure to obtain the smallest~$k$ that satisfies our bandwidth constraint, but we guarantee the satisfaction of the inequality, since $|M_k|\leq b(k)$.

Finding the smallest $k$ such that $b(k)<p$ requires solving the limit case equation $N^{\frac{k}{2}-1}/k!=p$, for which there is no analytic solution. We propose a simple method to solve for $k$ numerically. Due to the fast grow of $k!$ and to increase stability, we apply logarithms on both sides of the inequality. Denoting the natural logarithm of $b(k)$ by $B(k)$, we have:
\begin{eqnarray}
B(k)&=&\ln b(k)\,=\,\left(\frac{k}{2}-1\right)\ln
N-\ln k! \label{eq:log1}\\ &\simeq& k \left(
\frac{\ln N}{2}\! + \!1\! \right) - k\, \ln k \nonumber \\
&&- \frac{1}{2} \ln k - \ln N -
\frac{1}{2} \ln 2\pi\,,\label{eq:log2}
\end{eqnarray}
where \eqref{eq:log1} uses the definition of $b(k)$ in \eqref{eq:final} and \eqref{eq:log2} uses the Stirling's approximation $\ln k!\simeq k\ln k-k+\frac{1}{2}\ln 2\pi k$ \citep[see, {\it e.g.},][]{paris01}.

\begin{figure}[htb]
        \centering
               \includegraphics[width=8.5cm]{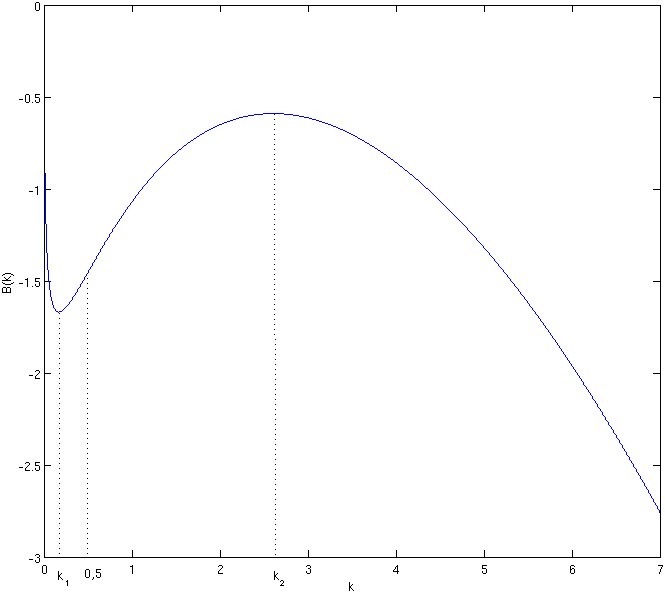}
        \caption{Upper bound $B(k)$ for the (logarithm of the) magnitude of the
        spectrum of the analytic signature (or, equivalently, magnitude of the principal moments), for $N=10$ points.}
        \label{fig:Bk}
\end{figure}

To analyze the behavior of $B(k)$ given by \eqref{eq:log2}, we relax $k$ to the reals and express the first two derivatives:
\[
B'(k) = \frac{\ln N}{2} -\ln k - \frac{1}{2k} \,,\qquad
B''(k) = -\frac{1}{k} + \frac{1}{2k^2}\,.
\]
From these expressions, we see that $B(k)$ has an inflection at $k=1/2$, where $B''(k)=0$, and two extrema at $k=k_1<1/2$ and $k=k_2>1/2$, where $B'(k)=0$ (assuming $N\geq 2$). 
Furthermore, $B(k)$ monotonically decreases for $k>k_2$, where $B'(k)<0$, being $\lim_{k\to+\infty} B(k) = -\infty$. The plot in Fig.~\ref{fig:Bk} illustrates the behavior of $B(k)$ for $N=10$. To find $k$ such that $B(k)$ is below a given threshold $\ln p$, we thus propose the following strategy in two steps: first, solve $B'(k) = 0$ in the interval $k\in[1/2, +\infty)$, obtaining $k_2$. Then, solve $B(k) - \ln(p) = 0$ in the interval $k\in[k_2, +\infty)$, obtaining
$k=k_B$, the desired upper bound for the bandwidth of the ANSIG. Note that the first step is necessary to specify the lower limit of the search region for the second one: without that limit, we could obtain a spurious solution $k_B < k_2$.

\begin{figure}[htb]
        \centering
              \includegraphics[width=8.5cm]{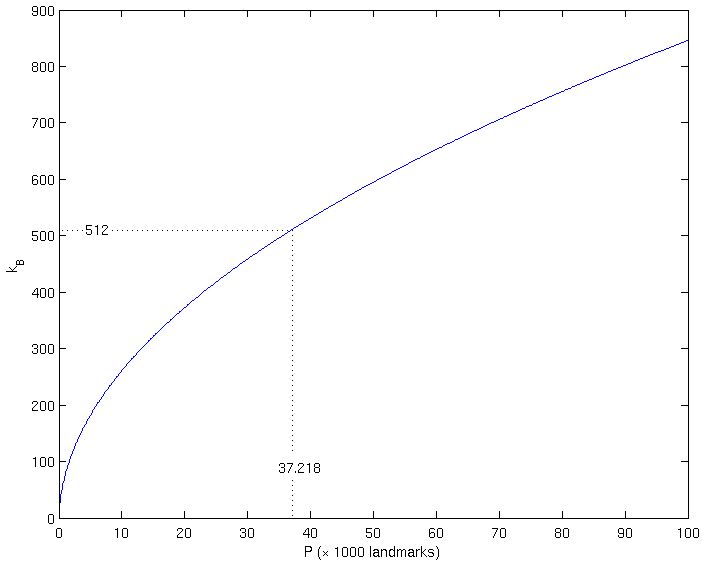}
        \caption{The number of coefficients $k_B$ needed to represent a shape described by $N$ landmarks.}
        \label{fig:poupar}
\end{figure}

We thus conclude that most of the energy (the parameter $p$ controls the amount) of the ANSIG of a shape described by $N$ landmarks is contained in a number $k_B$ of complex coefficients, which, naturally, depends on $N$. Fig.~\ref{fig:poupar} plots the number of coefficients $k_B$, computed as described above, as a function of the number of landmarks~$N$, for $p=0.1$ $(-20\mbox{dB})$. Obviously, $k_B$ can be indistinctly interpreted as either the required number of Fourier series coefficients, {\it i.e.}, the number of PMs, or the required number of samples in the unit-circle to represent the shape. In fact, since the fundamental frequency is $\omega_0=1$, the approximate bandwidth of the signal is $\omega_B = k_B \omega_0 = k_B$. Since the spectrum of $h(\bz,\theta)$  is zero for negative frequencies, {\it c.f.} \eqref{eq:hk}, it suffices to sample at a rate (number of points) of $N = \omega_s = \omega_B = k_B$ (the Nyquist sampling rate $\omega_s$ of twice the bandwidth is only required for two-sided spectra~\citep{oppss,opp}). Naturally, to recover the original continuous ANSIG from these samples, we should use a (complex coefficient) filter with passband $\omega \in [0, \omega_s)$ (in opposition to the traditional low-pass filter with cutting frequency $\omega_s/2$).

The plot in Fig.~\ref{fig:poupar} also compares the required number $k_B$ of samples, or of PMs, with 512, the fixed number of samples used in~\cite{rodrigues08a}: while for shapes described by an huge number of points (more than $\simeq 40000$), $512$ samples may not be enough, for the majority of cases that may arise in practice (a few hundreds of landmarks), the required number is much smaller (a few dozens). Note further that $k_B$ is smaller than $N$, making the representation based on PMs loose its maximal invariance to permutations, in a strict sense, when using $k_B$ coefficients. Nevertheless, the discrimination loss that results from this is small, since most of the energy of the signature is captured by the first $k_B$ coefficients \footnote{Naturally, the compactness of the representation is due to the decay of $|M_k|$ with $k!$ imposed by the normalization factor in \eqref{eq:pm}, which is now motivated. In Appendix~\ref{app:norm}, we further discuss the issue of normalizing power sums.}.

Having motivated the usage of only $k_B$ PMs as a way to compactly represent shapes in terms of its complexity, {\it i.e.}, its number of landmarks, we now
discuss the comparison of descriptions of shapes of distinct complexity. Since the first $k_B$ PMs are (the most relevant) coefficients of the ANSIG Fourier series, it suffices to pad with zeros the smaller vector before performing the comparison in frequency domain. If, in opposition, the shapes are equivalently described by the sparse set of $k_B$ ANSIG samples, say $N_1$ samples for one of the shapes and $N_2$ samples for the other, it is necessary to use multirate signal processing techniques to convert both to a common sampling rate \citep[see, {\it e.g.},][]{opp}. For example, perform upsampling of the ANSIGs by a factor of, respectively, $L_1 = \mbox{lcm}(N_1, N_2)/N_1$ and $L_2 = \mbox{lcm}(N_1, N_2)/N_2$, where $\mbox{lcm}$ stands for the least common multiple, followed by interpolation using (complex coefficient) filters with passband, respectively, $\omega \in [0, 2\pi/L_1)$ and $\omega \in [0, 2\pi/L_2)$.

\section{Maximal Rotation Invariance}
\label{sec:mri}

The representation of a 2D shape by its PMs forms a complete set of invariants with respect to permutation. The extension of the set to include translation and scale invariance is easily obtained by removing the mean and normalizing the energy, according to \eqref{eq:translation_scale_norm}. However, taking care of the orientation is not obvious. In the following, we propose an extension to include maximal invariance to rotations by computing a unique orientation, definable for an arbitrary shape.

A natural and the most common way to attempt to obtain rotation invariance consists in finding an angle $\theta(\bz)$ such that, through rotation, any shape $\bz$ is brought to its ``normalized'' version
\begin{equation}\label{eq:norm}
\bw(\bz)=\bz e^{-j\,\theta(\bz)}\,.
\end{equation}
In fact, if the desired invariance is satisfied, {\it i.e.}, if,
\begin{equation}\label{eq:inv}
\forall_\phi\,,\quad \bw\left(\bz e^{j\phi}\right)=\bw(\bz)\,,
\end{equation}
the normalization in~\eqref{eq:norm} produces a maximal invariant. However, current methods to compute the shape orientation $\theta(\bz)$ either fail to process particular shapes (see examples in Figs.~\ref{fig:ambill} and~\ref{fig:equalill} and others in the following section) or do not guarantee the equality in~\eqref{eq:inv}.

The success of the normalization in~\eqref{eq:norm} hinges then on finding an appropriate function $\theta: {\mathbb C}^N\rightarrow (-\pi,\pi]$ that unambiguously defines $\theta(\bz)$, the orientation of the shape $\bz$. We now show that, to guarantee the desired invariance, it suffices that this function
satisfies a natural condition: that the orientation of a rotated shape is the sum of the orientation of the original shape with the rotation angle. Formally, this condition means that the function $\theta(\cdot)$ has to satisfy
\begin{equation} \label{eq:equiv_thetac}
\forall_\phi\,,\quad    \theta\left(\bz e^{j \phi}\right) = \theta(\bz) +
\phi\,,
\end{equation}
where the equality is modulo $2\pi$. Simple manipulations show that \eqref{eq:equiv_thetac} suffices to guarantee \eqref{eq:inv}:
\begin{align}
    \bw\left(\bz e^{j \phi}\right) &= \bz e^{j \phi}\exp\left(-j\,\theta\!\left(\bz e^{j
    \phi}\right)\right)\label{eq:part1}\\
 &=\bz e^{j \phi}e^{-j\left(\theta\left(\bz\right)+\phi\right)}\label{eq:part2}\\
 &=\bz e^{-j\,\theta\left(\bz\right)}\nonumber\\
 &=\bw(\bz)\label{eq:part4}\,,
\end{align}
where \eqref{eq:part1} and \eqref{eq:part4} use the definition \eqref{eq:norm} and \eqref{eq:part2} uses \eqref{eq:equiv_thetac}.

In Section \ref{sec:pma}, we propose an orientation function $\theta(\cdot)$ satisfying \eqref{eq:equiv_thetac}. This is a relevant contribution to maximally invariant representations for 2D shapes, since the pre-processing step \eqref{eq:norm} extends any permutation-invariant representation, like the ones proposed by~\cite{jebara03} and~\cite{rodrigues08a}, or the PMs introduced in Section \ref{sec:pms} of this article, to also accommodate maximal invariance with respect to arbitrary geometric transformations, {\it i.e.}, including rotation.

\section{Limitations of Previous Approaches to Moment-based Orientation Normalization}
\label{sec:mom}

To obtain the orientation $\theta(\bz)$ of a shape $\bz$, we use the PMs of the points describing the shape. Since image moments have been used in the past, we first overview moment-based estimation of orientation and motivate the need to revisit the problem. The usage of Complex Moments~(CMs) to define orientation was proposed in \cite{mostafa85}. CMs stand for compact representations of linear combinations of ordinary ({\it i.e.}, real) geometric moments. In that work, the authors define and compute the orientation by imposing the phase of one of the moments $C_{q+1,q}$ in \eqref{eq:cms} to be zero. When applying this method to a shape $\bz$, {\it i.e.}, to an image composed by a set of $N$ mass points describing the shape, we obtain, through \eqref{eq:cm1}, the moments
\begin{equation}\label{eq:cm1_qqp1}
C_{q+1,q}(\bz)=\sum_{n=1}^{N}\left|z_n\right|^{2q+1}e^{j\arg z_n}\,,
\end{equation}
where, as introduced in Section~\ref{sec:pms}, $z_n$ collects the coordinates of the $n^{\mbox{\scriptsize th}}$ shape point.

\begin{figure}[hbt]
\centerline{\includegraphics[width=7cm]{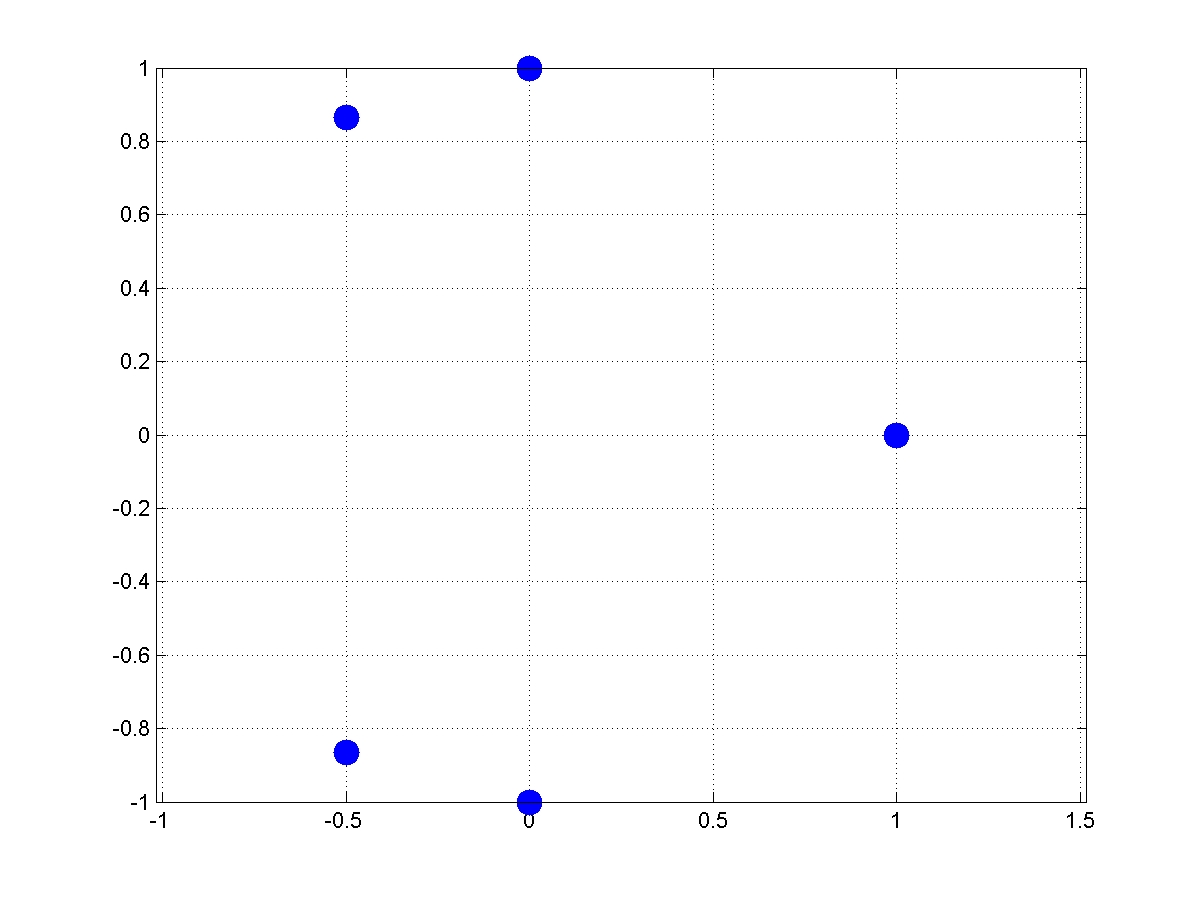}} \caption{Example of a shape
(composed by the points $\left\{1,\pm j, \exp(\pm j2\pi/3)\right\}$), for which
all the complex moments $C_{q+1,q}$ are zero, making impossible to define its
orientation in terms of the phase of such moments.\label{fig:ex-nrsi}}
\end{figure}

\begin{figure}[hbt]
\centerline{\includegraphics[width=7cm]{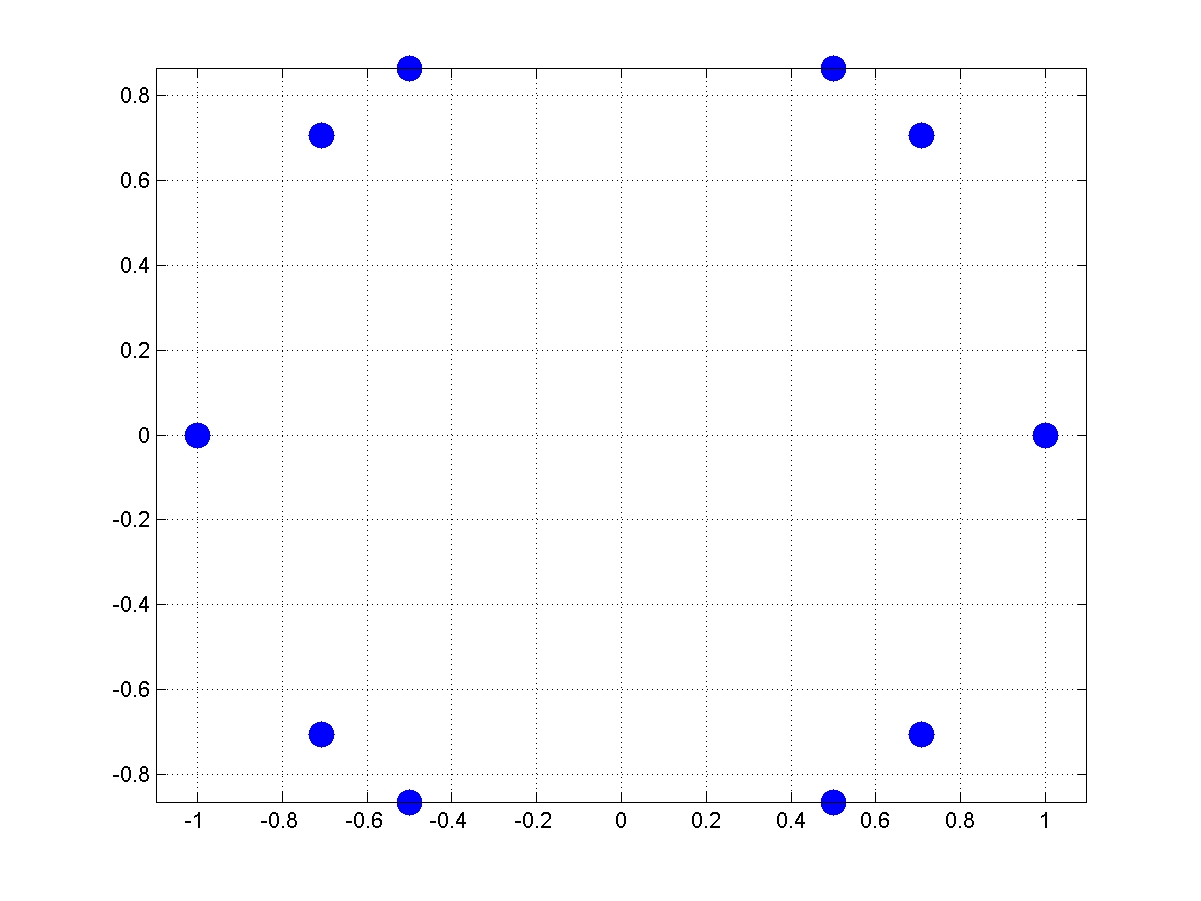}} \caption{Example of a 2-fold
rotationally symmetric shape (composed by the points $\{\pm 1,\pm \exp(\pm
j\pi/3),\pm \exp(\pm j\pi/4)\}$), for which all the complex moments $C_{q+2,q}$
are zero, making impossible to define its orientation in terms of the phase of
such moments.\label{fig:ex-rsi2}}
\end{figure}

Although the method just described is adequate to deal with shapes $\bz$ that lead to a moment $C_{q+1,q}(\bz)$ with large magnitude, there are shapes for which this does not happen for any $q$. It was known that this is the case of rotationally symmetric shapes \citep{mostafa85}, but we now show it may also happen with general ones. Just look at the example in Fig.~\ref{fig:ex-nrsi}, where $\bz=[1,j,-j,\exp(j2\pi/3),\exp(-j2\pi/3)]^T$. For this shape, from~\eqref{eq:cm1_qqp1}, we obtain that all the moments $C_{q+1,q}(\bz)$ are zero, regardless of $q$:
\[
C_{q+1,q}(\bz)=e^{j0}+e^{j\pi/2}+e^{-j\pi/2}+e^{j2\pi/3}+e^{-j2\pi/3}=0\,.
\]
For $S$-fold rotationally symmetric shapes, \cite{mostafa85} propose to use the phase of one of the moments $C_{q+S,q}$. However, again, there are $S$-fold rotationally symmetric shapes for which these are all zero. For example, it is straightforward to show that the $2$-fold rotationally symmetric shape in Fig.~\ref{fig:ex-rsi2} leads to $C_{q+2,q}(\bz)=0,\,\forall_q$.
Although the examples in Figs.~\ref{fig:ex-nrsi} and~\ref{fig:ex-rsi2} (or others the reader may come up with) serve as mere illustrations of extreme cases, they also make clear that in practice it is not adequate to rely on the angle of these moments to robustly compute shape orientation, since when the magnitude of those moments is small, their phase results very sensitive to the noise.

\begin{figure}[hbt]
\centerline{\includegraphics[width=7cm]{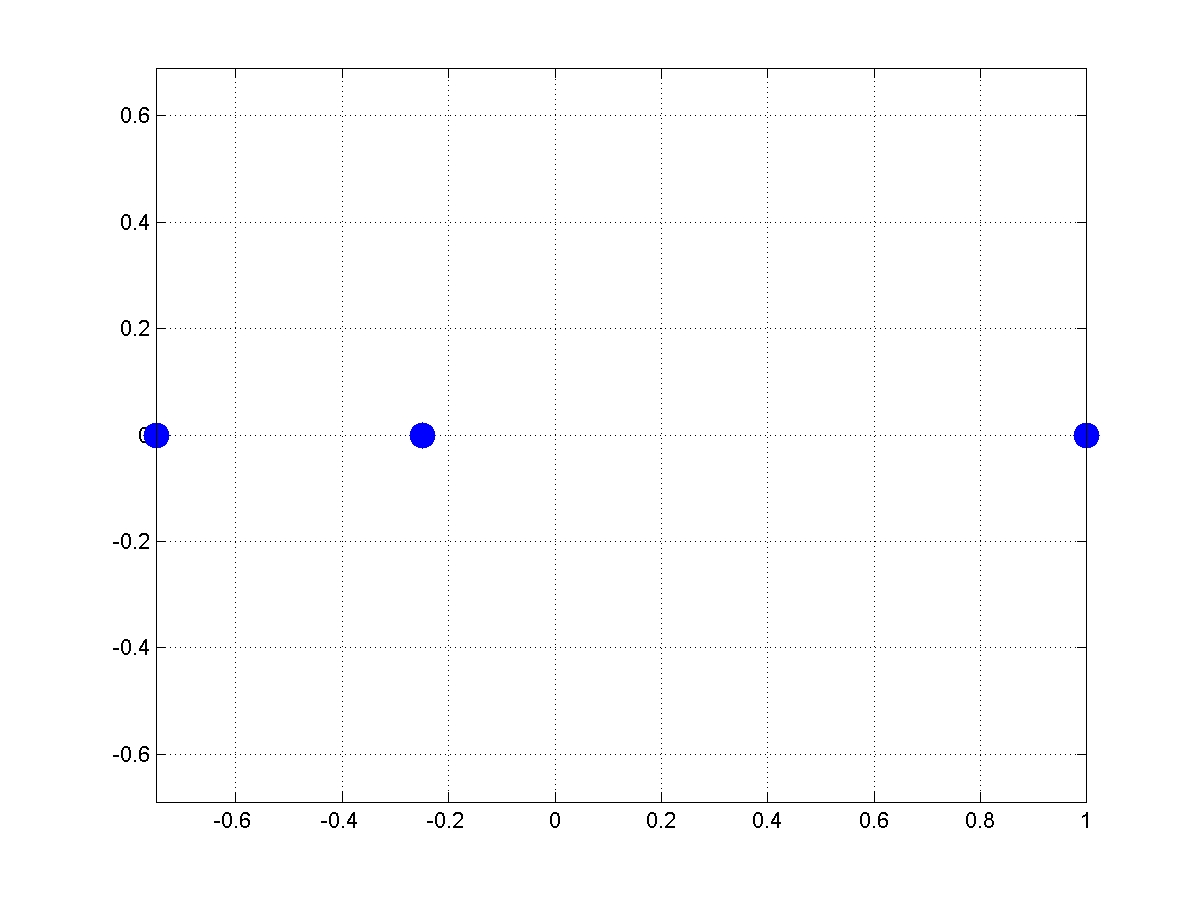}} \caption{Example of a shape
(composed by the points $\left\{1,-1/4,-3/4\right\}$), for which
all the generalized complex moments $GC_{1q}$ are the same as for its reflexion ($\left\{-1,1/4,3/4\right\}$), illustrating that it is impossible to unambiguously define their orientations in terms of those moments.\label{fig:ex-nrsi-gc}}
\end{figure}

Moment-based orientation was later addressed by using Generalized Complex~(GC) moments \citep{shen97,shen99}. GC moments, simply termed rotational moments in a previous review (including Legendre, Zernike, and CMs) by \cite{teh88} are given by \eqref{eq:gci} and can be seen as the coefficients of the Fourier series of radial projections of the image. To deal with ambiguities that arise when attempting to define and compute shape orientation from a single moment of a particular order, \cite{shen97,shen99} use three non-zero GC moments with a fixed index $p$. The method is not simple: from $GC_{pq_1}$ and $GC_{pq_2}$, \emph{the possibility} is inferred that the shape is rotationally symmetric; in case there is that possibility, the unambiguous detection of symmetry requires an exhaustive search; if the shape is classified as rotationally symmetric, a third moment $GC_{pq_3}$ is also used to compute the orientation. The simple example in Fig.\ref{fig:ex-nrsi-gc} shows that this method may fail: consider $\bz_1=[1,-1/4,-3/4]^T$ and $\bz_2$ its reflection, {\it i.e.}, $\bz_1$ rotated by $\pi$, $\bz_2=-\bz_1$, and the choice of GC index $p=1$. Using~\eqref{eq:gcz}, we get
\[
GC_{1q}(\bz_1)=1+\frac{1}{4}e^{jq\pi}+\frac{3}{4}e^{jq\pi}=1+(-1)^q=GC_{1q}(\bz_2)\,,\\
\]
showing that it is impossible to distinguish between the orientations of $\bz_1$ and $\bz_2$ from the moments $GC_{1q}$ (note that the shape in $\bz_1$
and $\bz_2$ is not rotationally symmetric, thus different orientation angles $\theta(\bz_1)$ and $\theta(\bz_2)$ must be computed). Although
\cite{shen97,shen99} propose to tune the index $p$ by maximizing a so-called alternating energy of the radial projection (which also requires exhaustive
search), this method fails to exclude $p=1$ for the example above.

\section{The Unique Orientation of a 2D Shape Through Principal Moment Analysis}
\label{sec:pma}

We now present our algorithm to compute the unique orientation of an arbitrary shape, {\it i.e.}, we derive a function $\theta(\bz)$ that satisfies property
\eqref{eq:equiv_thetac}. Our approach to define and compute the orientation $\theta(\bz)$ is based on the PMs of the shape, given by \eqref{eq:pm}. As we will only deal with their argument, for simplicity, we strip off the scaling factor and work directly with the power sums
\begin{equation}
\mu_k(\bz)=z_1^k+z_2^k+\cdots+z_N^k=\sum_{n=1}^{N}z_n^k\,, \label{eq:pm_simple}
\end{equation}
where $k\in\left\{1,2,3,\ldots\right\}$. Note, nevertheless, that both the power sums $\{\mu_k(\bz)\}$ \eqref{eq:pm_simple} and the PMs $\{M_k(\bz)\}$ \eqref{eq:pm} lead to the same algorithm to be explained, and that both forms can be used to compute the (same) unique orientation of the shape. For this reason, in this section, we refer to $\mu_k(\bz)$ as the $k^{\mbox{\scriptsize th}}$-order PM of the shape.

\subsection{A Single Moment is Not Sufficient}

If we were to choose $\theta(\bz) = \arg \mu_1(\bz)$ (equivalent to applying the simplest form of the method in \cite{mostafa85}, using $C_{10}=\mu_1$), we would satisfy \eqref{eq:equiv_thetac}. In fact, from~\eqref{eq:pm_simple}, $\mu_1(\bz e^{j\phi})=\mu_1(\bz)\, e^{j\phi}$, thus $\arg \mu_1(\bz e^{j\phi})=\arg \mu_1(\bz)+\phi$. Nevertheless, since $\mu_1(\bz)=\sum_n z_n$ is proportional to the shape center, its angle is not a characteristic of the shape format, but only of the shape localization. In practice, to obtain translation invariance, it is common to apply the pre-processing step \eqref{eq:translation_scale_norm}, where the shapes are centered by subtracting their center of mass, thus $\mu_1$ is zero for all shapes and useless to determine a shape orientation.

The choice $\theta(\bz) = \arg \mu_1(\bz)$ is equivalent to imposing the argument of the first-order PM of the rotationally normalized shape \eqref{eq:norm} to
be zero, {\it i.e.}, imposing $\arg \mu_1(\bz e^{-j\, \theta(\bz)})=0$. Our approach is to generalize this method by doing the same to the $k^{\mbox{\scriptsize th}}$-order PM, assumed to be non-zero:
\begin{equation}
\arg \mu_k\left(\bz e^{-j\, \theta(\bz)}\right) = 0\,. \label{eq:pma_k1}
\end{equation}
Note that one non-zero PM must exist, as otherwise all landmarks of the shape are at the origin. To solve for $\theta(\bz)$, use the definition \eqref{eq:pm_simple} to rewrite \eqref{eq:pma_k1} as
\[
\arg \sum_{n=1}^N z_n^k\, e^{-jk\,\theta(\bz)} = 0\,. 
\]
Since complex arguments are defined modulo $2\pi$, we get
\begin{equation}
\arg \sum_{n=1}^N z_n^k  - k\,\theta(\bz) + 2\pi l = 0\,, \label{eq:pma_k3}
\end{equation}
where $l$ is an integer. Now, we express the solution(s) for the normalization angle as
\begin{equation}
\theta(\bz) = \frac{\arg \mu_k(\bz)}{k} + \frac{2\pi}{k} l\,,\qquad
                 l \in \{ 0, 1, \ldots, k-1 \} \,, \label{eq:pma_k}
\end{equation}
where we used \eqref{eq:pm_simple} again and noted that only $k$ values of $l$ lead to distinct solutions for $\theta(\bz)$. Expression \eqref{eq:pma_k} makes clear that an ambiguity arises when attempting to define the normalization angle using the $k^{\mbox{\scriptsize th}}$-order PM alone (for $k\neq 1$): there are $k$ (modulo $2\pi$) different values of $\theta(\bz)$ that annihilate the argument of $\mu_k(\bz e^{-j\, \theta(\bz)})$.

As previously referred, the normalization angle $\theta(\bz)$ needs to satisfy property \eqref{eq:equiv_thetac}, {\it i.e.}, the normalization angle of a rotated shape must be equal to the one of the original shape plus the rotation angle. We derive what this condition imposes to the solution for $\theta(\bz)$ that must
be picked from the set in \eqref{eq:pma_k}. By proceeding in a similar way as in \eqref{eq:pma_k1}--\eqref{eq:pma_k3}, we express the argument of the $k^{\mbox{\scriptsize th}}$-order PM of a rotated shape as
\begin{equation}
\arg \mu_k(\bz e^{j \phi}) = \arg \mu_k(\bz) + k\phi + 2\pi \hat l\,,
\label{eq:pma_k_arg_mkrot}
\end{equation}
where $\hat l$ is an integer that guarantees that the argument of $\mu_k(\bz e^{j \phi})$ falls within the interval where this operator is defined, {\it e.g.},
$(-\pi,\pi]$. Using this result and \eqref{eq:pma_k}, we obtain the normalization angle of the rotated shape:
\begin{equation}
    \theta(\bz e^{j \phi}) = \frac{\arg \mu_k(\bz)}{k} + \phi + \frac{2\pi}{k}(l + \hat l)\,.
    \label{eq:pma_k_rot_pedro}
\end{equation}
We now see that the verification of the desired property \eqref{eq:equiv_thetac} hinges on the choice of the integer $l$ in the definition \eqref{eq:pma_k} of the normalization angle. In fact, if one decides to simply pick a given fixed $l$, {\it i.e.}, to choose the same $l$ for all shape vectors $\bz$, \eqref{eq:pma_k_rot_pedro} becomes $\theta(\bz e^{j \phi})=\theta(\bz) + \phi + (2\pi/k)\hat l$, 
showing that property \eqref{eq:equiv_thetac} is satisfied if and only if $\hat l = 0$ (mod $k$). However, this can not be guaranteed, since in general
equality \eqref{eq:pma_k_arg_mkrot} requires distinct values of $\hat l$ for distinct $\phi$: just imagine $\phi$ ranging from $0$ to $2\pi$ and note that
$\arg \mu_k(\bz e^{j \phi})$ would exhibit jumps (in order to maintain its value within $(-\pi,\pi]$) corresponding to a changing value of $\hat l$, at values of $\phi$ spaced by intervals of length $2\pi/k$.

\subsection{Using a Pair of Moments}

The crux of our approach is to define the normalization angle in \eqref{eq:pma_k} by selecting a value for $l$ that {\it depends on the shape}
vector $\bz$. We will show that it is always possible to select $l(\bz)$ according to our method, {\it i.e.}, that it unambiguously defines a
normalization procedure, and that the resulting normalization angle $\theta(\bz)$ satisfies property \eqref{eq:equiv_thetac}, thus guaranteeing
maximal invariance to rotations in shape-based classification.

To achieve this, we use a supplementary non-zero PM, $\mu_m(\bz)$, with $k$ and $m$ coprime, {\it i.e.}, with $\gcd(k,m) = 1$, where $\gcd$ denotes the greatest common divisor. The case where there do not exist coprime non-zero moments will be treated in the next subsection. Note additionally that there are no shapes (with a finite number of points) with all but one non-zero PM \footnote{A simple way to show this is to use the fact that the first $N$ PMs fully specify a shape with $N$ points \citep{kanatani}. Consider all of them are 0 but $\mu_k=1$, for an arbitrary choice of $k$. From this, compute the shape and from the shape compute the remaining (higher order) PMs (or compute these directly using Newton's identities). It will be clear from the resulting expression that they can not be all $0$.}. Consequently, all shapes (except the one with all landmarks at the origin) have at least two non-zero moments, although not compulsorily coprime.

Our choice for $l(\bz)$ is based on the arguments of the $k^{\mbox{\scriptsize th}}$- and $m^{\mbox{\scriptsize th}}$-order PMs. For simplicity, denote the argument of the $m^{\mbox{\scriptsize th}}$-order PM of
the normalized shape by $\nu(\bz,l)$, where $l \in \{ 0, 1, \ldots, k-1 \}$ stands for the particular integer used in \eqref{eq:pma_k} to compute the
normalization angle. Then,
\begin{align}
    \nu(\bz, l) &= \arg \mu_m \left(\bz e^{-j\, \theta(\bz, l)} \right) \nonumber\\
                   &= \arg \sum_{n=1}^N z_n^m - m \,\theta(\bz, l) \label{eq:pma_d3} \\
                   &= \arg \mu_m(\bz) - \frac{m}{k} \arg \mu_k(\bz) - \frac{m}{k} 2\pi l \,, \label{eq:pma_d4}
\end{align}
where we used the definitions of the PMs \eqref{eq:pm_simple} in \eqref{eq:pma_d3} and of the normalization angle \eqref{eq:pma_k} in \eqref{eq:pma_d4}, emphasizing its dependence on $l$. It is easy to verify that, since $k$ and $m$ are coprime,
the set $\{-(m/k)2\pi l:\, l = 0,1,\ldots,k-1\}$ is the same as the one expressed by $\{(1/k)2\pi l:\, l=0,1,\ldots,k-1\}$ modulo $2\pi$, {\it i.e.}, for each element of the first set, there is an element of the second one that differs by a multiple of $2\pi$ and vice-versa. Thus, the set $\{\nu(\bz,
l):\, l=0,1,\ldots,k-1\}$ contains $k$ different elements (mod $2\pi$) spaced by intervals of length $2\pi/k$. We propose to unambiguously choose
$l(\bz)$ so that $\nu(\bz, l(\bz))$ falls within an arbitrary but \emph{fixed} interval of length $2\pi/k$:
\begin{equation} \label{eq:pma_choice_l}
    \nu\left(\bz, l(\bz)\right) \in I \quad (\mathrm{mod}~2\pi)\,,
\end{equation}
where the interval
\[
    I = \left\{\lambda: \,\lambda_0 < \lambda \leq \lambda_0 + 2\pi/k\right\}\,,
\]
defined by an arbitrary but fixed $\lambda_0 \in \R$, is independent of $\bz$.

The ambiguity in the definition of the normalization angle $\theta(\bz)$ in \eqref{eq:pma_k} is now solved through the choice of $l(\bz)$ in \eqref{eq:pma_choice_l}, but we still have to check that this solution satisfies property \eqref{eq:equiv_thetac}, {\it i.e.}, that the normalization angle of a rotated shape equals the one of the original shape plus the rotation angle. As derived above, the normalization angle of a rotated shape is given by
\begin{equation}
\theta(\bz e^{j \phi})= \frac{ \arg \mu_k(\bz)}{k} + \phi + \frac{2\pi}{k}
\left(l(\bz e^{j \phi})+\hat l\right) \label{eq:pma_v4} \,.
\end{equation}
This last expression is a simple rewrite of \eqref{eq:pma_k_rot_pedro}, now emphasizing the dependance of $l$ on the (rotated) shape; $l(\bz e^{j \phi})$ is thus the solution of \eqref{eq:pma_choice_l} for the rotated shape $\bz e^{j \phi}$. To express the right side of \eqref{eq:pma_v4}
in terms of $\theta(\bz)$, we must relate $l(\bz e^{j \phi})$ to $l(\bz)$. This is done by expressing the argument of the $m^{\mbox{\scriptsize
th}}$ PM of the normalized rotated shape in terms of the one of the original shape:
\begin{eqnarray}
\nu\left(\bz e^{j \phi}, l(\bz e^{j \phi})\right) &=&\nonumber\\&&
\!\!\!\!\!\!\!\!\!\!\!\!\!\!\!\!\!\!\!\!\!\!\!\!\!\!\!\!\!\!\!\!\!\!\!\!\!\!\!\!\!\!\!\!\!\!
=\arg \mu_m(\bz e^{j \phi})- \frac{m}{k} 2\pi l(\bz e^{j \phi})
-\frac{m}{k} \arg \mu_k(\bz e^{j \phi})  \label{eq:pma_v5}\\&&
\!\!\!\!\!\!\!\!\!\!\!\!\!\!\!\!\!\!\!\!\!\!\!\!\!\!\!\!\!\!\!\!\!\!\!\!\!\!\!\!\!\!\!\!\!\!
=\arg \mu_m(\bz) + m \phi-\frac{m}{k} 2\pi l(\bz e^{j \phi})
\nonumber\\&&\;\;\;- \frac{m}{k} \left(\arg \mu_k(\bz)+ k\phi + 2\pi\hat l\right) \label{eq:pma_v6}\\&&
\!\!\!\!\!\!\!\!\!\!\!\!\!\!\!\!\!\!\!\!\!\!\!\!\!\!\!\!\!\!\!\!\!\!\!\!\!\!\!\!\!\!\!\!\!\!
=\arg \mu_m(\bz) - \frac{m}{k} \arg \mu_k(\bz)
 - \frac{m}{k} 2\pi \left(l(\bz e^{j \phi})+\hat l\right) \label{eq:pma_v7}\\&&
\!\!\!\!\!\!\!\!\!\!\!\!\!\!\!\!\!\!\!\!\!\!\!\!\!\!\!\!\!\!\!\!\!\!\!\!\!\!\!\!\!\!\!\!\!\!
= \nu\left(\bz, l(\bz e^{j \phi})+\hat l\right) \,,
\label{eq:pma_v8}
\end{eqnarray}
where \eqref{eq:pma_v5} uses \eqref{eq:pma_d4}, \eqref{eq:pma_v6} uses \eqref{eq:pma_k_arg_mkrot}, \eqref{eq:pma_v7} are simple manipulations, and
\eqref{eq:pma_v8} uses \eqref{eq:pma_d4} again. From \eqref{eq:pma_v8} and our choice \eqref{eq:pma_choice_l} applied to $\bz e^{j \phi}$, we see that $\nu(\bz e^{j \phi}, l(\bz e^{j \phi})) = \nu(\bz, l(\bz e^{j \phi})+\hat l) \in I$. Due to the fact that $I$ is fixed, {\it i.e.}, that it does not depend on the shape $\bz$, and due to the uniqueness (mod $k$) of the solution in \eqref{eq:pma_choice_l} with respect to $l$, $l(\bz e^{j \phi}) + \hat l$ must be equal to $l(\bz)$ (mod $k$). Replacing this concluding equality into \eqref{eq:pma_v4} and using the definition of $\theta(\bz)$ in \eqref{eq:pma_k}, property \eqref{eq:equiv_thetac} is immediately obtained.

We call our method Principal Moment Analysis (PMA).

\subsection{Dealing with Rotational Symmetry}
\label{subsec:rot}

We now deal with the case when there are no coprime $k$ and $m$ such that $\mu_k\neq0$ and $\mu_m\neq0$, in which case it is impossible to apply the
method just described directly.
Let $\gamma$ be the greatest common divisor with respect to all the orders of non-zero PMs, {\it i.e.}, $\gamma = \gcd(K),\; K = \{k \in \mathbb{Z}^+:
\mu_k\neq0\}$. When there are no $k$ and $m$ coprime such that $\mu_k\neq0$ and $\mu_m\neq0$, we have $\gamma>1$. This is equivalent to the existence of a $\gamma$-fold rotational symmetry, {\it i.e.}, that the shape is invariant to rotations of $2\pi/\gamma$. A simple way to derive this equivalence is to use the fact that, as shown in Section~\ref{sec:ansig}, the PMs are the coefficients of the Fourier series of the restriction $h(\bz,\theta)$ in \eqref{eq:cauchy} of the ANSIG to the unit circle: since the rotation of a shape propagates into its ANSIG \citep{rodrigues08a}, $\gamma$-fold rotationally symmetric shapes lead to $h(\bz,\theta)$ with period $2\pi/\gamma$, whose non-zero Fourier series coefficients will only occur at multiples of~$\gamma$; conversely, if the Fourier series coefficients only occur at multiples of $\gamma$, $h(\bz,\theta)$ has period $2\pi/\gamma$ and, as such, using the same propagation property, the shape is $\gamma$-fold rotationally symmetric.
In this case, all normalization angles of the form
\begin{equation}
    \theta = \theta_{0} + \hat k\, 2\pi/\gamma,\; \hat k \in \mathbb{Z} \label{eq:theta_rot_symmetry}
\end{equation}
lead to the same normalized shape. Hence, to compute a normalization angle, it suffices to compute the Fourier series coefficients of the function $h(\bz, \theta/\gamma)$ instead of the ones of $h(\bz, \theta)$ (in the variable $\theta$), then to use these coefficients ({\it i.e.}, the PMs) in the PMA as described in the previous subsection, obtaining an angle $\theta_0'$, and, finally, to invert the expanding effect of $h$ through the contraction of $\theta_{0}'$, {\it i.e.}, assign $\theta_0 = \theta_{0}'/\gamma$. Any $\theta$ in \eqref{eq:theta_rot_symmetry} can then be used.

In terms of the PMs, it is easy to see through the properties of the Fourier series \citep{oppss} that this procedure is simply equivalent to using the PMs of orders $\gamma k$ and $\gamma m$ instead of the original ones of orders $k$ and $m$, respectively, and then contracting the resulting angle. Finally, a last equivalent method is to compute the PMs \eqref{eq:pm} or \eqref{eq:pm_simple} directly from the ``powered'' shape vector $[z_1^\gamma, z_2^\gamma, \ldots, z_N^\gamma]^T$ (equal to the spectrum of $h(\bz, \theta/\gamma)$ up to a real positive scaling factor), apply the PMA and contract the result.

\subsection{Improving Robustness}

Until now, we presented a theoretical proof for the correctness of PMA to unambiguously compute the orientation of arbitrary shapes using a pair of moments. Since in practice it is also important to obtain robustness to noise, we now describe how to improve the robustness of PMA by using a larger set of PMs. In fact, PMA can be used with any pair of coprime indices $(k,m)$, provided that $\mu_k \neq 0$ and $\mu_m\neq 0$. In order to improve robustness, we integrate the contributions of several pairs $(k_1, m_1)$, $(k_2, m_2)$, $\ldots$, $(k_M, m_M)$, by computing pairwise estimates $\theta_{i}(\bz), i=1,2,\ldots,M$, and defining a robust normalization angle $\theta(\bz)$ as the (angular) weighted average of them:
\begin{equation}
    \theta(\bz) = \arg \sum_{i=1}^M w_i \,
    e^{j\,\theta_{i}(\bz)}\,.\label{eq:angav}
\end{equation}
The reason for the angular average is its ability to deal with angles close to the region of circular discontinuity. For instance, we want the average of
$1^\circ$ and $359^\circ$ to be $\arg(\exp(j\,1^\circ)+\exp(j\,359^\circ)) = 0^\circ$, not $(1^\circ+359^\circ)/2 = 180^\circ$. The proof that property \eqref{eq:equiv_thetac} holds for the robust normalization angle defined in \eqref{eq:angav} is straightforward:
\begin{align}
    \theta(\bz e^{j \phi}) &=
    \arg \sum_{i=1}^M w_i \, e^{j\,\theta_{i}(\bz e^{j\phi})} \nonumber\\
    &= \arg \sum_{i=1}^M w_i \, e^{j(\theta_{i}(\bz) + \phi)} \label{eq:ppp}\\
    &= \arg \sum_{i=1}^M w_i \, e^{j\,\theta_{i}(\bz)} + \phi \nonumber\\
    &= \theta(\bz) + \phi\,,\label{eq:ooo}
\end{align}
where \eqref{eq:ppp} uses the fact that the individual $\theta_{i}(\bz)$ satisfy \eqref{eq:equiv_thetac} and \eqref{eq:ooo} uses definition \eqref{eq:angav}.

\subsection{Implementation Details}

In practice, PMA was implemented the following way: we start by computing the first $20$ power sums in \eqref{eq:pm_simple} and normalizing their magnitude by dividing $\mu_k$ by its absolute value raised to $k$, {\it i.e.}, by using ${\hat \mu}_k = \mu_k/|\mu_k|^k$ instead of $\mu_k$. This normalization aims to cancel the growth of the magnitude of the power sums with $k$. Note that alternatives to this normalization include the one discussed in Appendix~\ref{app:norm} or the usage of the PMs as defined~\eqref{eq:pm}. Then, we detect non-zero PMs by thresholding their magnitude, {\it i.e.}, we only use ${\hat \mu}_k$ if $|{\hat \mu}_k|$ is above a threshold, say,
$10^{-3}$.
In what respects to experimental tuning, our method only requires dealing with the two parameters just referred.
Finally, we compute the greatest common divisor of all the indices corresponding to non-zero PMs and run the algorithm described in the previous section.

In the second step of the algorithm (the core of the method), for each non-zero moment ${\hat \mu}_k$, we find the smallest index $m$ that is coprime with $k$ (the
first step described above guarantees that this index always exists).
Then, we search for $l(\bz)\!\in\!\{0,1,2,\ldots,k\!-\!1\}$ until we find $\nu(\bz, l(\bz))\!\!\mod 2\pi\!\in\! I$ (we choose $I\!= \!(-\pi/k, \pi/k]$). The value
found for $l(\bz)$ is used to compute the pairwise estimates $\theta_i(\bz)$, which are then averaged using weights given by $w_i \!=\!|{\hat \mu}_{k_i}\,{\hat \mu}_{m_i}|$. The rationale for the choice of these weights is that PMs with larger magnitude have an argument less sensitive to noise. Note that the PMs used in the weights are the normalized ones, being this normalization thus also important to avoid over-weighting angles $\theta_{i}(\bz)$ computed from PMs of large order.

\section{Extension to Grey-level Images}
\label{sec:grey}

The algorithm presented in the previous section computes a unique orientation $\theta(\bz)$ for an arbitrary set of landmarks $\{z_1, z_2, \ldots, z_N\}$, satisfying property \eqref{eq:equiv_thetac}. We now generalize the concept to compute a unique orientation of a continuous image $g(x, y)$. For that purpose, generalize the moments of \eqref{eq:pm_simple} to the equivalent ones of \cite{mostafa85}, {\it i.e.}, make $\mu_k(g) = C_{k0}(g)$. From \eqref{eq:cms}, the PMs of the grey-level image $g(x,y)$ are thus
\begin{equation}
\mu_k(g)=\int\!\int_{-\infty}^{+\infty} \left(x+jy\right)^k g(x,y) \,dx\,dy\,,\label{eq:pms_image}
\end{equation}
with $k \in \{0, 1, 2, \ldots\}$. In what respects to representation, the generalization to continuous images loses completeness: in opposition to the case of a set of points, discussed in Sections \ref{sec:pms} and \ref{sec:ansig}, the PMs in \eqref{eq:pms_image} do not determine the image $g(x,y)$ univocally. An immediate way to conclude that is focusing on radial images. Start by rewriting \eqref{eq:pms_image} in polar coordinates:
\begin{equation}
\mu_k(g)=\int_{-\pi}^{\pi}\int_{0}^{\infty} r^k e^{jk\theta} g(r\cos \theta, r\sin \theta)\, r\,dr\,d\theta\,.\label{eq:pms_polar}
\end{equation}
Now, for the radial image $g(r\cos \theta , r\sin \theta) = R(r)$, from \eqref{eq:pms_polar}, we easily get $\mu_k(g)=0$, for $k \in \{1, 2, 3, \ldots\}$, and $\mu_0(g)=2\pi \int_{0}^{\infty} R(r)\,r\,dr$.
As this integral does not define the function $R(r)$ univocally \footnote{For example, the functions $R_1(r) = H(r) - H(r-\sqrt{2/3})$ and $R_2(r)= r(H(r) - H(r-1))$, where $H(\cdot)$ denotes the Heaviside step function, lead to the same value for the moment $\mu_0=2\pi/3$.}, the PMs do not determine $g(x,y)$.

Naturally, the lack of completeness just referred does not impede the extension of PMA, our rotational normalization algorithm presented in Section \ref{sec:pma}, for continuous images. A simple way to derive this extension is by using the derivations in Sections~\ref{sec:mri} and \ref{sec:pma} with new definitions of the objects at hand, {\it i.e.}, with moments and rotations of images instead of shapes. The symbol $\bz$ is now interpreted as an image $g(x,y)$ and $\bz e^{j\phi}$ is interpreted as the image that results from the (counterclockwise) rotation of $\bz$ by the angle $\phi$.
With these definitions, its trivial to show that it still holds $[\bz e^{j\phi}] e^{j\psi} = \bz e^{j(\phi+\psi)} = [ \bz e^{j\psi} ] e^{j\phi}$ (because image rotation is associative and commutative), thus the derivations in Section~\ref{sec:mri} remain valid for the interpretation in terms of continuous images.

As far as the derivation of the PMA algorithm in Section \ref{sec:pma} is concerned, the reader should note that it is entirely based on the property $\mu_k(\bz e^{j\phi}) = \mu_k(\bz) e^{j k \phi}$. We have to show that this property extends to the interpretation in terms of grey-level images, {\it i.e.}, that the $k^{\mbox{\scriptsize th}}$-order PM of an image rotated by $\phi$ equals to the product of the $k^{\mbox{\scriptsize th}}$-order PM of the original image by $e^{j k \phi}$ (notice that the multiplication of $\mu_k$ by $e^{j k \phi}$ is an ordinary one, not a rotation, since $\mu_k$ is a complex number, not an image). Using the definition of PMs in \eqref{eq:pms_polar}, the desired property is immediate:
\begin{align}
   \mu_k(\bz e^{j\phi}) \!&=\!\int_{\!-\!\pi}^{\pi}\!\!\int_{0}^{\infty}\!\!\! r^k e^{jk\theta} g(r\!\cos (\theta\!\!-\!\!\phi), r\!\sin (\theta\!\!-\!\!\phi))\, r\,dr\,d\theta\nonumber\\
    &= \int_{\!-\!\pi}^{\pi}\int_{0}^{\infty} r^k e^{jk(\theta+\phi)} g(r\cos \theta, r\sin \theta)\, r\,dr\,d\theta\nonumber\\
    &= e^{jk\phi} \int_{\!-\!\pi}^{\pi}\int_{0}^{\infty} r^k e^{jk\theta} g(r\cos \theta, r\sin \theta)\, r\,dr\,d\theta\nonumber\\
    &=e^{jk\phi}\mu_k(\bz)\,.\nonumber
\end{align}

Having shown how to extend PMA to grey-level images, we end this section by emphasizing that some care must be taken with the claims about the evidence of rotational symmetry and the universality of the algorithm. When dealing with sets of landmarks, as derived in Section~\ref{sec:pma}, the moments $\mu_k$ are nonzero only for indices $k$ multiples of $\gamma>1$ if and only if the shape is $\gamma$-fold rotationally symmetric. However, this equivalence is not fulfilled for the case of continuous grey-level images. To get insight on what happens in this case, start by interpreting the moments in \eqref{eq:pms_polar} in terms of the Fourier series of periodic signals obtained by circularly slicing the image. In particular, the $k^{\mbox{\scriptsize th}}$-order PM of the polar coordinate image $f(r,\theta) \;\stackrel{\mathrm{def}}{=}\; g(r\cos \theta , r\sin \theta)$ can be written as
\begin{equation}
\mu_k=\int_{0}^{\infty} r^{k+1}F(r,k)\,dr\,,\label{eq:pmsf}
\end{equation}
where
\begin{equation}
F(r,k)=\int_{-\pi}^{\pi} f(r,\theta) e^{jk\theta} \,d\theta \label{eq:sfgrey}
\end{equation}
are the coefficients of the Fourier series of the $2\pi$-periodic (in $\theta$) signal $f(r,\theta)$.
Naturally, an image $f(r,\theta)$ is $\gamma$-fold rotationally symmetric if and only if it is $2\pi/\gamma$-periodic in $\theta$, for all $r>0$. Hence, the coefficients $F(r,k)$ of its Fourier series are nonzero only for $k$ multiple of $\gamma$ if and only if there is $\gamma$-fold symmetry in $f(r,\theta)$ \citep[see, {\it e.g.},][]{oppss}. In turn, the nonzero moments $\mu_k$ in \eqref{eq:pmsf} only occur for $k$ multiple of $\gamma$ in this case and PMA can be used as described in Subsection~\ref{subsec:rot}. However, this last statement is not an equivalence, since nonzero coefficients $F(r,k)$ may be destroyed by the integration in \eqref{eq:pmsf}. In fact, as we detail in Appendix~\ref{app:exc}, there exist very particular grey-level images that are not rotationally symmetric but have moments $\mu_k$ that are nonzero only for indices $k$ multiples of $\gamma>1$. There even exist images with a single non-zero moment (other than the radial ones referred in the first paragraph of this section; these last ones are not normalizable in what respects to orientation). Naturally, as any other method based on these moments, PMA fails to process these images.

\section{Experiments}
\label{sec:exp}

We now describe experiments. The following subsections focus on illustrating the compactness of the PM-based shape representation (Subsection \ref{subsec:exppms}), its usage in classification (Subsection \ref{subsec:exppmsclass}), and the results of PMA for rotational normalization of shapes (Subsection \ref{subsec:exppma}) and grey-level images (Subsection \ref{subsec:exppmagrey}).

\subsection{PMs for Shape Representation}
\label{subsec:exppms}

\begin{figure}[htb]
\centerline{\includegraphics[width=7cm]{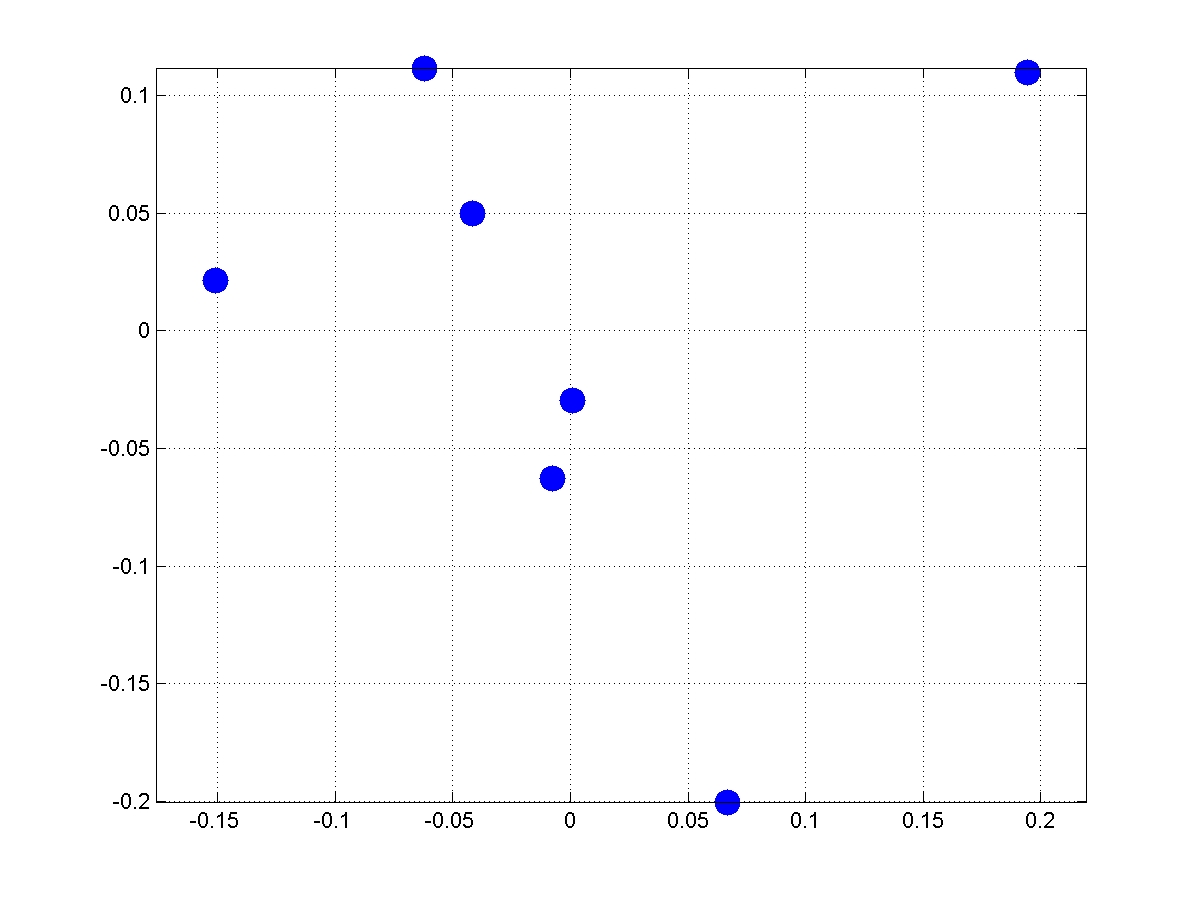}}
        \caption{A 2D shape described by a small number of landmarks.}
        \label{fig:ill1}
\end{figure}

\begin{figure}[htb]
\centerline{\includegraphics[width=8.5cm]{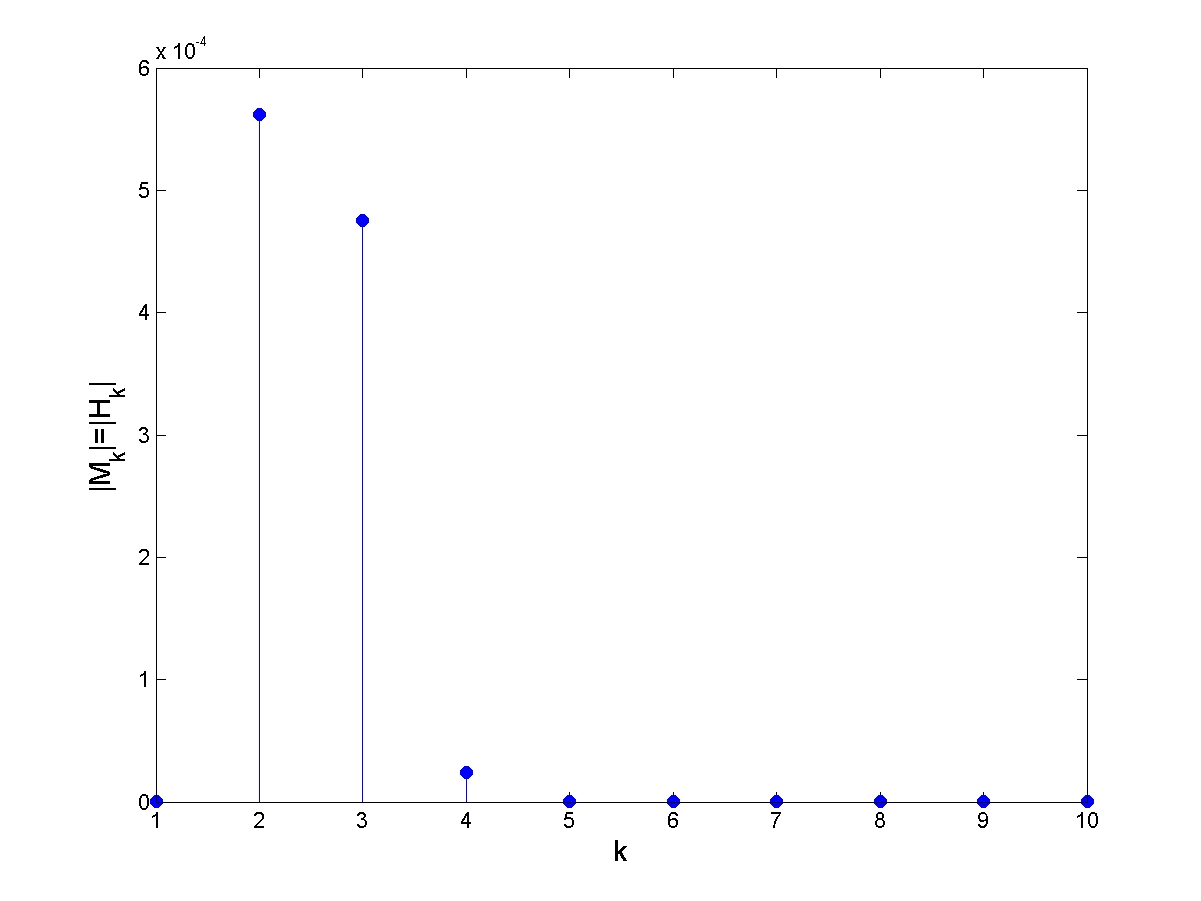}}
        \caption{Magnitude of the PMs of the shape in Fig.~\ref{fig:ill1} (or of the coefficients of the Fourier series of its analytic signature).}
        \label{fig:ill1a}
\end{figure}

In this subsection, we focus on showing that the computational saving that arises from using our PMs does not degrade performance when compared with the discriminative ANSIG, the densely sampled signature introduced by \cite{rodrigues08a}. We illustrate this point with the shape shown in Fig.~\ref{fig:ill1}, which is described by $7$ landmarks.
The plot in Fig.~\ref{fig:ill1a} shows the magnitude of the PMs $\{M_k\}$ of this shape.
Proceeding as described in Section~\ref{sec:ansig}, we obtain the required number of PMs for this shape, $k_B=6$.
As easily perceived from Fig.~\ref{fig:ill1a}, the magnitude of the $6^{\scriptsize\mbox{th}}$ PM is very small, indicating that the first $6$, $\left\{M_k, 0\leq k \leq 5\right\}$, containing the majority of the energy,
adequately describe the shape. Since the shape was pre-processed as in \eqref{eq:translation_scale_norm}, we obtain $M_1=0$ and $M_0=1$ (this last PM is not represented in the plot).

\begin{figure}[htb]
\centerline{\includegraphics[width=8.5cm]{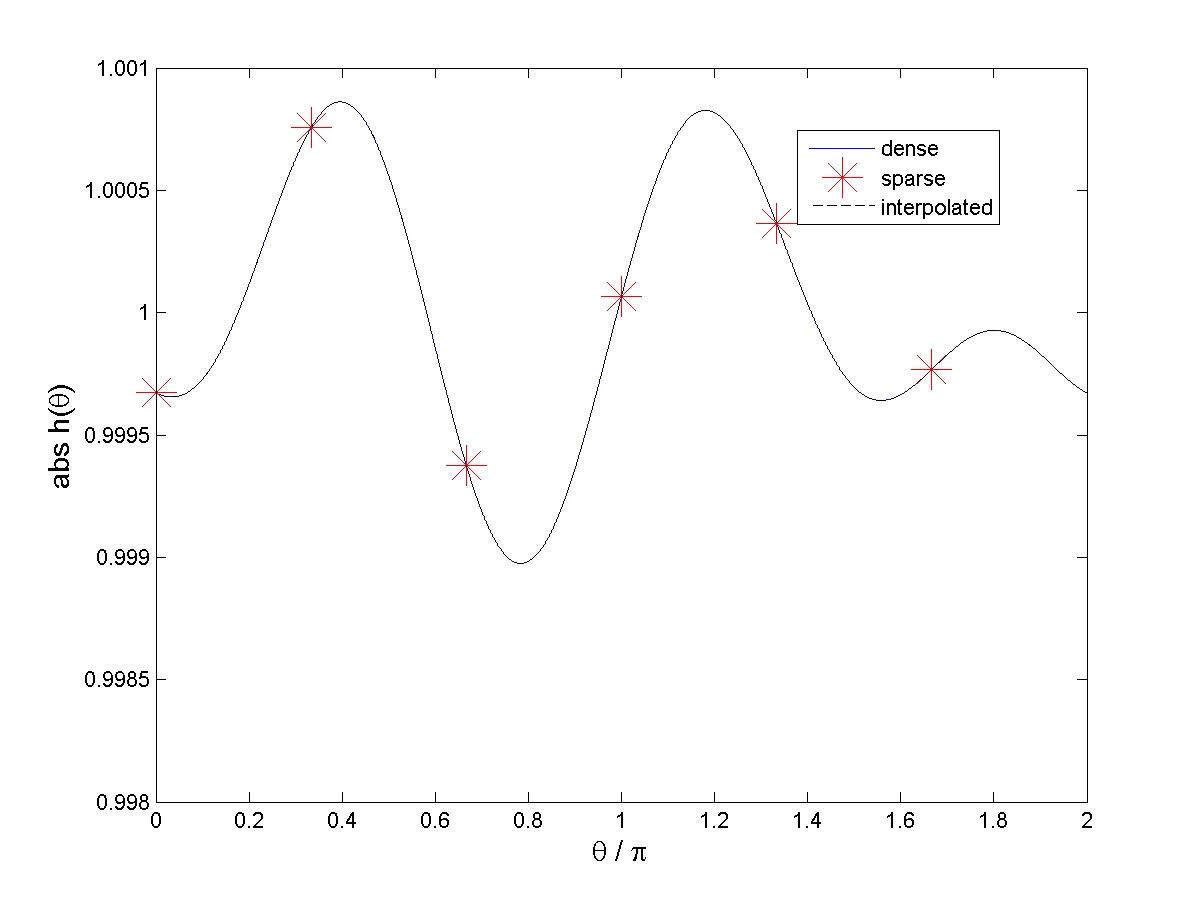}}
\centerline{\includegraphics[width=8.5cm]{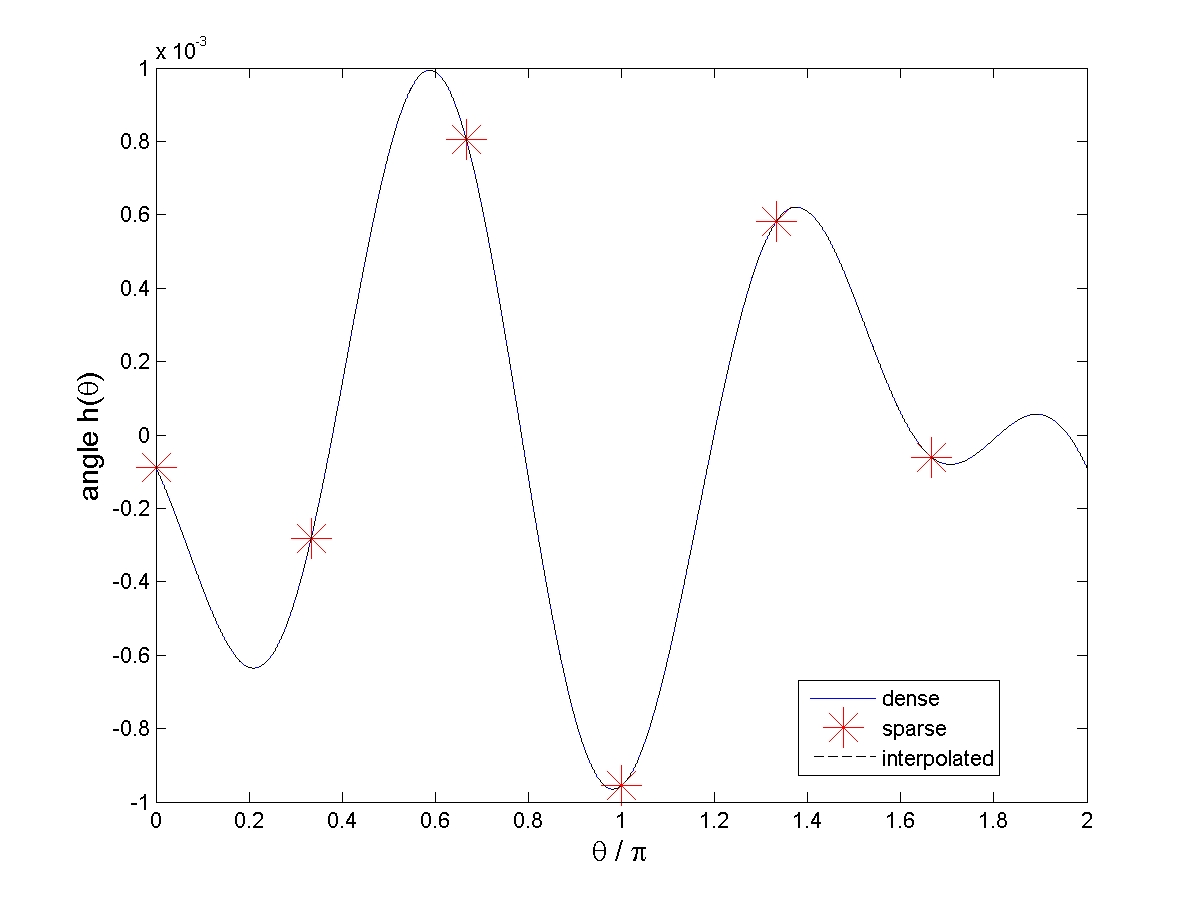}}
\caption{Magnitude and phase of the analytic signature of the shape in Fig.~\ref{fig:ill1}. The very sparse
representation we propose leads to plots that result visually indistinguishable
from those obtained by dense sampling (after reconstruction).}
        \label{fig:ill2}
\end{figure}

In Fig.~\ref{fig:ill2}, we represent, with solid lines, the magnitude and phase of the densely sampled ANSIG of the shape in Fig.~\ref{fig:ill1}. As shown in Section~\ref{sec:ansig}, the coefficients $\{H_k\}$ of the Fourier series of this periodic complex signal are given by the PMs of the shape. To verify this in practice, we computed the {\it Fast Fourier Transform} (FFT) of the vector collecting the dense sampling of one period of the ANSIG, since it is straightforward to derive that
this FFT is equal to $\{H_k\}$ multiplied by the number of samples \citep{oppss,opp}. As expected, we concluded that the Fourier series coefficients $\{H_k\}$ coincide with the PMs $\{M_k\}$, whose magnitude is represented in Fig.~\ref{fig:ill1a}.

\begin{figure}[htb]
\centerline{\includegraphics[width=8.5cm]{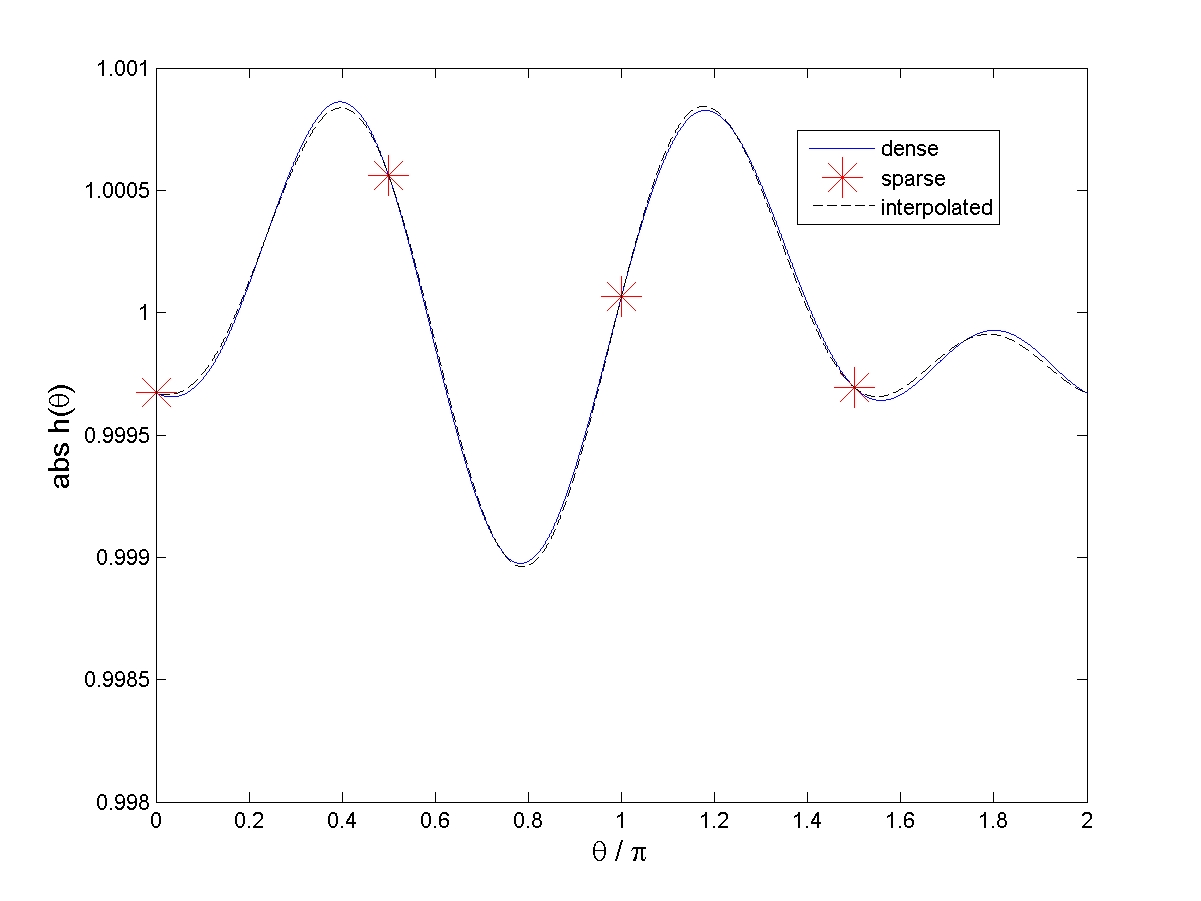}}
\centerline{\includegraphics[width=8.5cm]{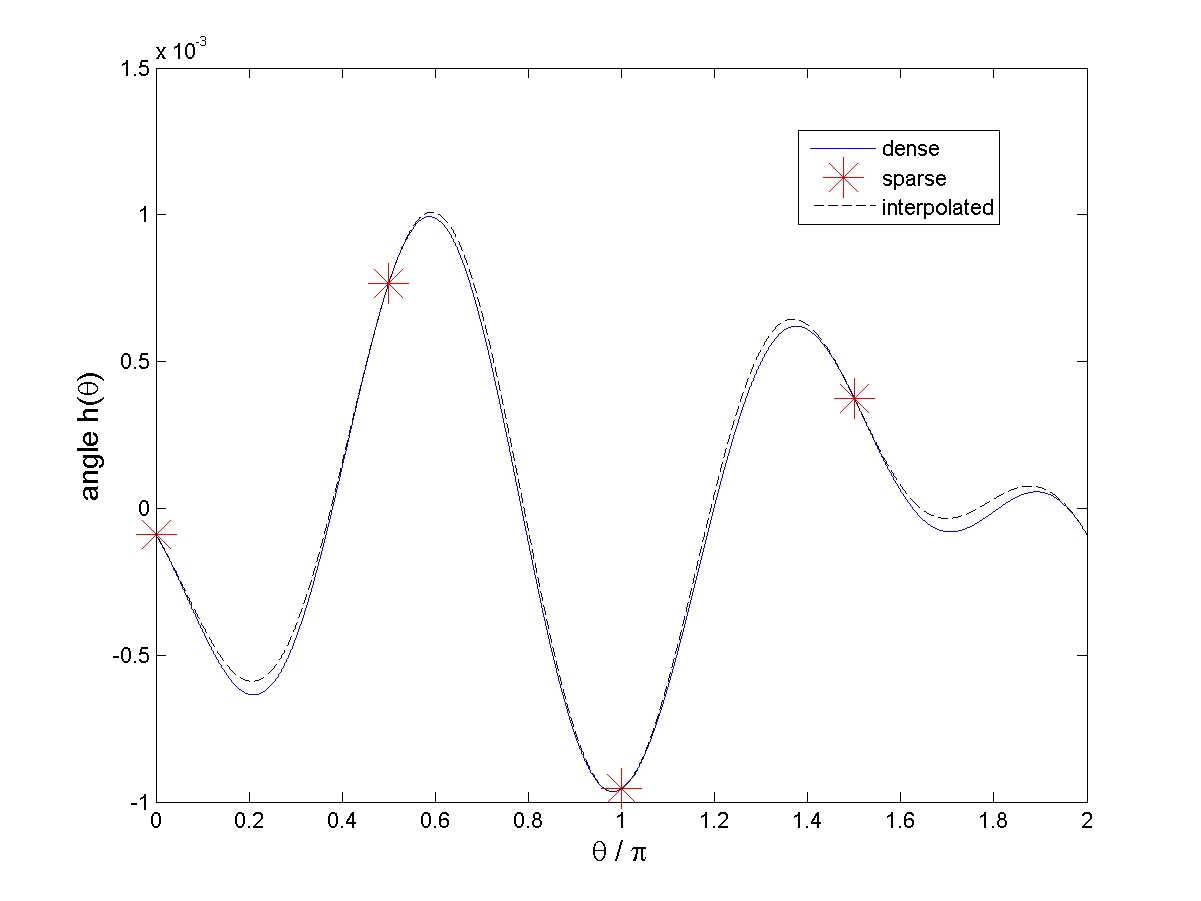}}
\caption{Same as in Fig.~\ref{fig:ill2}, now with a number of samples smaller than the one required by our study. Note how the interpolated signature now differs from the dense sampled one.}
        \label{fig:ill2less}
\end{figure}

In the plots of Fig.~\ref{fig:ill2}, we also compare the densely sampled ANSIG with a signature obtained by interpolating our very compact representation. As derived in Section~\ref{sec:ansig}, the required number $k_B=6$ of coefficients needed to represent the shape can be interpreted either as the minimum number of PMs or the minimum number of samples of the ANSIG. In Fig.~\ref{fig:ill2}, we represent these $k_B=6$ samples with stars, showing how much sparser this representation is when compared with the densely sampled ANSIG. Finally, we superimpose, represented by dashed lines, the reconstruction obtained by interpolating this sparse set as described in Section~\ref{sec:ansig}. We see that the lines of both plots are visually indistinguishable, showing that the PMs are adequate to represent the ANSIG and, consequently, the underlying shape. To illustrate what happens when using less samples than the minimum required by our study, we repeat the procedure by interpolating from $4$ samples, obtaining Fig.~\ref{fig:ill2less}. As easily seen, the reconstructed signature differs from the dense sampled one. Note, nevertheless, that there is no guarantee that our bound $k_B$ is tight.

\subsection{PMs for Shape Classification}
\label{subsec:exppmsclass}

In our experiments, the behavior illustrated in the previous subsection was observed in general, {\it i.e.}, a set of $k_B$ PMs always suffices to accurately describe the densely sampled ANSIG. Although this is enough to guarantee that the same results are obtained when classifying shapes described by either their small sets of PMs or their dense ANSIGs, we also verified this directly. In particular, we generated noisy versions of
prototype shapes 
and classified them by using 1-NN, {\it i.e.}, by just selecting the prototype
that had most similar description. The number of PMs $k_B$ ranged from $15$ to
$22$, thus our descriptions are much shorter than the vectors of $512$ ANSIG samples
used in~\cite{rodrigues08a}. We performed hundreds of tests for each shape,
obtaining the same performance ($100\%$ correct classifications,
except for shapes that are visually indistinguishable), for both the densely
sampled ANSIG and the PMs. Since the ANSIG was extensively demonstrated in shape-based classification of
real images \citep{rodrigues08a,rodrigues08b} and we have shown that the PMs have similar behavior, we do not report here other experiments on PM-based shape classification.

\subsection{PMA for Shape Normalization}
\label{subsec:exppma}

Although in the paper we have theoretically proven the correctness of PMA for shape normalization, {\it i.e.}, that it succeeds in unambiguously computing a unique orientation for any shape, the impact of PMA in shape-based recognition
applications is also determined by the sensitivity to the noise, since
observations of similar shapes must originate similar normalization angles. In
this subsection, we illustrate that PMA is able to deal with these situations.

\begin{figure}[htb]
\centerline{
\includegraphics[width=4.5cm]{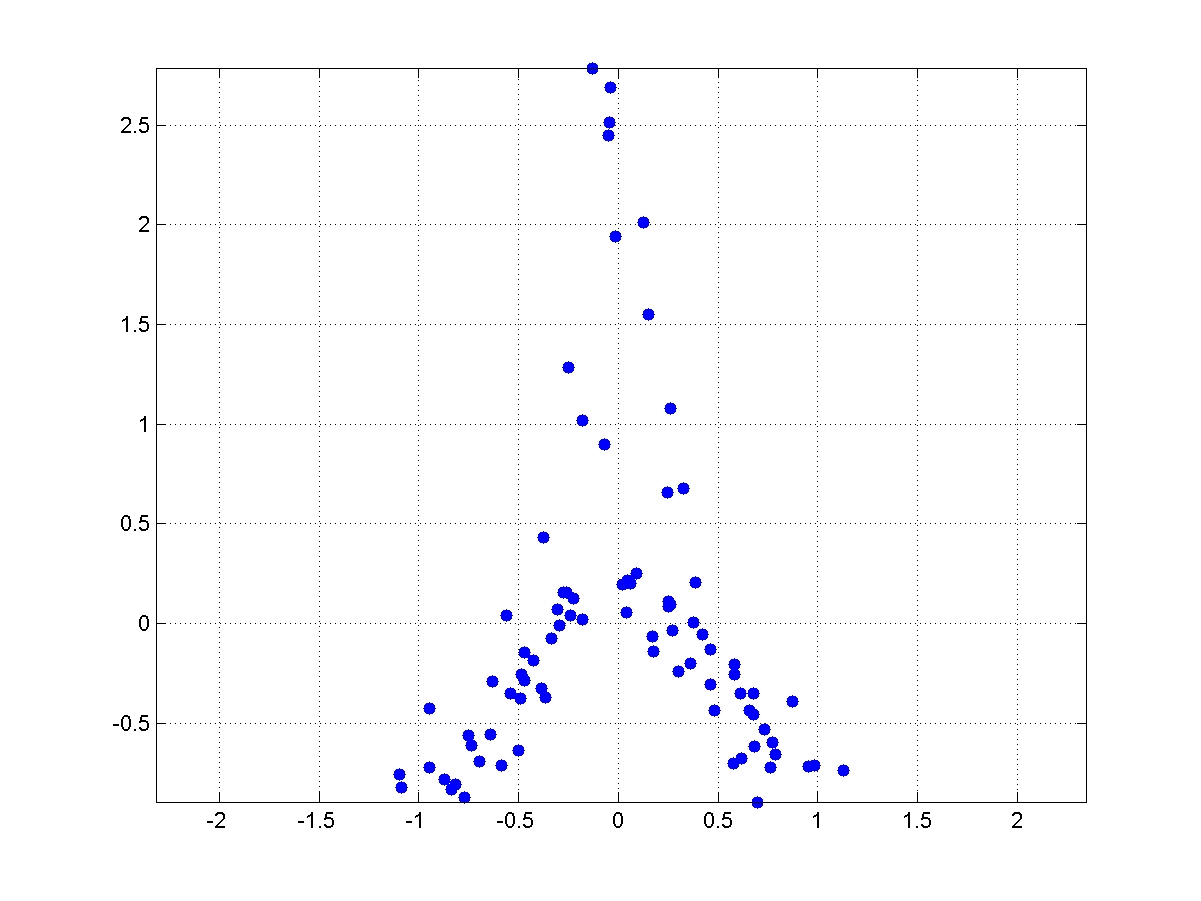}
\includegraphics[width=4.5cm]{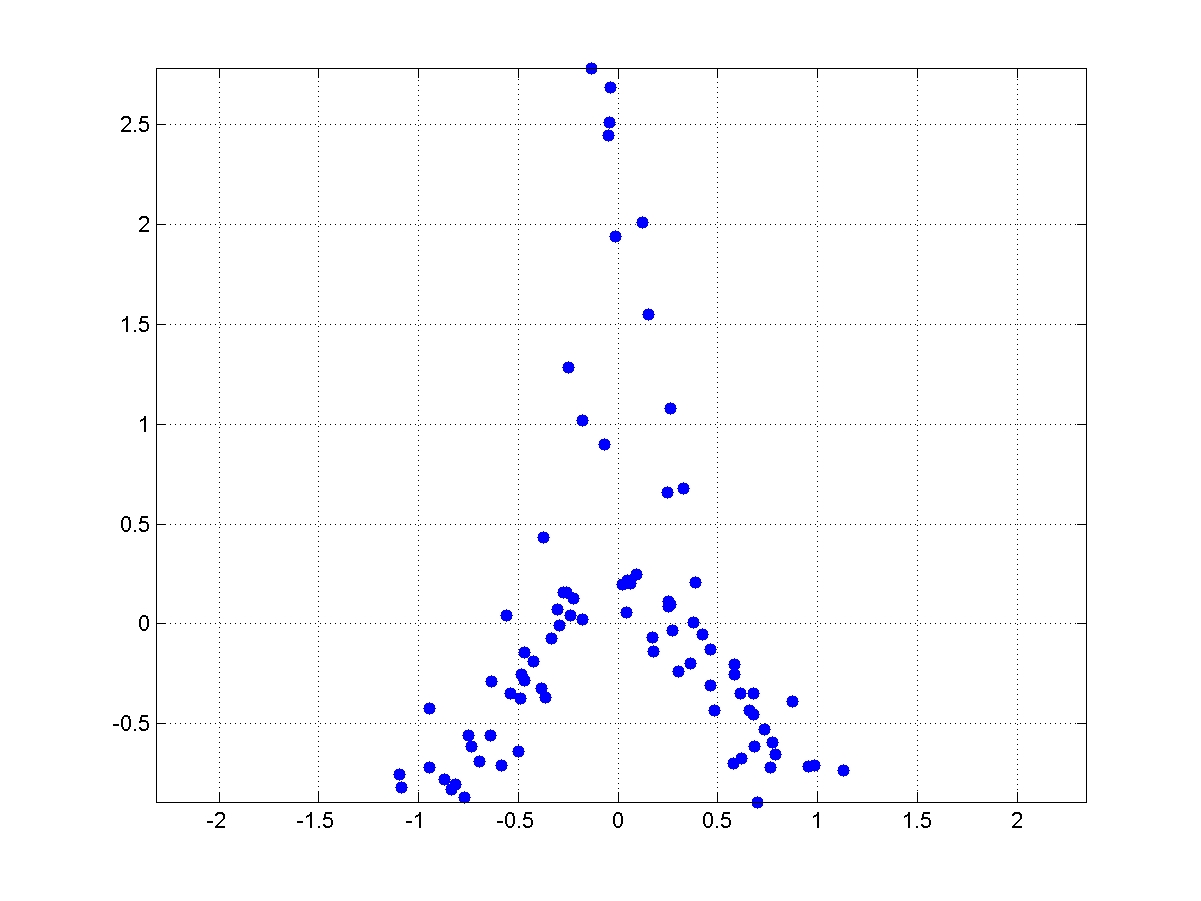}}
\centerline{
\includegraphics[width=4.5cm]{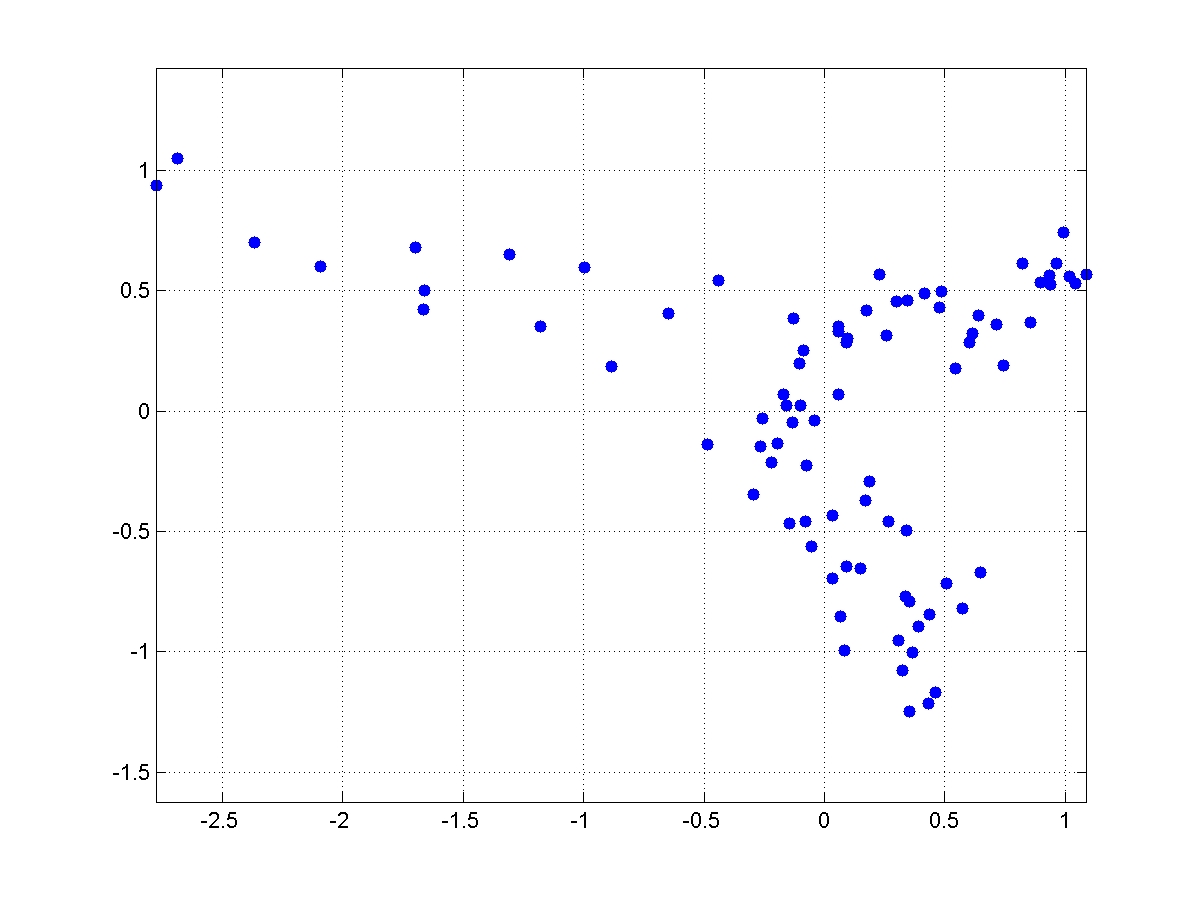}
\includegraphics[width=4.5cm]{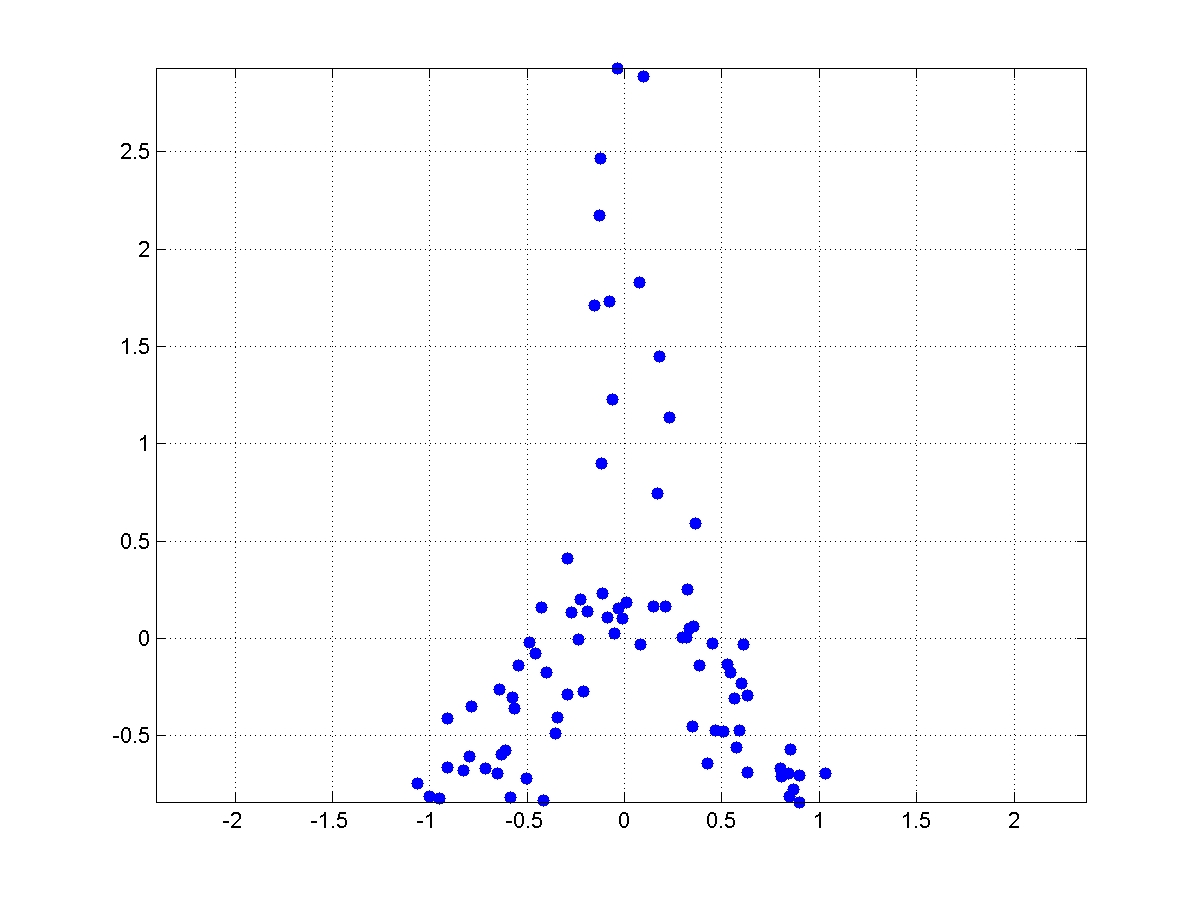}}
\centerline{
\includegraphics[width=4.5cm]{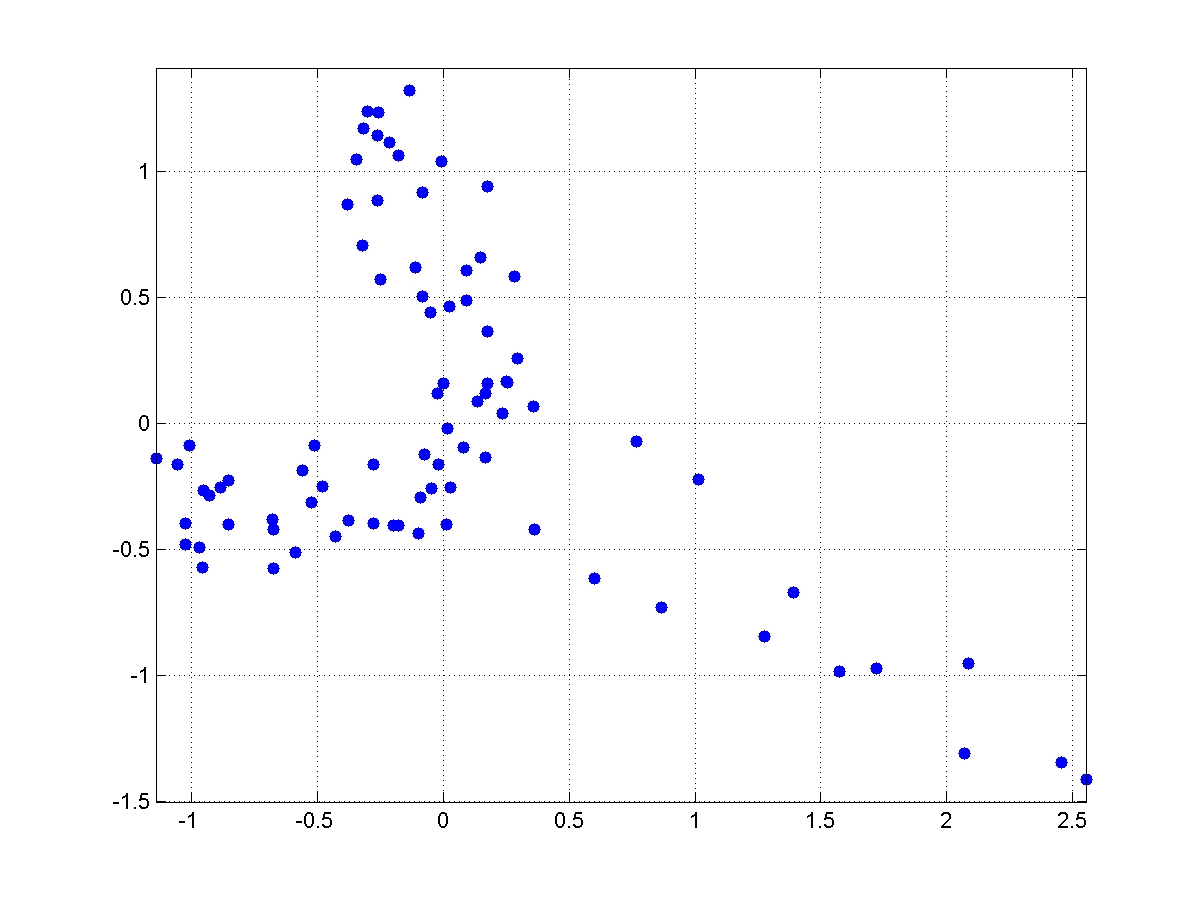}
\includegraphics[width=4.5cm]{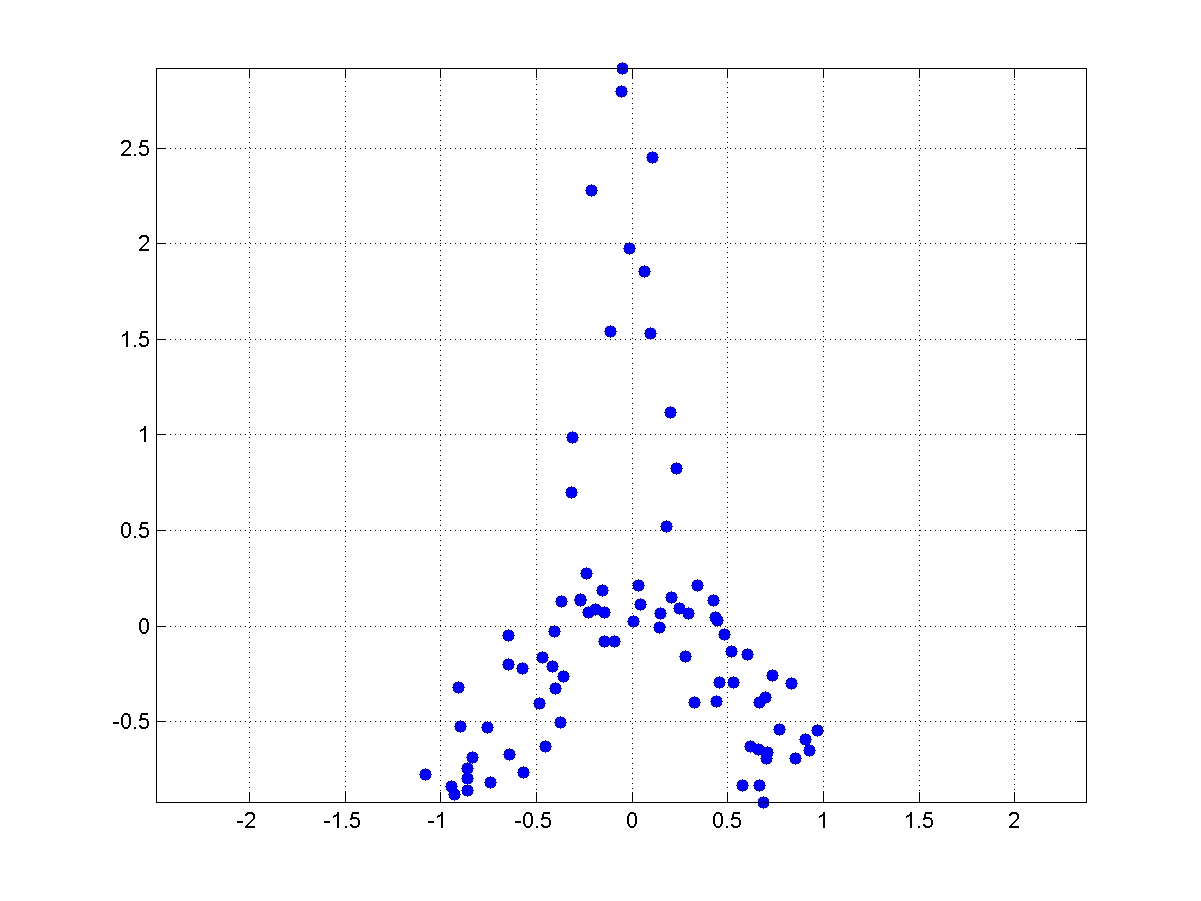}}
\centerline{
\includegraphics[width=4.5cm]{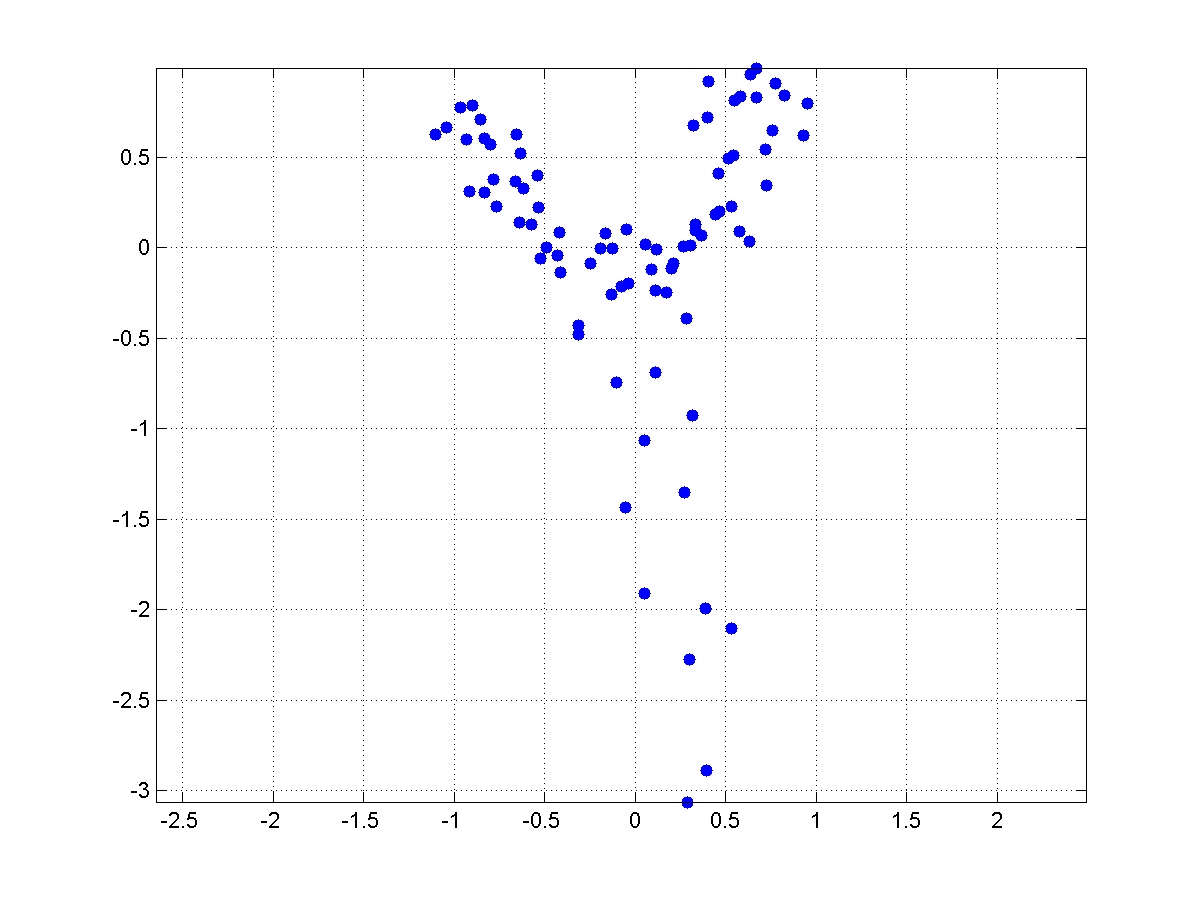}
\includegraphics[width=4.5cm]{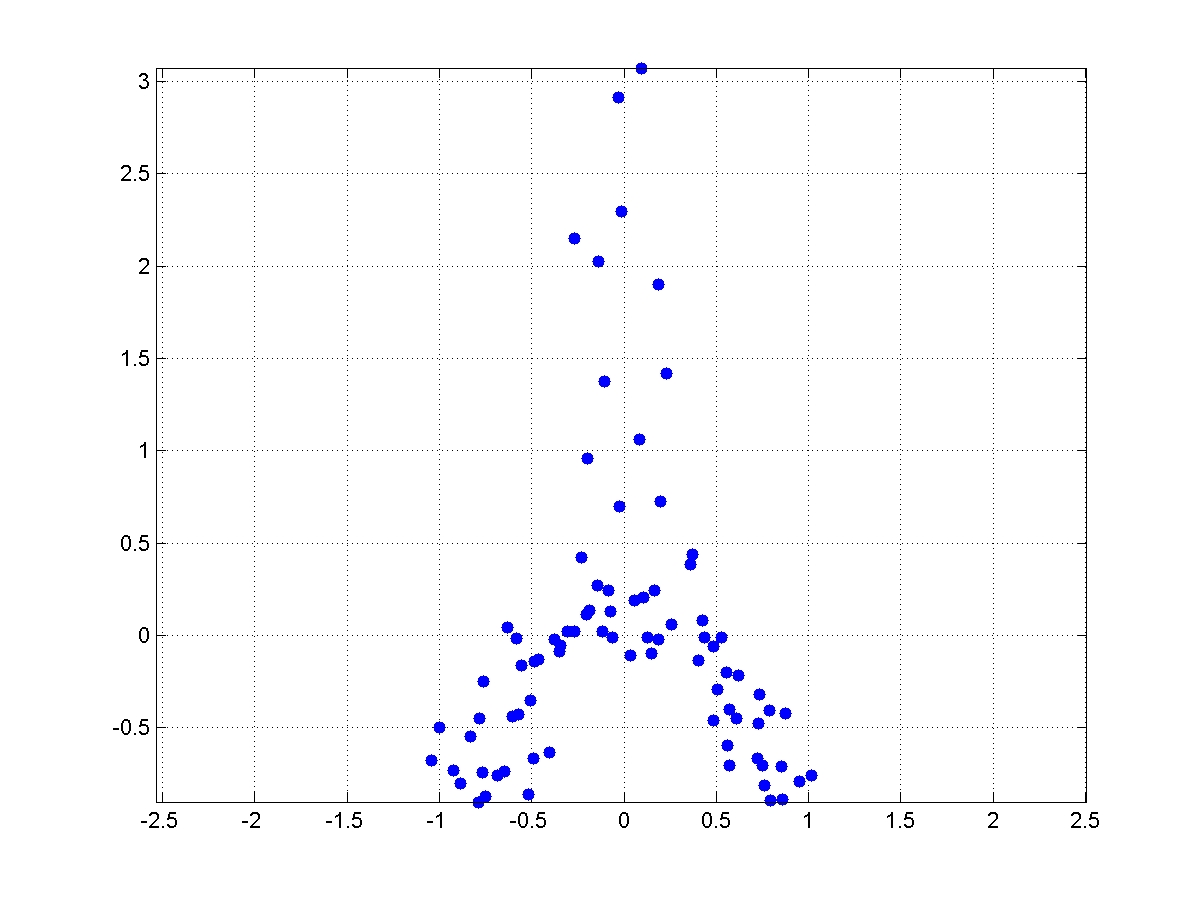}}
\caption{Examples of using PMA for computing the orientation of general shapes.
Left: original shapes; right: PMA oriented shapes. Relative to the ambiguity
illustrated in Fig.~\ref{fig:ambill}, note that PMA disambiguates the direction
of the principal axis.\label{fig:ambex}}
\end{figure}

We start by illustrating that our method disambiguates the direction of the
principal axis of generic shapes, {\it i.e.}, shapes without rotational
symmetry. We used noisy versions of the shape in Fig.~\ref{fig:ambill}, see
examples on the left column of Fig.~\ref{fig:ambex}, and computed the
correspondent normalization angles using PMA. The right column of
Fig.~\ref{fig:ambex} shows the resultant shapes, {\it i.e.}, the rotationally
normalized versions of the corresponding shapes on the left. We see that,
regardless of the noise, all the normalized shapes exhibit similar orientation.

\begin{figure}[htb]
\centerline{\includegraphics[width=9.5cm]{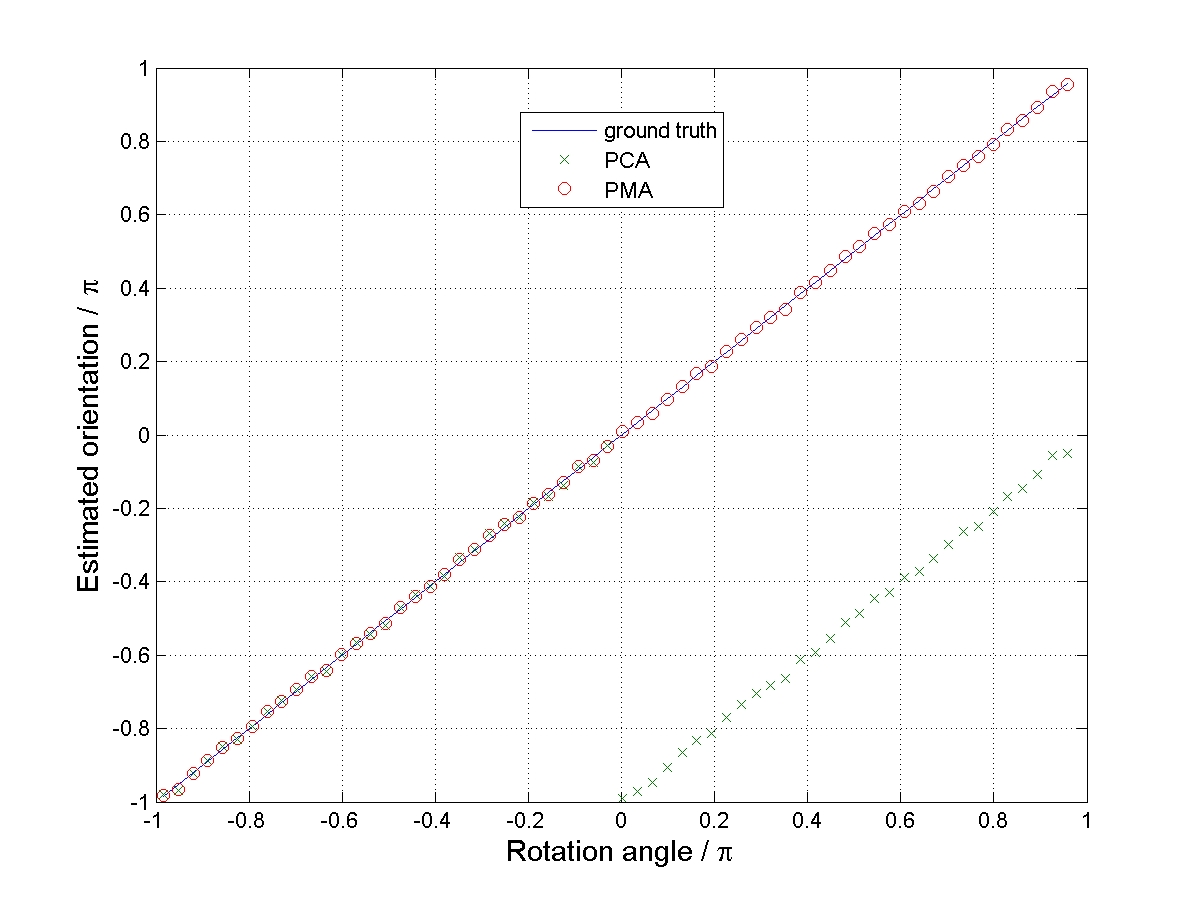}} \caption{Comparison of PMA
and PCA for general shapes.\label{fig:ambplot}}
\end{figure}

Although the examples in Fig.~\ref{fig:ambex} illustrate the disambiguation of
direction, the accuracy of the estimates of the normalization angle is better
evaluated by contrasting them with the ground truth. We thus performed
experiments by rotating noisy shapes according to a known angle, ranging from
$-\pi$ to $\pi$, and then estimating the orientation. The plot in
Fig.~\ref{fig:ambplot} summarizes the results. We see that the estimates
obtained by using PMA are very close to the correspondent true orientation,
for all values of the rotation angle. For illustration purposes, the plot also
shows the results obtained by using PCA, which naturally exhibit the
directional ambiguity discussed in Section~\ref{sec:int} and illustrated in
Fig.~\ref{fig:ambill}.

\begin{figure}[htb]
\centerline{
\includegraphics[width=4.5cm]{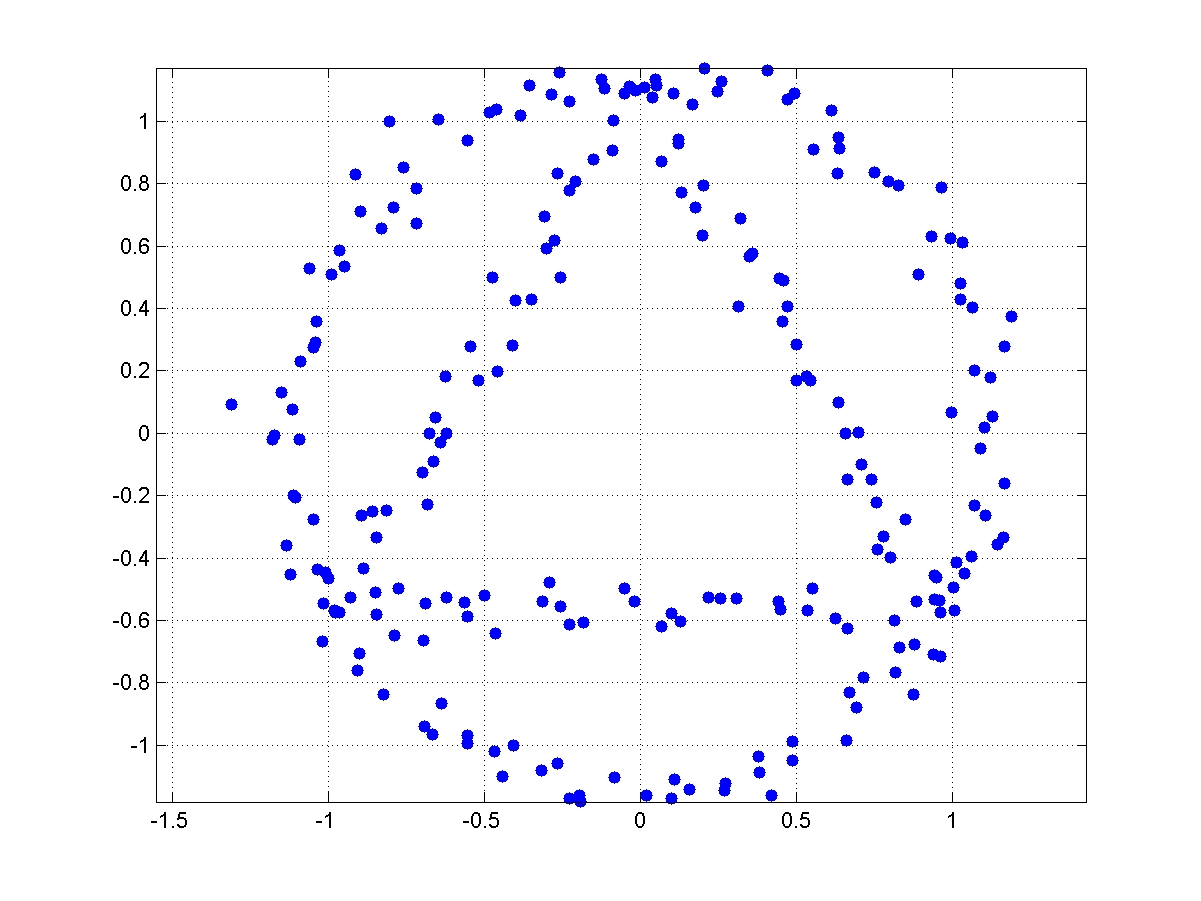}
\includegraphics[width=4.5cm]{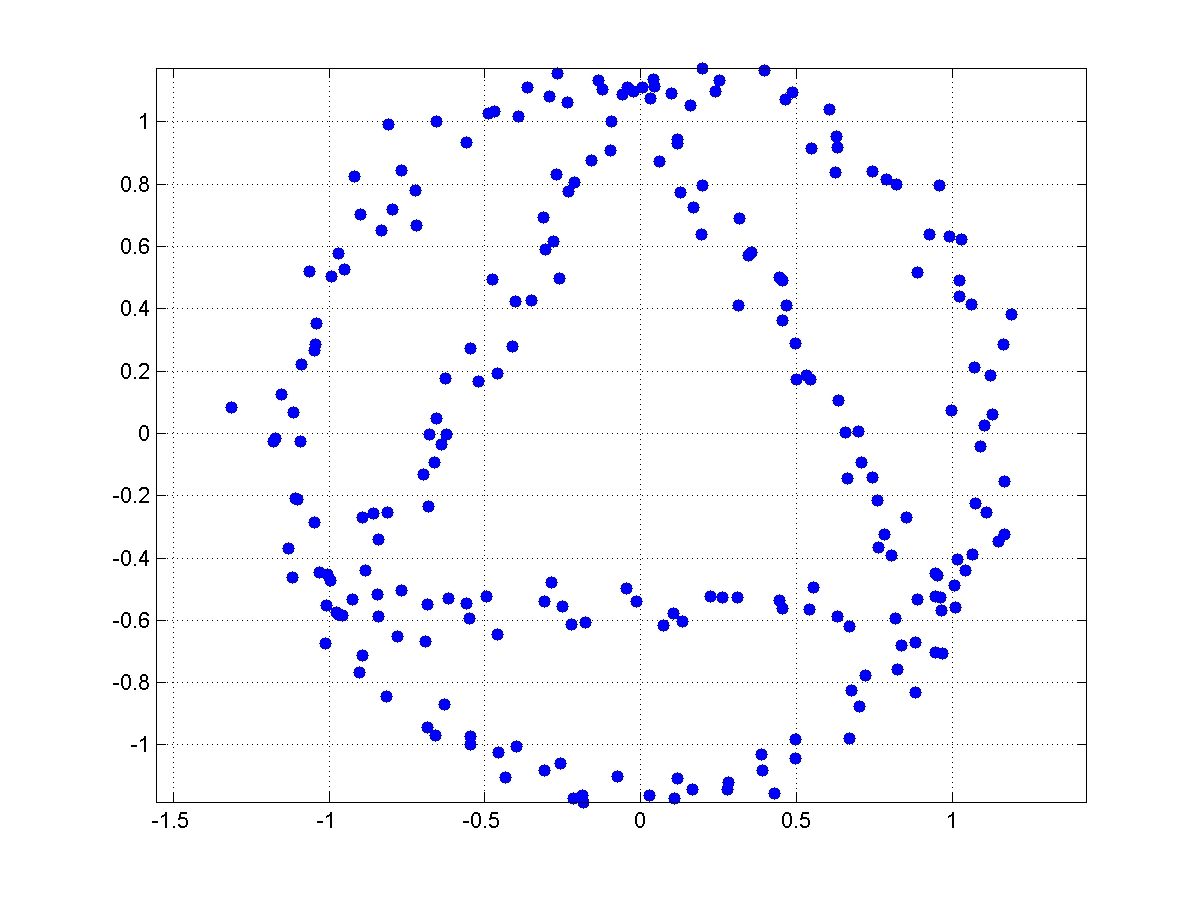}}
\centerline{
\includegraphics[width=4.5cm]{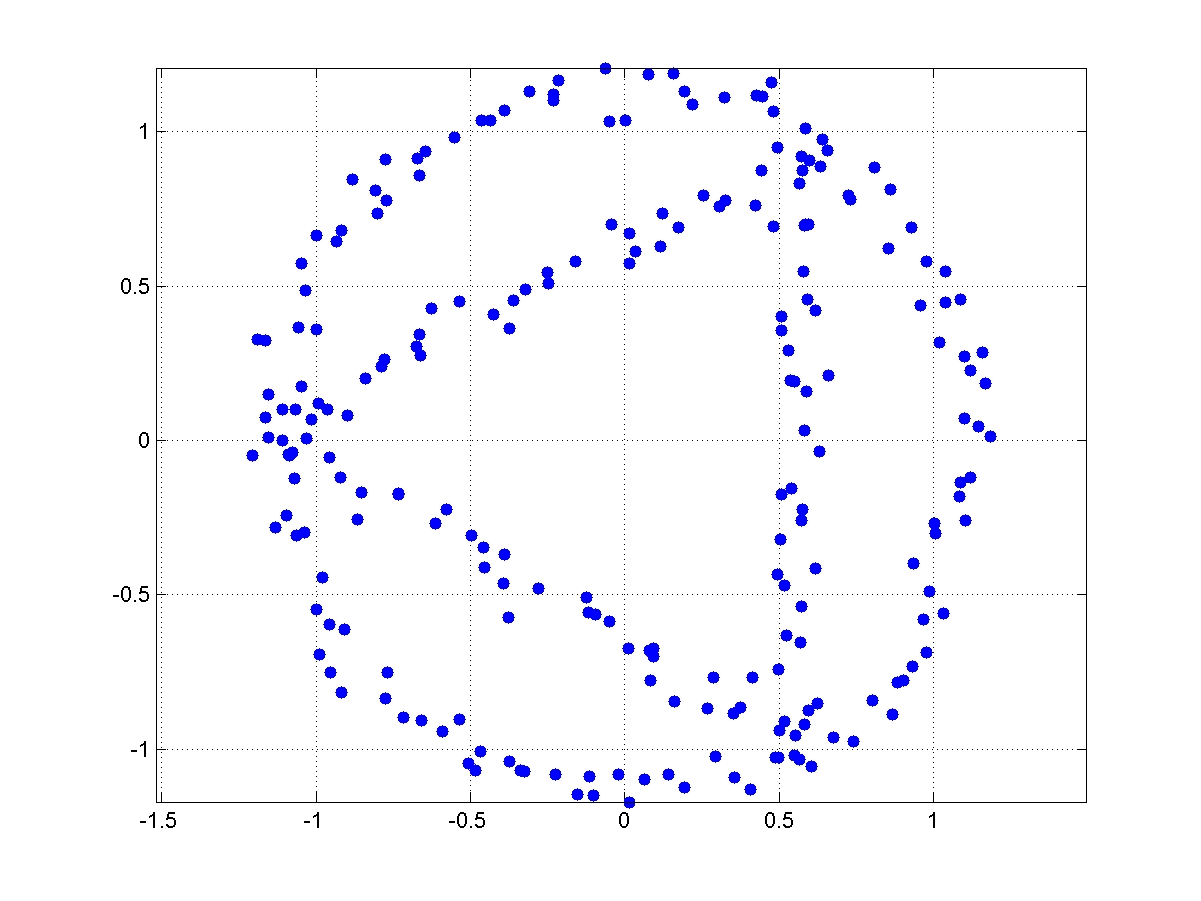}
\includegraphics[width=4.5cm]{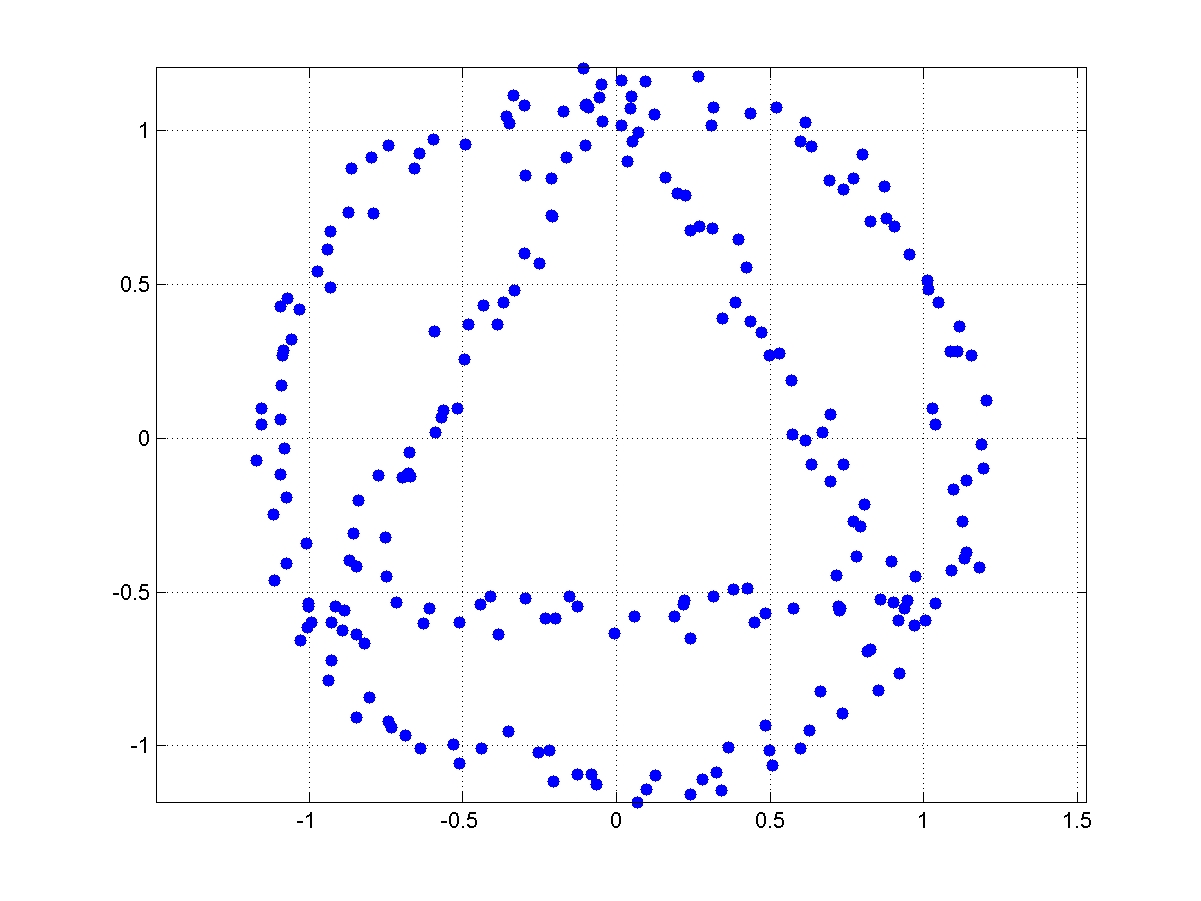}}
\centerline{
\includegraphics[width=4.5cm]{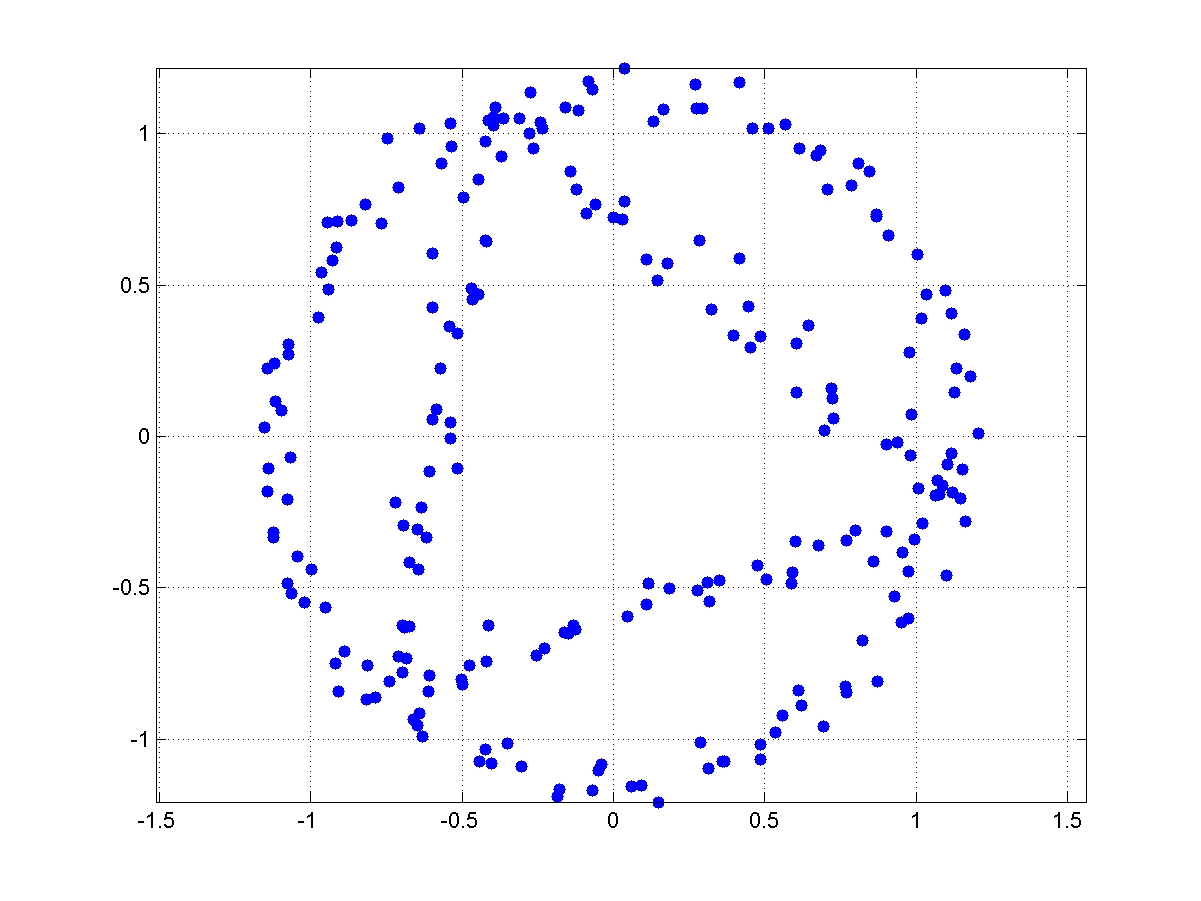}
\includegraphics[width=4.5cm]{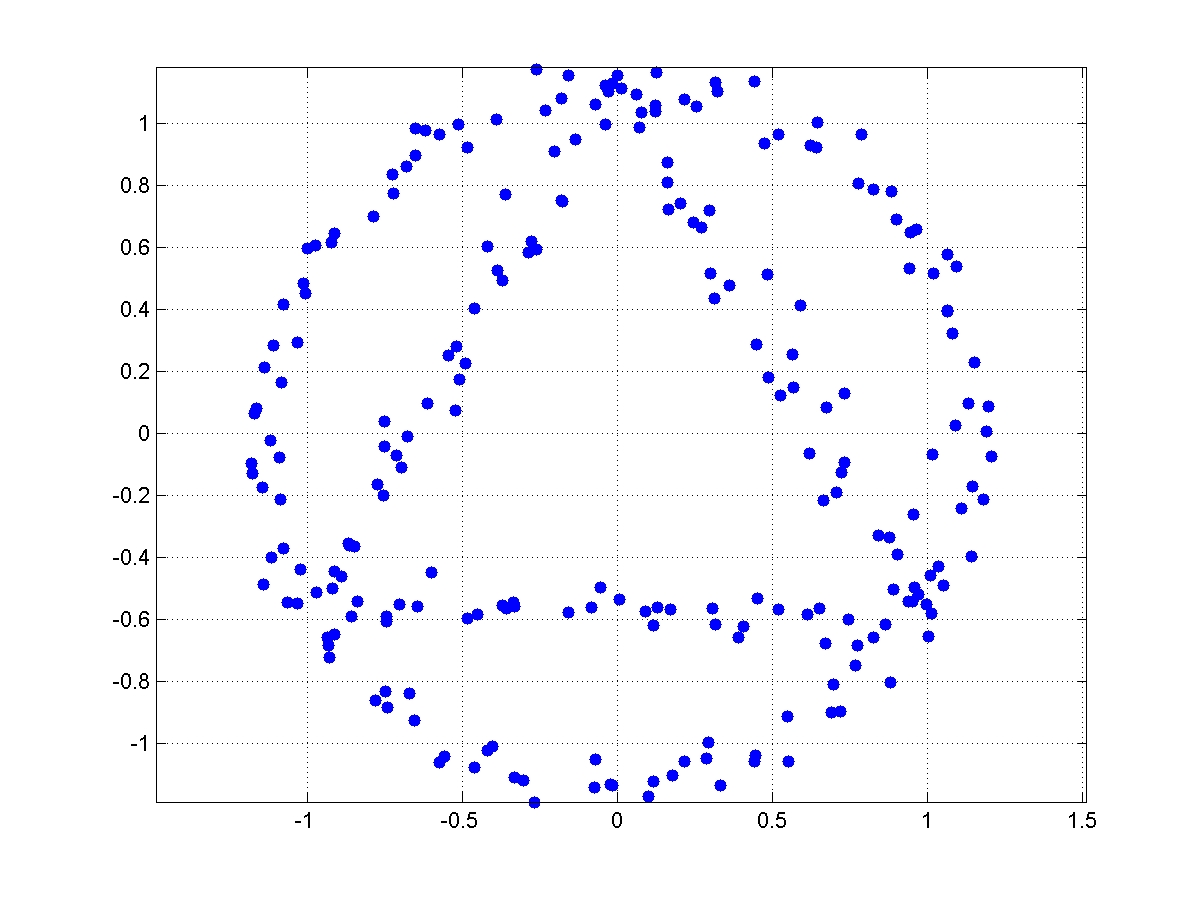}}
\centerline{
\includegraphics[width=4.5cm]{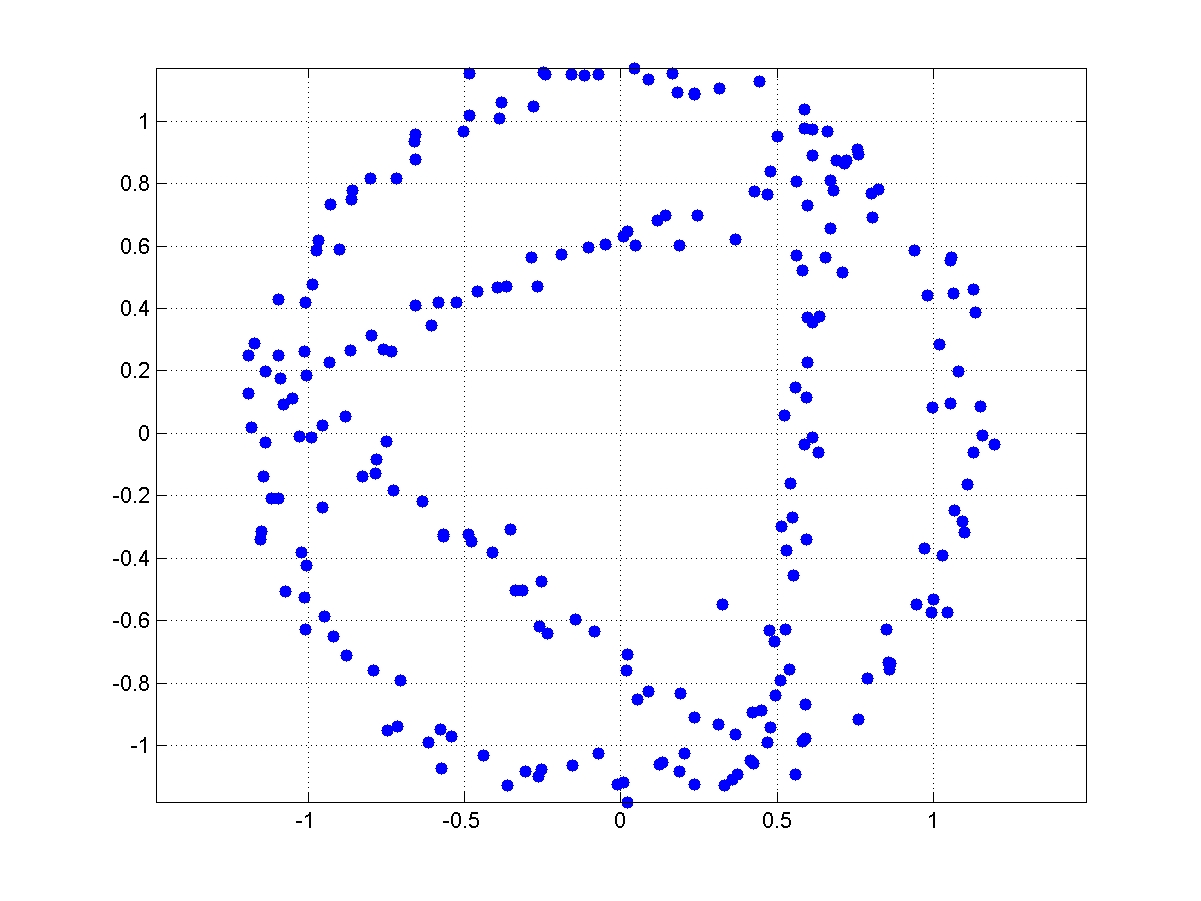}
\includegraphics[width=4.5cm]{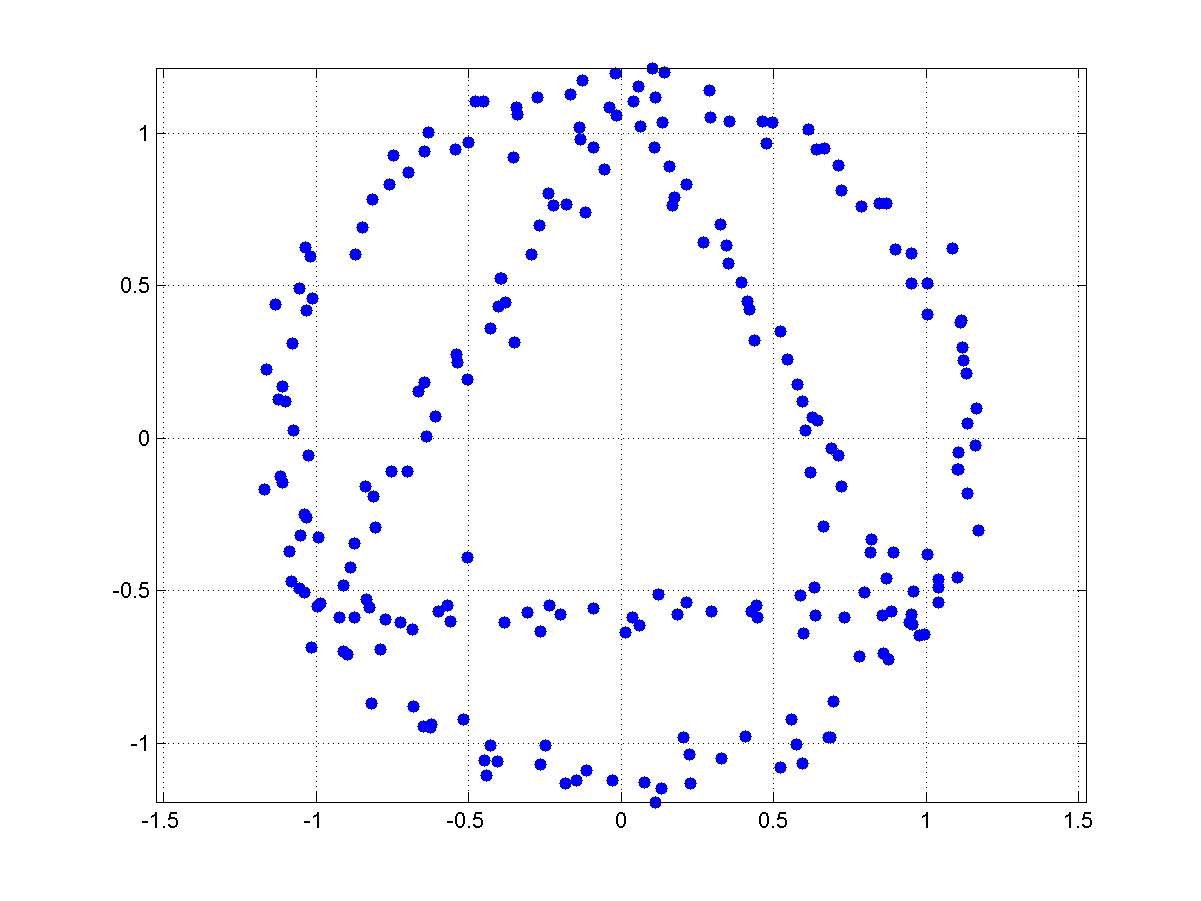}}
\caption{Examples of using PMA for computing the orientation of rotationally
symmetric shapes. Left: original shapes; right: PMA oriented shapes. In spite
of the absence of a principal axis, and the high level of noise, PMA provides
consistent orientations.\label{fig:equalex}}
\end{figure}

\begin{figure}[htb]
\centerline{\includegraphics[width=9.5cm]{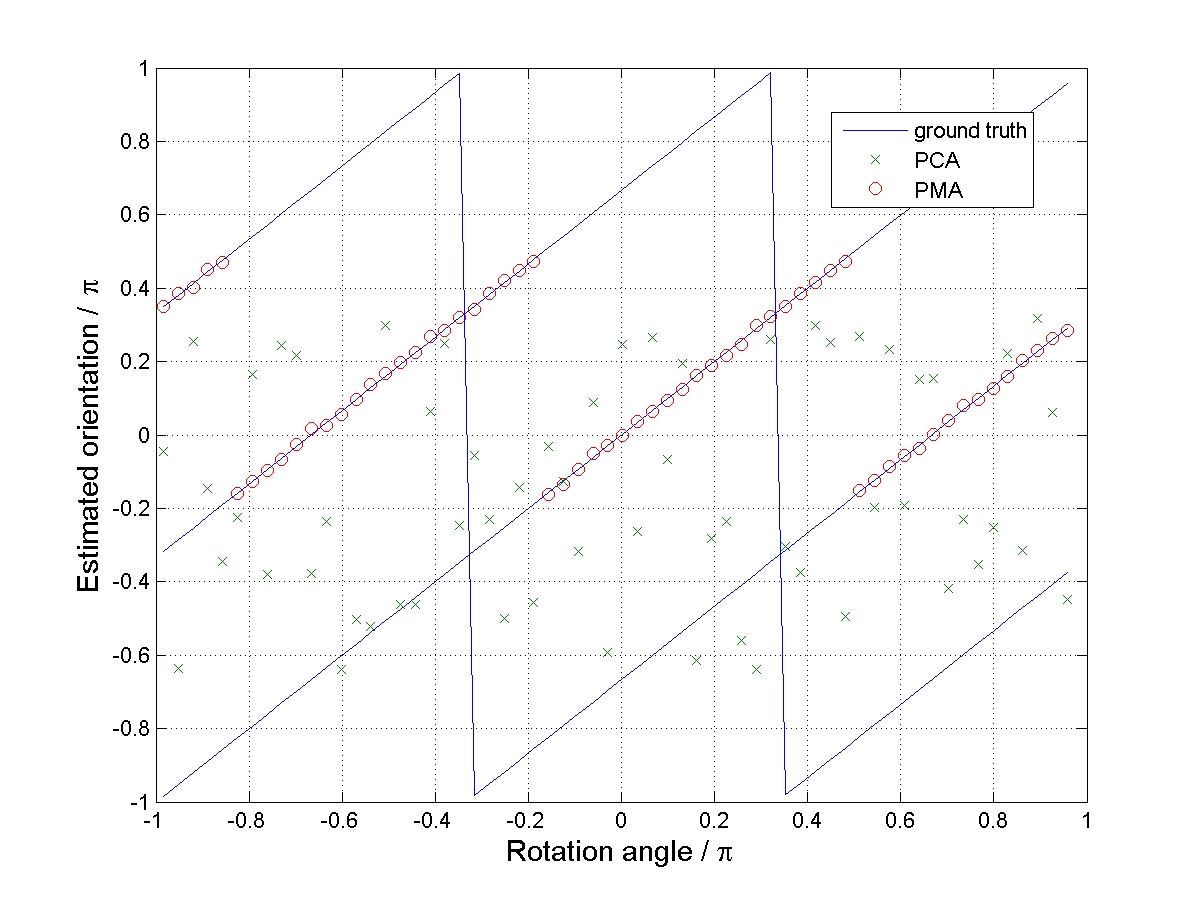}} \caption{Comparison of
PMA and PCA for rotationally symmetric shapes. In this case, with a three-fold
symmetric shape, the ground true is represented by three lines corresponding to
the three consistent orientations, {\it i.e.}, separated by
$2\pi/3$.\label{fig:equalplot}}
\end{figure}

In a similar way, we now illustrate that PMA also deals with noisy observations
of rotationally symmetric shapes.
We used noisy versions of a three-fold rotationally symmetric shape that would
be impossible to orient using PCA, see the left column of
Fig.~\ref{fig:equalex}, and normalized their orientations using PMA, obtaining
the visually correct results on the right column of Fig.~\ref{fig:equalex}. By
proceeding in a similar way as described above, we contrasted the estimates
with their ground truth, obtaining the plot in Fig~\ref{fig:equalplot}. Note
that, in this case, there are three values for the true rotation angle,
corresponding to three consistent orientations, due to the three-fold symmetry
of the shape, see expression \eqref{eq:theta_rot_symmetry}. Since the shape is rotationally symmetric, PCA is useless for the
determination of an orientation, providing results only determined by the noise.
In opposition, the plot in Fig~\ref{fig:equalplot} shows that the estimates
obtained through PMA are robust to noise.

\begin{figure}[htb]
\centerline{\includegraphics[width=3cm]{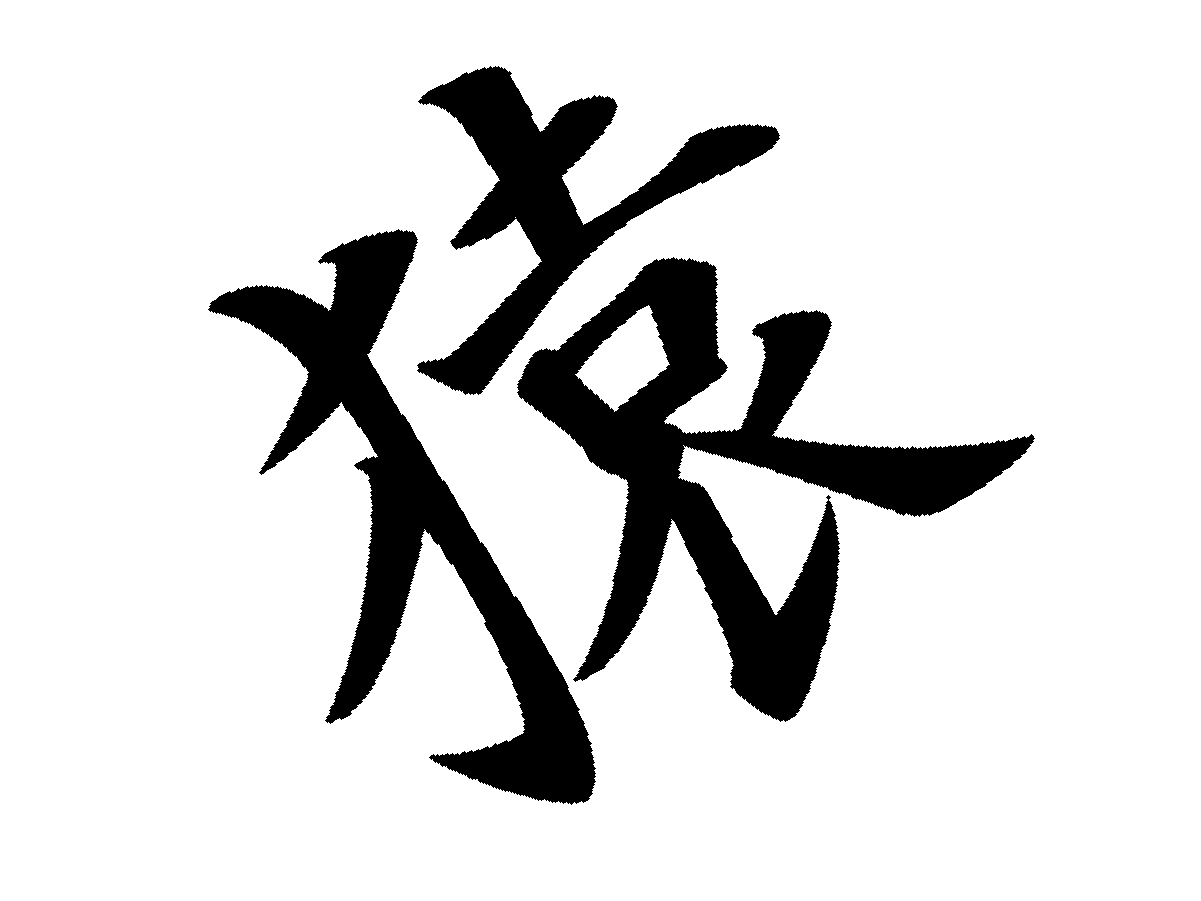} \hspace*{1cm} \includegraphics[width=3cm]{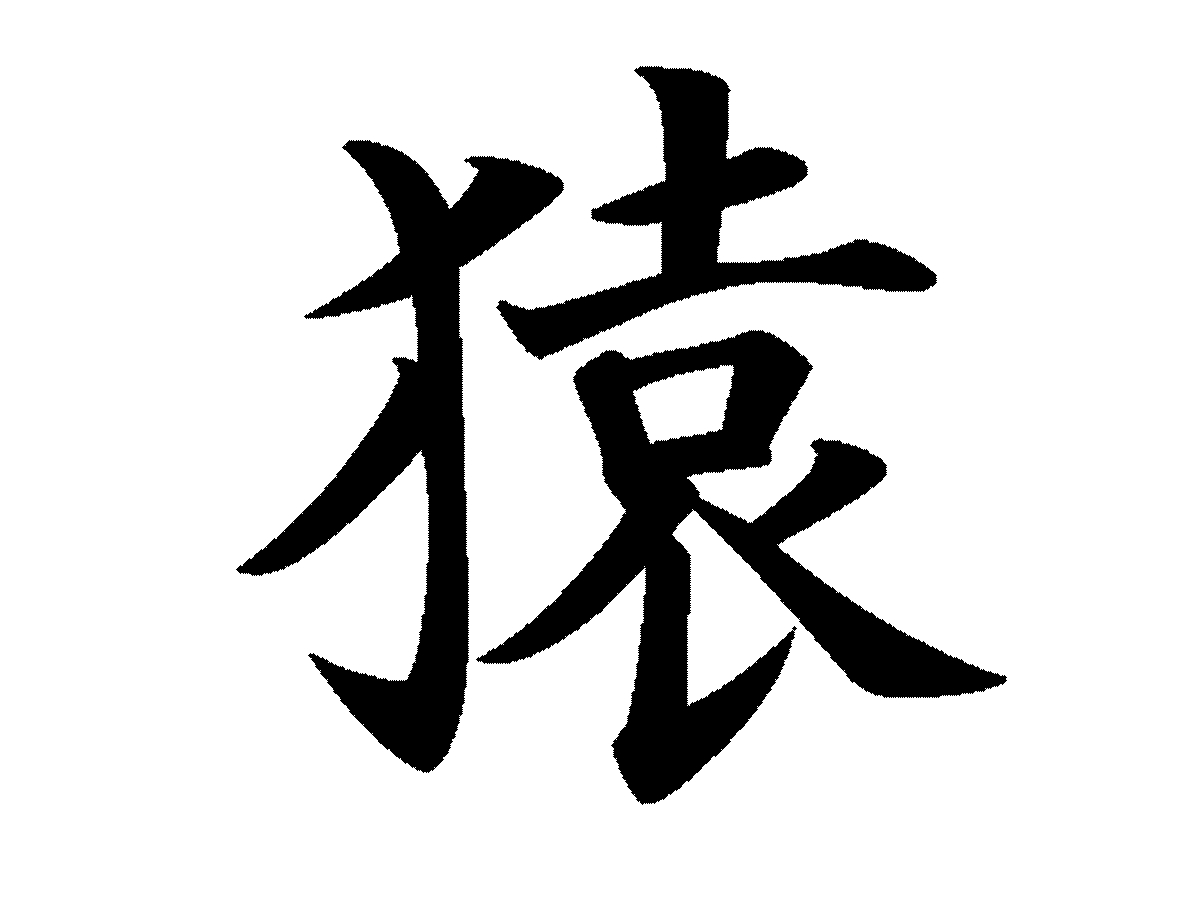}}
\centerline{\includegraphics[width=3cm]{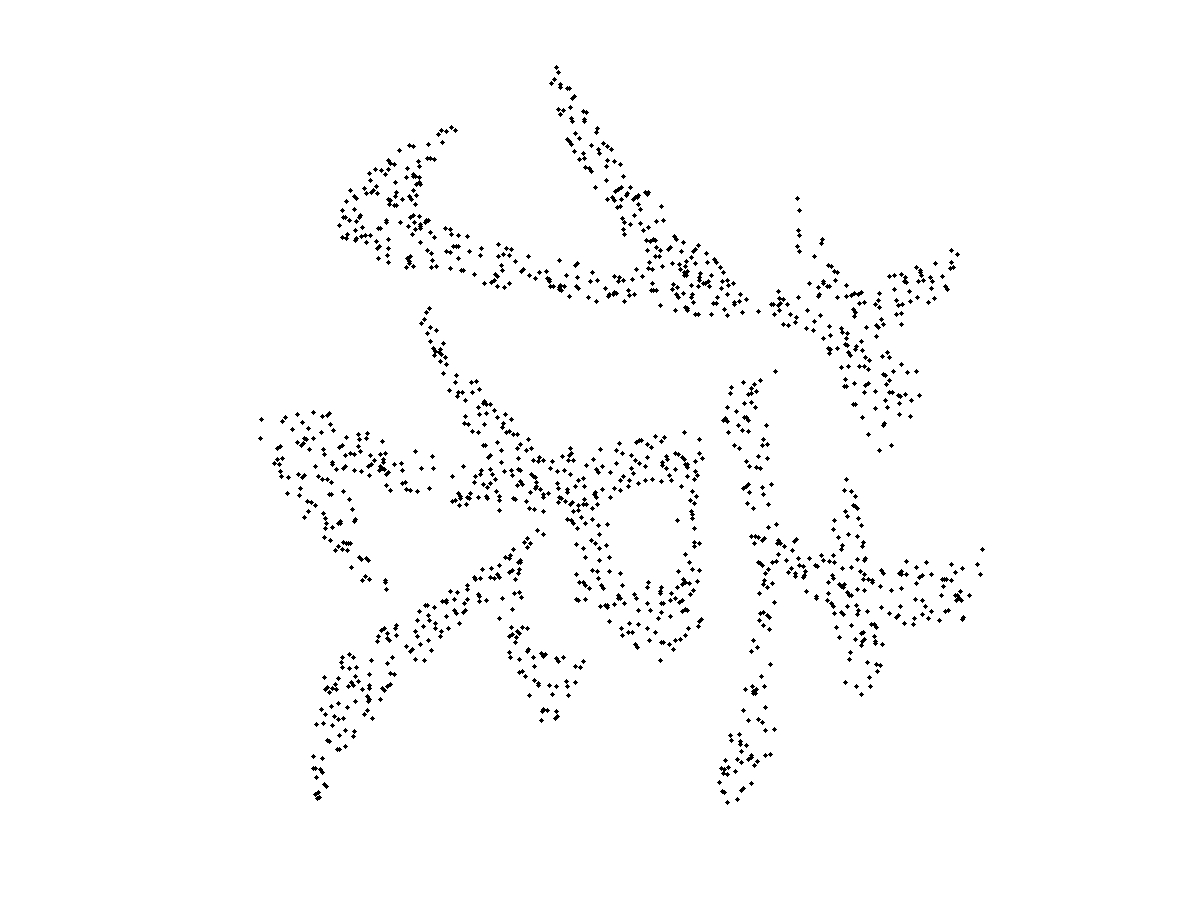} \hspace*{1cm}\includegraphics[width=3cm]{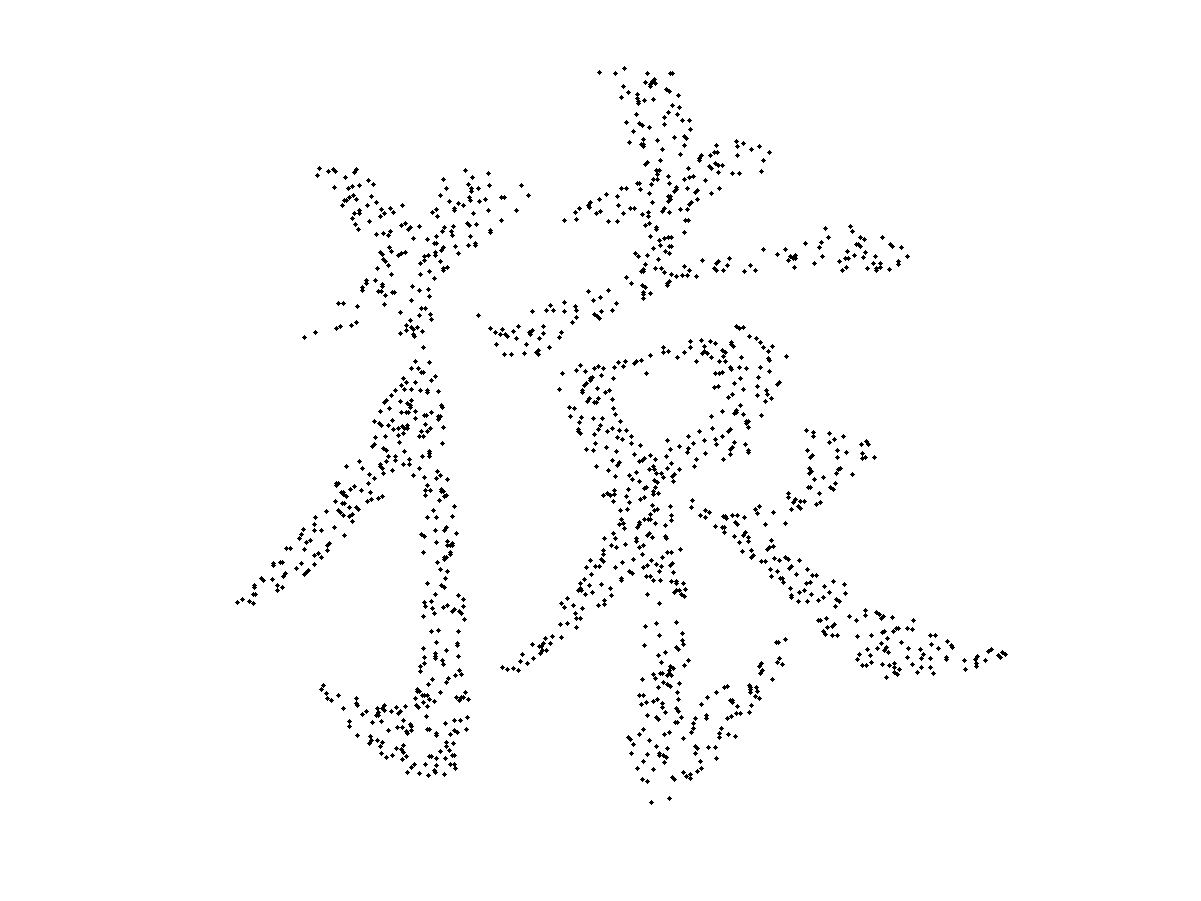}}
\centerline{\includegraphics[width=3cm]{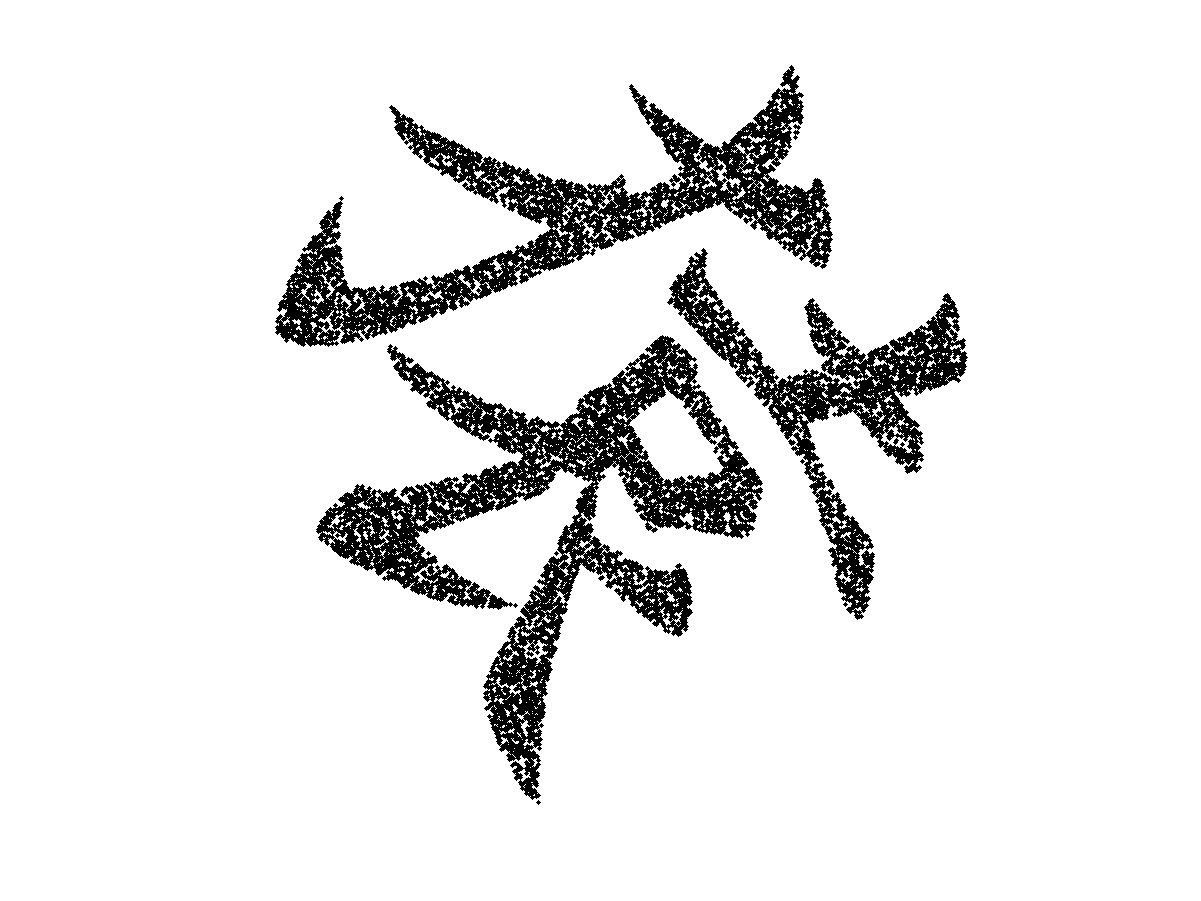}\hspace*{1cm} \includegraphics[width=3cm]{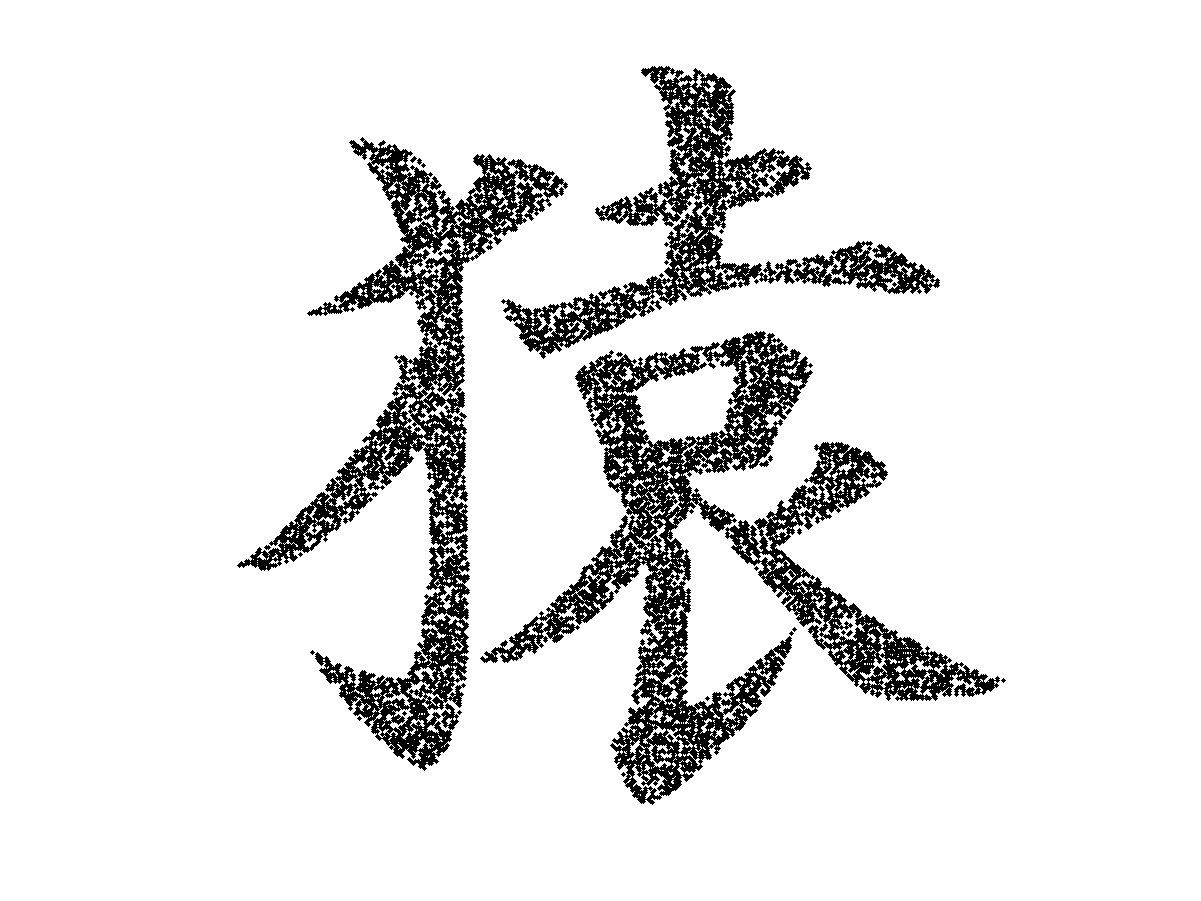}}
\centerline{\includegraphics[width=3cm]{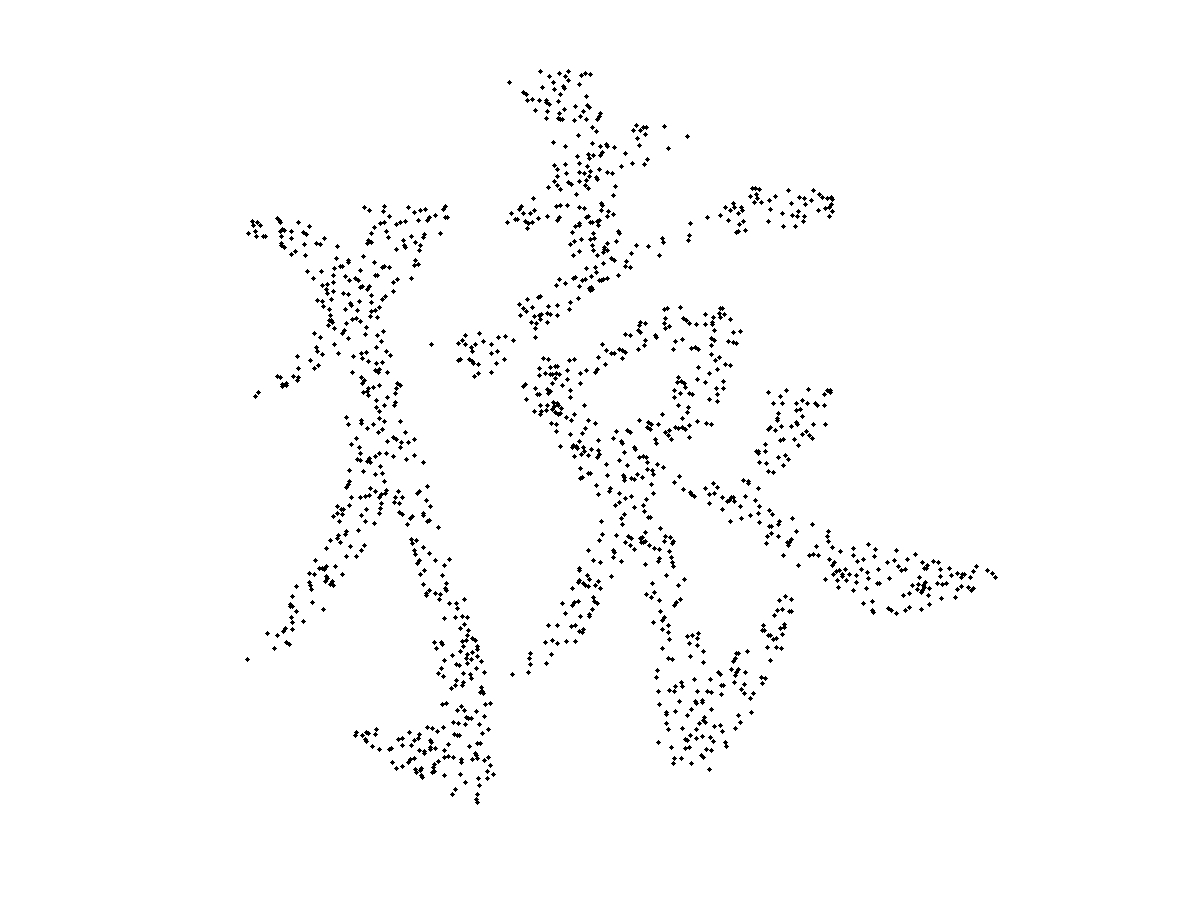} \hspace*{1cm}\includegraphics[width=3cm]{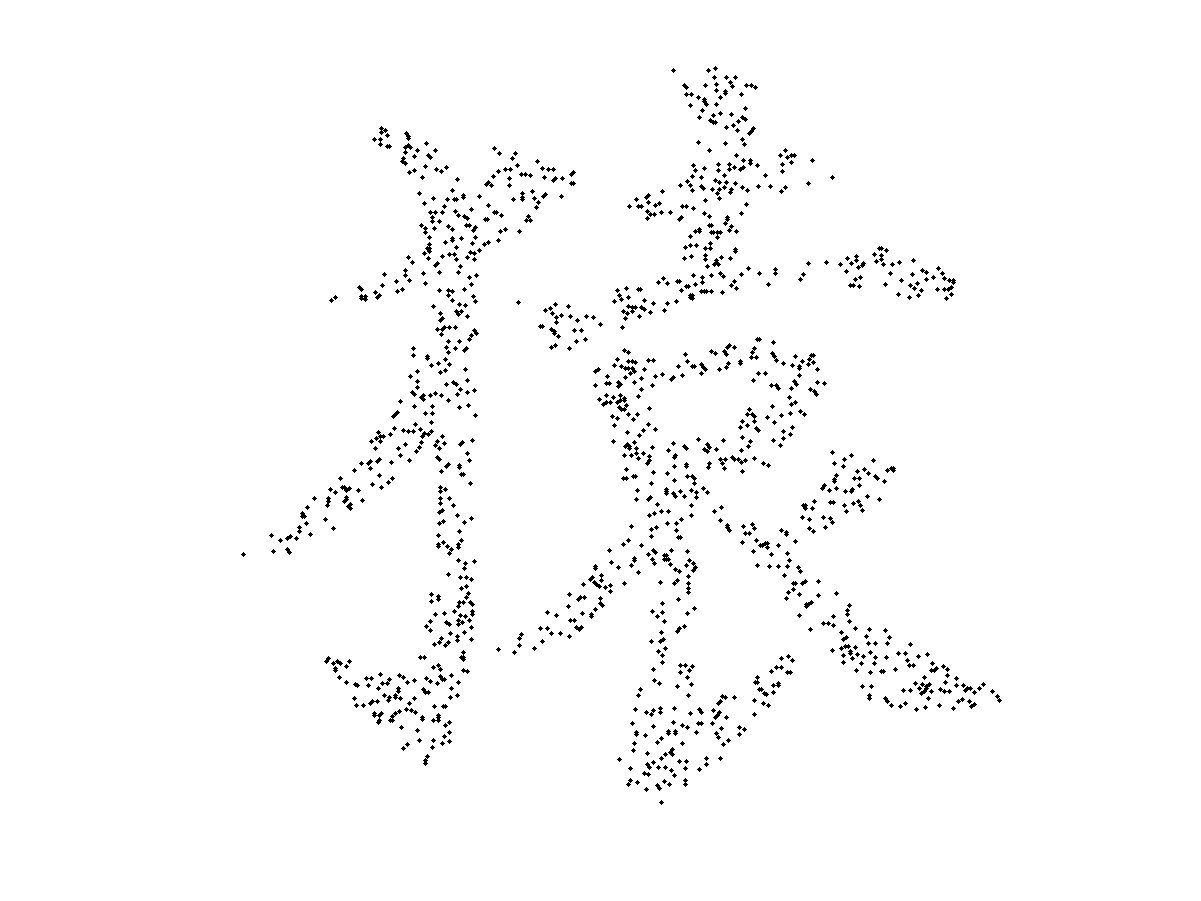}}
\centerline{\includegraphics[width=3cm]{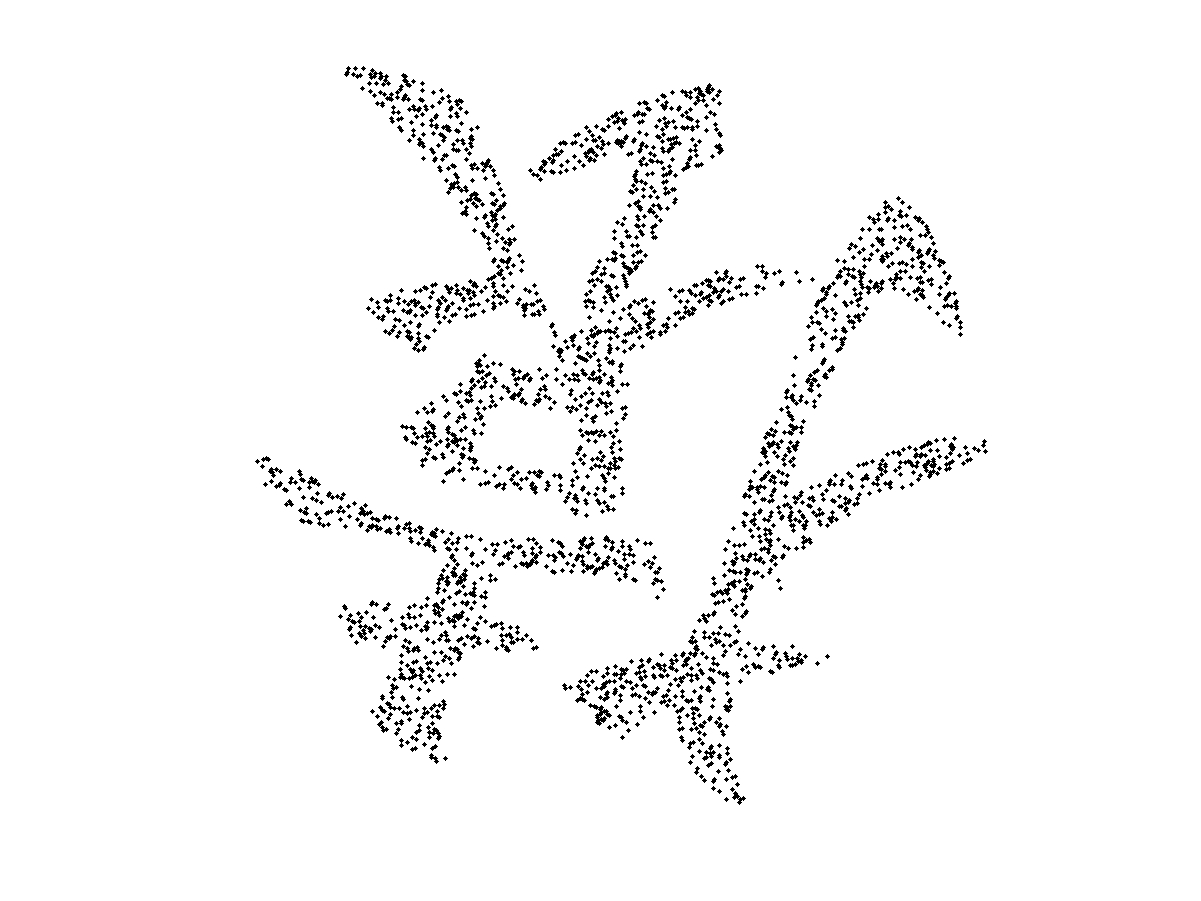} \hspace*{1cm}\includegraphics[width=3cm]{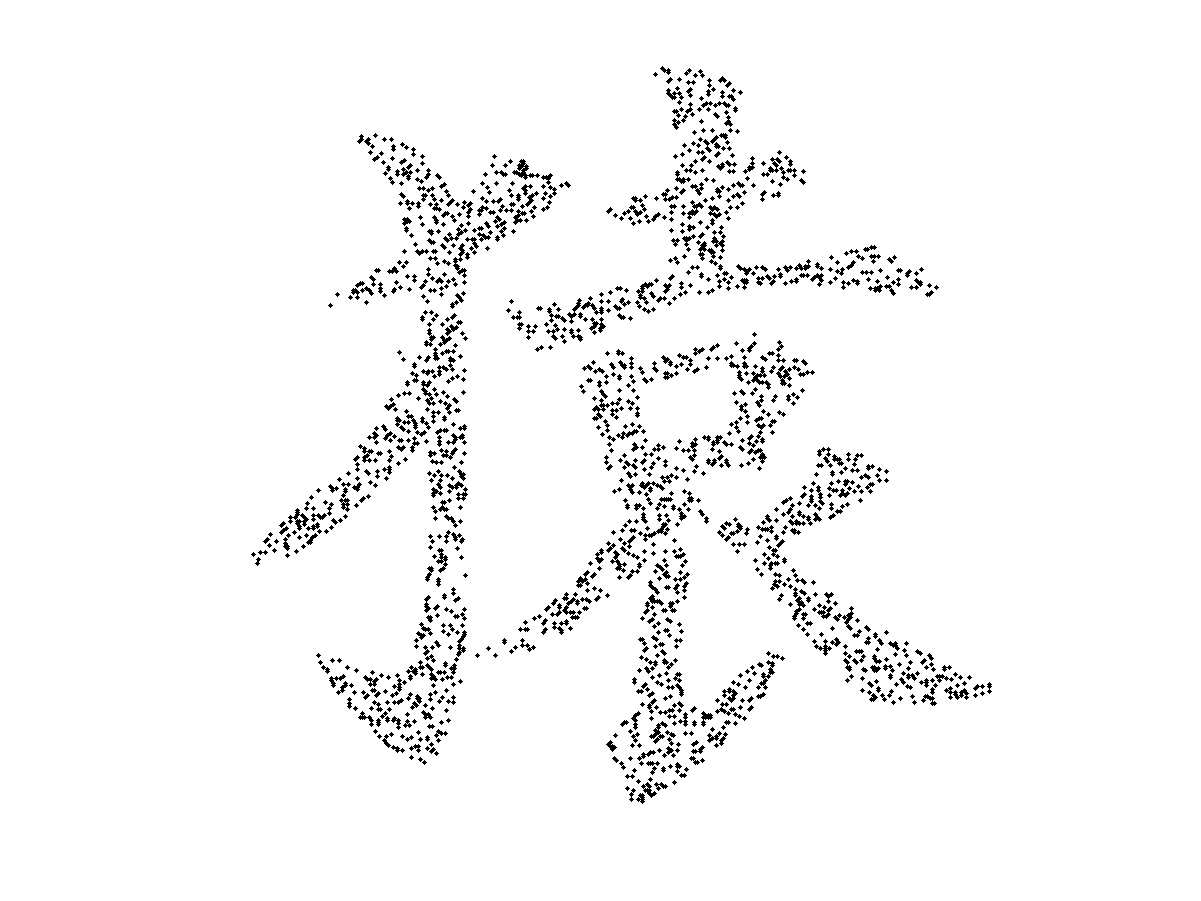}}
\centerline{\includegraphics[width=3cm]{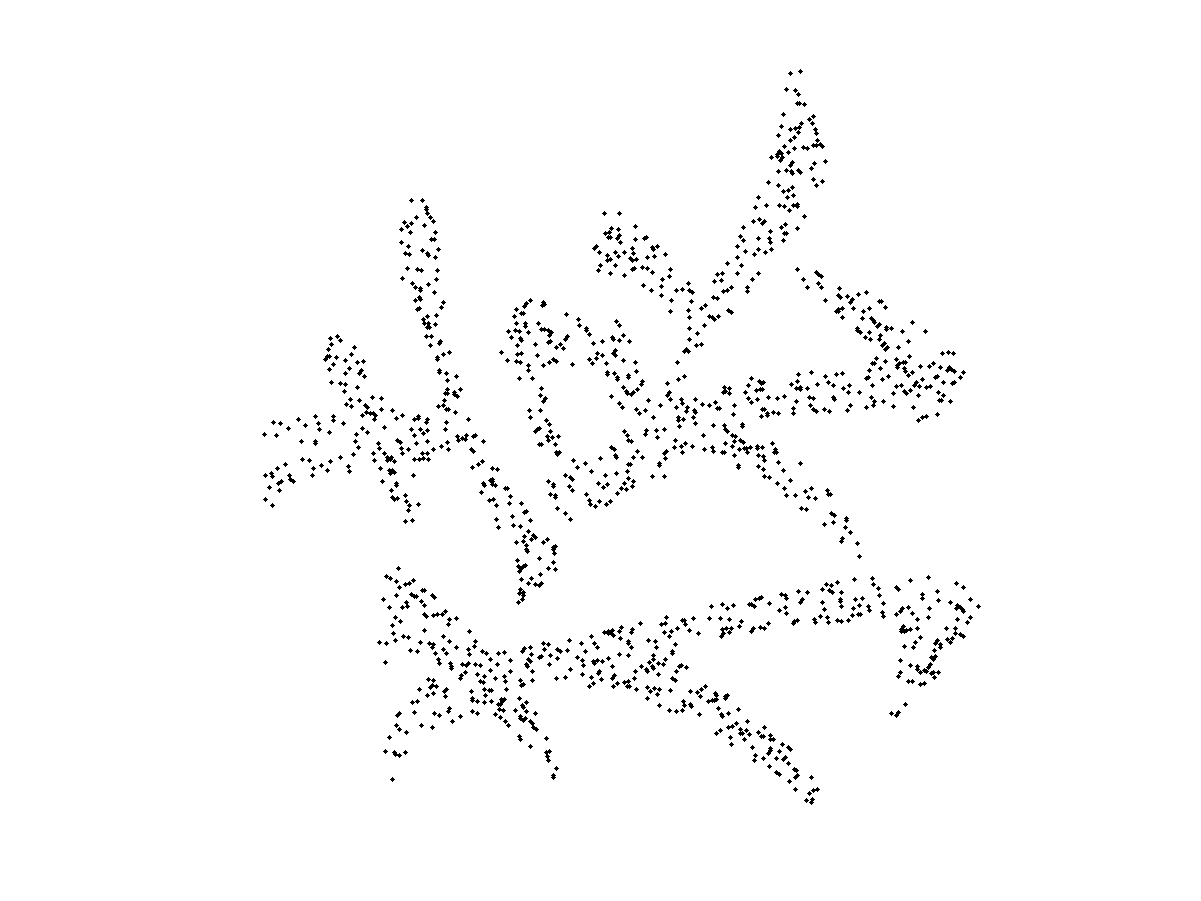} \hspace*{1cm}\includegraphics[width=3cm]{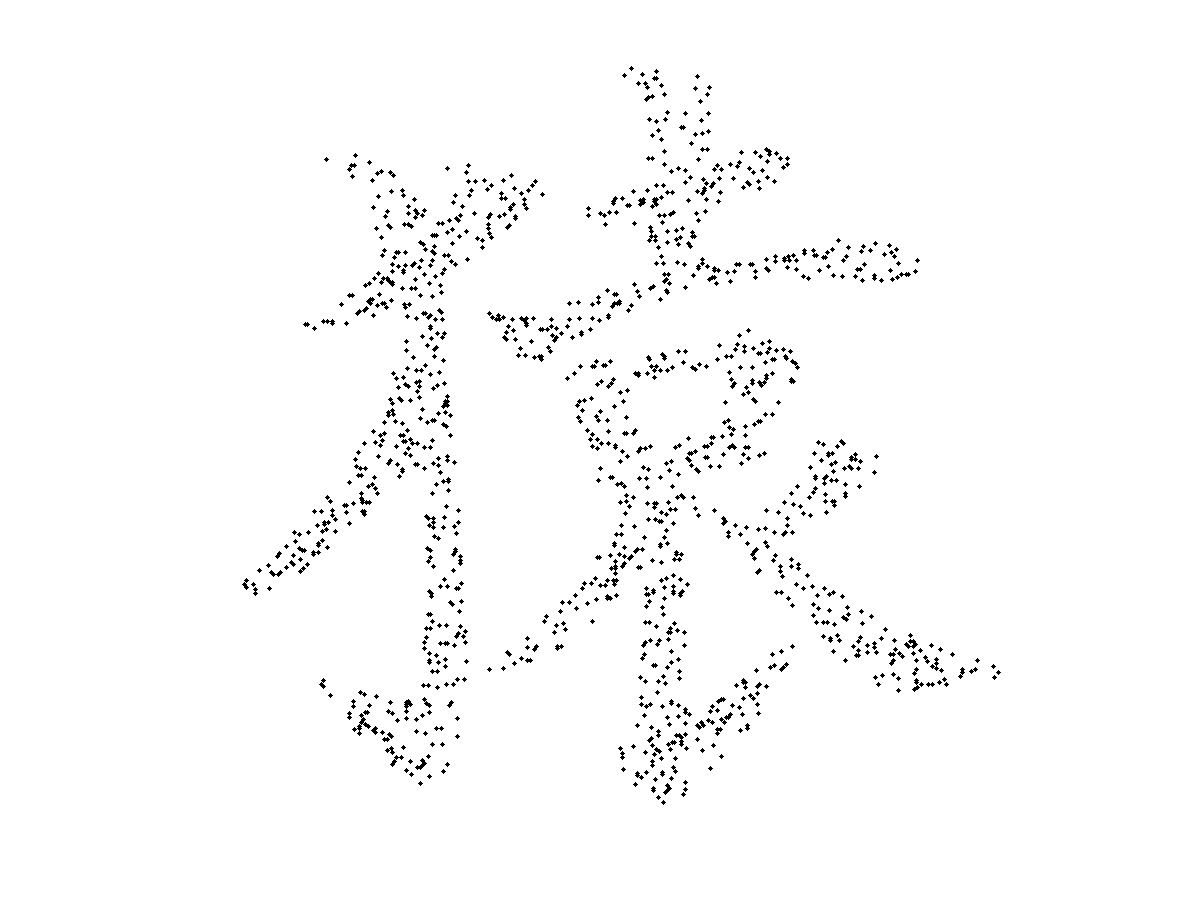}}
\caption{Using PMA to normalize corrupted Japanese characters.\label{fig:monkey}}
\end{figure}

To illustrate the robustness of PMA to the shape sampling density, we used
Japanese characters. We synthesized corrupted versions of those characters by
removing up to $95\%$ of the shape points, obtaining shape vectors of the form of expression
\eqref{eq:shapevector} but of very distinct cardinality. We then processed these
vectors by using PMA. In Fig.~\ref{fig:monkey}, we single out five instances of
a specific character to illustrate the consistent orientations obtained for all
the corrupted characters.

\subsection{PMA for Normalization of Grey-level Images}
\label{subsec:exppmagrey}

\begin{figure}[htb]
\centerline{\includegraphics[width=4cm]{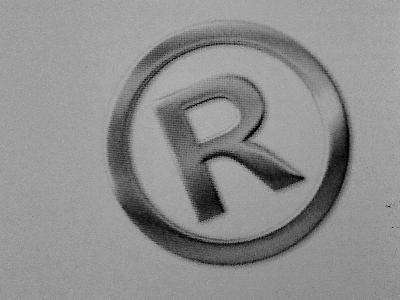}  \includegraphics[width=4cm]{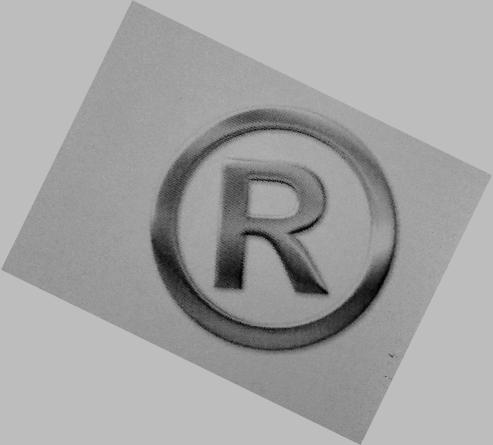}}\vspace*{.2cm}
\centerline{\includegraphics[width=4cm]{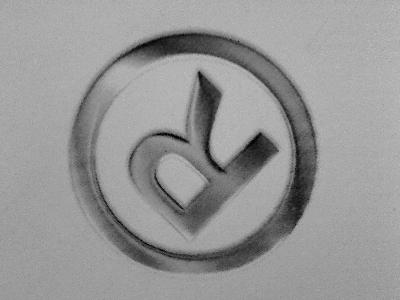}  \includegraphics[width=4cm]{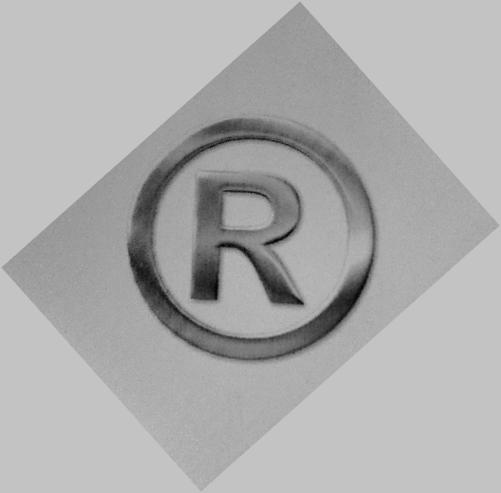}}\vspace*{.2cm}
\centerline{\includegraphics[width=4cm]{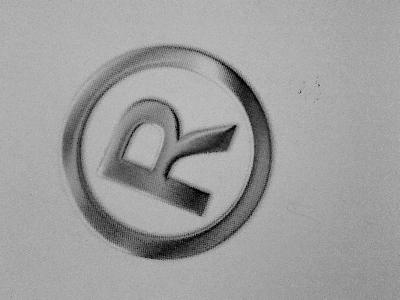}  \includegraphics[width=4cm]{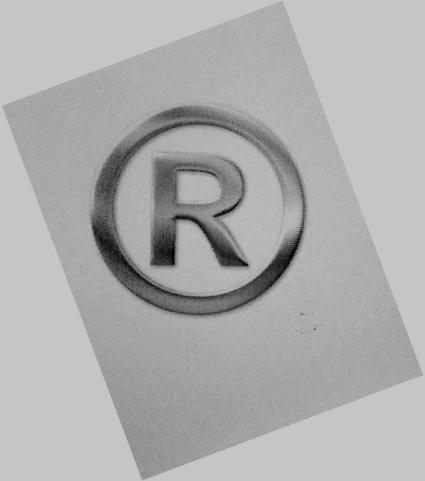}}\vspace*{.2cm}
\centerline{\includegraphics[width=4cm]{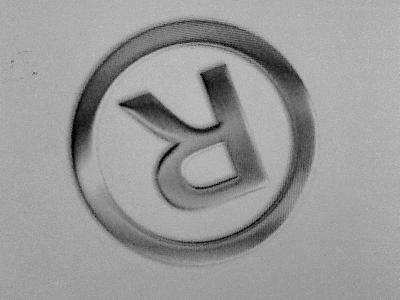}  \includegraphics[width=4cm]{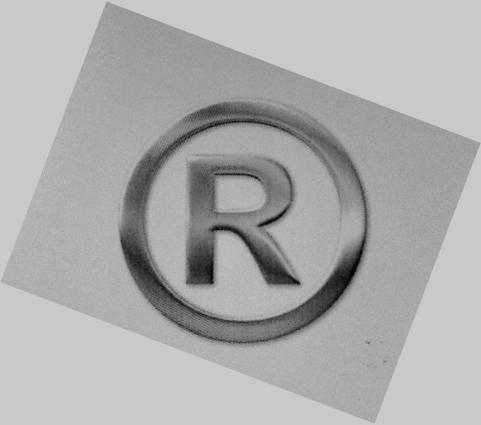}}
\caption{Using PMA with gray-scale photos of trademark logos.\label{fig:trade}}
\end{figure}

\begin{figure}[htb]
\centerline{\includegraphics[width=4cm]{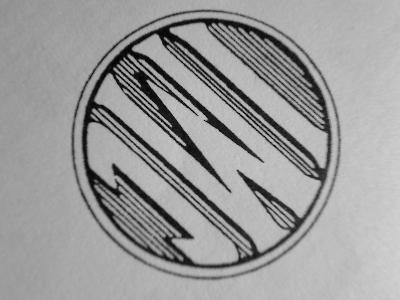}  \includegraphics[width=4cm]{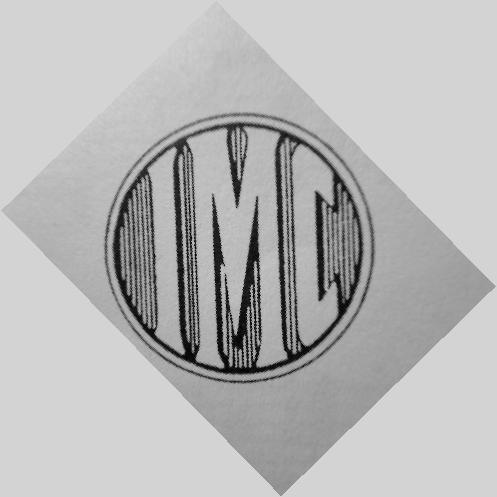}}\vspace*{.2cm}
\centerline{\includegraphics[width=4cm]{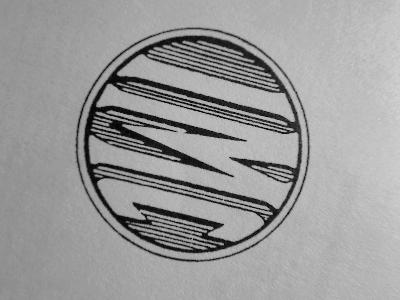}  \includegraphics[width=4cm]{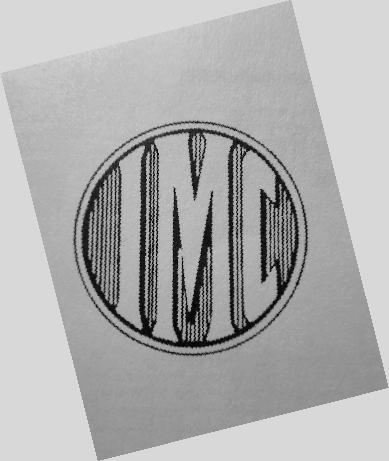}}\vspace*{.2cm}
\centerline{\includegraphics[width=4cm]{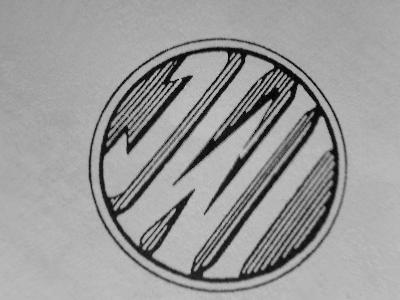}  \includegraphics[width=4cm]{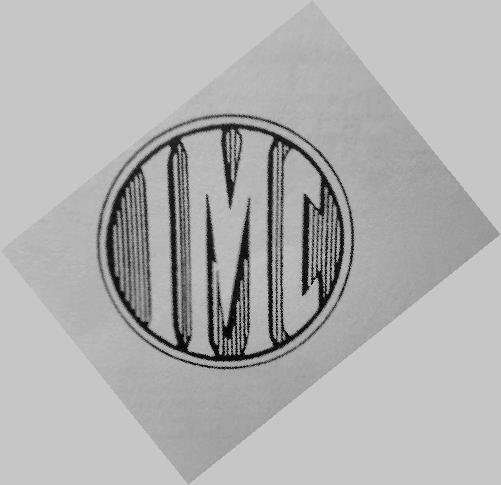}}\vspace*{.2cm}
\centerline{\includegraphics[width=4cm]{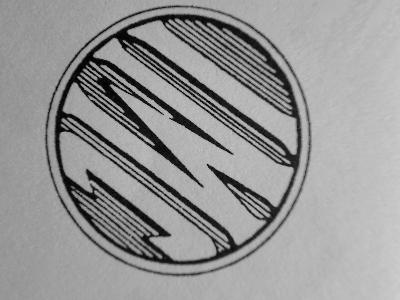}  \includegraphics[width=4cm]{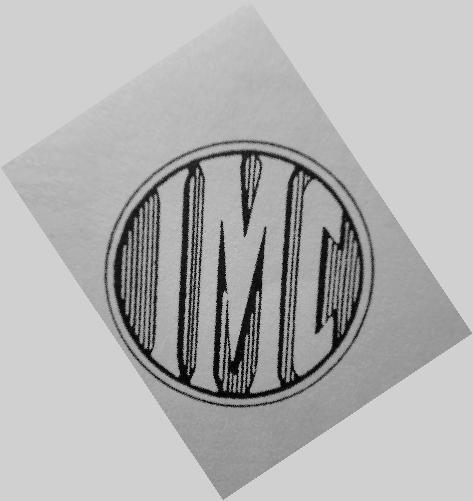}}
\caption{Using PMA with gray-scale photos of trademark logos.\label{fig:trade2}}
\end{figure}

\begin{figure}[htb]
\centerline{\includegraphics[width=4cm]{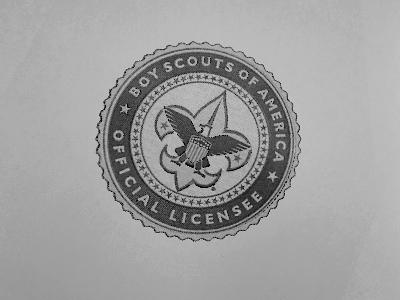}  \includegraphics[width=4cm]{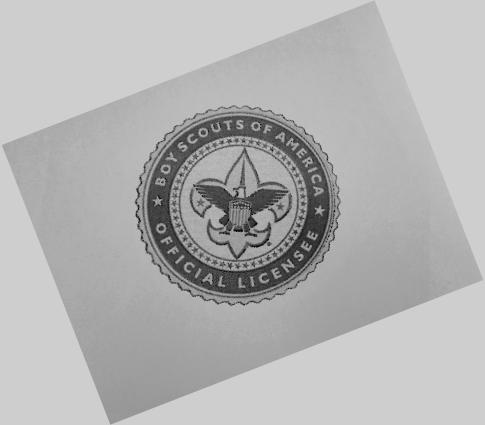}}\vspace*{.2cm}
\centerline{\includegraphics[width=4cm]{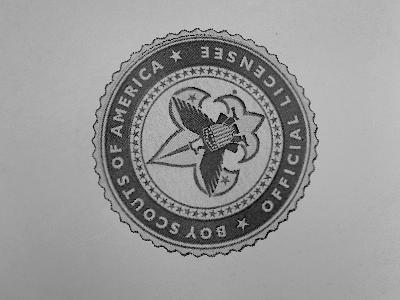}  \includegraphics[width=4cm]{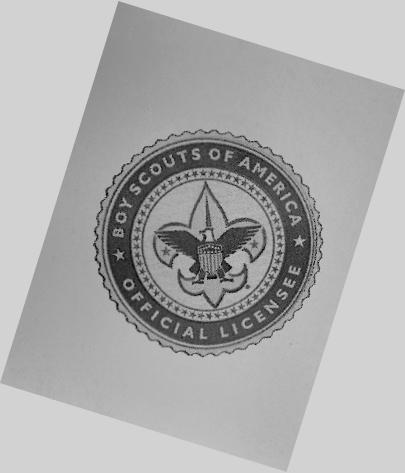}}\vspace*{.2cm}
\centerline{\includegraphics[width=4cm]{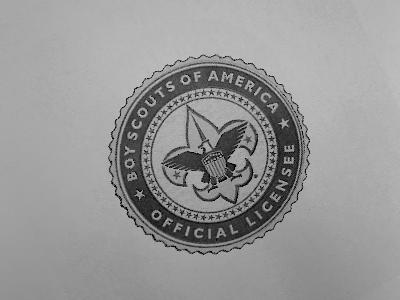}  \includegraphics[width=4cm]{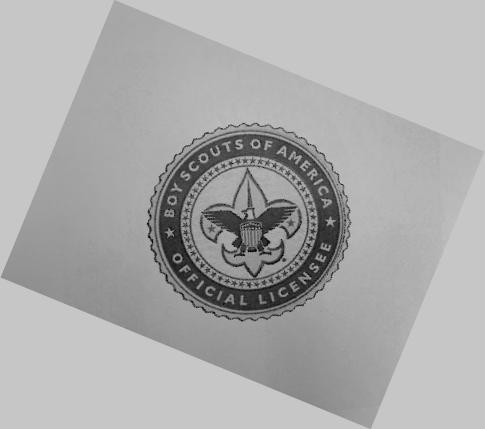}}\vspace*{.2cm}
\centerline{\includegraphics[width=4cm]{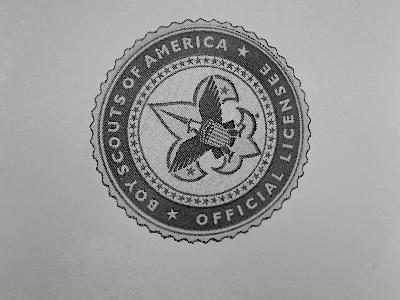}  \includegraphics[width=4cm]{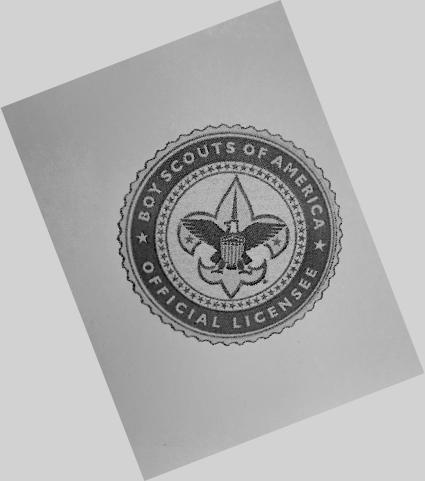}}
\caption{Using PMA with gray-scale photos of trademark logos.\label{fig:trade3}}
\end{figure}

Finally, we illustrate the usage of PMA to rotationally normalize grey-level images.
We used real images, consisting of photos of trademark logos,
see examples in the left columns of Figs.~\ref{fig:trade}-\ref{fig:trade3}.
By proceeding as described in Section~\ref{sec:grey}, we used PMA to normalize the orientation of the photographed logos.
In spite of geometric distortions ({\it e.g.}, perspective, radial) and
other intensity disturbances in the images, we got consistent results, as illustrated by the oriented versions of the logos
in the right columns of Figs.~\ref{fig:trade}-\ref{fig:trade3}. Note that these examples correspond to particularly
challenging grey-level images, due to approximate rotational symmetry.

Since we detailed several situations where current methods fail, see
Section~\ref{sec:mom}, we do not report here experimental results obtained with
those algorithms. In fact, it would be easy to produce examples where estimates
obtained through PMA would be much more accurate than those obtained by using
other methods (just imagine using shapes similar to the ones in
Figs.~\ref{fig:ex-nrsi}, \ref{fig:ex-rsi2}, or \ref{fig:ex-nrsi-gc}). However, we found it would be
more informative to present the discussion in Section~\ref{sec:mom}
regarding the core limitations of those methods, {\it i.e.}, to show how they
attempt to use information that is not available in all shapes, than to blindly
report sample experiments to support our approach.

\section{Conclusion}
\label{sec:conc}

We proposed to represent 2D~shapes, {\it i.e.}, sets of unlabeled points or landmarks,
via particular complex moments that we call {\it Principal Moments} (PMs).
This representation is complete and we show it is compact,
in the sense that the number of PMs needed to discriminate between shapes is small
(and dependent on their complexity). We further presented a new method,
{\it Principal Moments Analysis} (PMA) to unambiguously compute a unique orientation
for arbitrary 2D~shapes. This enables performing rotational normalization, thus obtaining
maximally invariant ({\it i.e.}, complete) representations for 2D~shapes. We finalized by
extending PMA to the normalization of grey-level images.
Besides theoretically sound, PMA results are robust to noise.

\section*{Acknowledgements}

This work was partially supported by FET, within the EU--FP7, under SIMBAD
project (contract 213250), and FCT, under ISR/IST plurianual funding (POSC
program, FEDER) and grant MODI-PTDC/EEA-ACR /72201/2006.

\appendix

\section{On the Normalization of Power Sums}
\label{app:norm}

In this appendix, we derive an expression for the expected growth of the magnitude of the power sum
\begin{equation}
\mu_k=\sum_{n=1}^{N}z_n^k\,,\qquad k\in\{1,2,3,\ldots\}\,, \label{eq:app:pm_simple}
\end{equation}
under reasonable assumptions for the set of 2D points $\left\{z_n\right\}$.

Let $\left\{z_n, n=1,2,\ldots,N\right\}$, be samples of a complex random variable (r.v.). Due to the common pre-processing of centering the shape \eqref{eq:translation_scale_norm}, we assume this r.v.~is zero mean, {\it i.e.}, $\E\{z_n\}=0, \forall_n$. We also assume the likelihood of each direction is the same, {\it i.e.}, the angle $\arg(z_n)$ is uniform in  $[0,2\pi)$ and independent of the absolute value $|z_n|$. Under these assumptions, we obtain
\begin{align}
\E\left\{z_n^k\right\}&=\E\left\{|z|^ke^{jk\arg(z_n)}\right\}\nonumber\\
&=\E\left\{|z|^k\right\}\E\left\{e^{jk\arg(z_n)}\right\}\label{eq:e1}\\
&=0\,,\label{eq:e2}
\end{align}
where \eqref{eq:e1} is due to the independence between $|z_n|$ and $\arg(z_n)$ and \eqref{eq:e2} is due to the uniformity of $\arg(z_n)$. Using this result, we conclude that the mean value of the power sums is zero:
\begin{equation}
\E\left\{\mu_k\right\}=\E\left\{\sum_{n=1}^{N} z_n^k\right\}=\sum_{n=1}^{N}\E\left\{z_n^k\right\}=0\,.
\end{equation}
The issue we address in the sequel is the expected grow of $|\mu_k|$.

We start by expressing $\E\{|\mu_k|^2\}$ in terms of a moment of the real r.v.~$|z|$, through the chain of equalities
\begin{eqnarray}
\E\left\{|\mu_k|^2\right\} &=&\E\left\{\mu_k\mu_k^*\right\} \nonumber\\
&=& \E\left\{\sum_{n=1}^{N} z_n^k \sum_{m=1}^{N}{z_m^k}^*\right\} \nonumber\\
&=& \sum_{n,m=1}^{N}\E\left\{\left(z_nz_m^*\right)^k\right\} \nonumber\nonumber\\
&=& \sum_{n=1}^{N}\E\left\{\left(z_nz_n^*\right)^k\right\} \label{ind} \\
&=& \sum_{n=1}^{N}\E\left\{|z_n|^{2k}\right\} \nonumber\\
&=& N\E\left\{|z_n|^{2k}\right\}\,, \label{final}
\end{eqnarray}
where ${}^*$ denotes the complex conjugate and \eqref{ind} is due to the independence between $z_n$ and $z_m$ for $n\neq m$.
Expression \eqref{final} states that $\E\{|\mu_k|^2\}$ is proportional to the moment of order $2k$ of the real r.v.~$|z|$ (this type of moments is often referred as {\it raw moments}, to emphasize that the corresponding r.v.~is not zero mean, as it is obviously the case of $|z|$).

Depending on the probability density function (p.d.f.) of the r.v.~$z$, we obtain different growing rates for $|\mu_k|$. For example, if the p.d.f.~of $z$, besides being circularly symmetric on the complex plane, is Gaussian, {\it i.e.}, if $z$ is a 2D Gaussian r.v.~with co-variance proportional to the identity matrix, its absolute $|z|$ is a Rayleigh r.v. \citep[see, {\it e.g.},][]{papoulis91}. The $p^{\mbox{\scriptsize th}}$-order raw moment of a Rayleigh r.v. is given by
\begin{equation}
\cM_p=\E\left\{|z_n|^{p}\right\}=\sigma^p 2^{p/2}\Gamma(1+p/2)\,,\label{eq:rm}
\end{equation}
where
\[
\Gamma(x)=\int_{0}^{\infty}t^{x-1}e^{-t}\,dt
\]
is the {\it Gamma function}, which, for an integer argument, is given by
\begin{equation}
\Gamma(q)=(q-1)!\label{eq:g}
\end{equation}
\citep[see][]{papoulis91}. From \eqref{final}, \eqref{eq:rm}, and \eqref{eq:g}, we finally get
\begin{eqnarray}
\E\{|\mu_k|^2\} &=& N\cM_{2k}\nonumber\\
&=& N\sigma^{2k} 2^k\Gamma(1+k)\nonumber\\
&=& N\left(2\sigma^2\right)^k k!\,.\label{eq:bigo}
\end{eqnarray}
Thus, for shapes respecting our assumptions, the moment $\mu_k$ should be normalized according to the square root of \eqref{eq:bigo}, which, using Stirling's approximation \citep{paris01}, can be shown to be $\sqrt{\E\{|\mu_k|^2\}}=o(k!)$.

\section{Grey-level Images That Make PMA Fail}
\label{app:exc}

In this appendix we discuss the conditions under which grey-level images have only a few (or even a single) nonzero moments, preventing PMA to work as desired. As anticipated in Section~\ref{sec:grey}, the limitations come from images that are not rotationally symmetric but ``appear to be", in the sense that their nonzero moments $\mu_k$ occur only for indices $k$ multiples of a given $\gamma>1$. These images $f(r,\theta)$ have nonzero Fourier series coefficients $F(r,k)$, given by~\eqref{eq:sfgrey}, that are canceled out in $\mu_k$ by the integral in \eqref{eq:pmsf}.

\begin{figure}[htb]
\centerline{
\includegraphics[width=4.5cm]{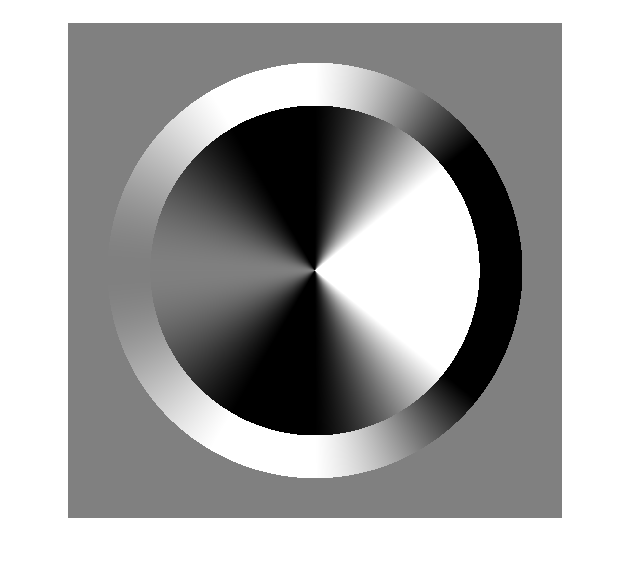}}
\caption{Example of a grey-level image for which PMA fails.\label{fig:exception}}
\end{figure}

Consider, as an example, the image
\begin{equation}
f(r,\theta)=\frac{1}{\pi}R(r)\left(\cos\theta+\cos2\theta\right)\,.\label{eq:ex1}
\end{equation}
It is clear that the image $f(r,\theta)$ is not rotationally symmetric, since the fundamental period of $\cos\theta+\cos2\theta$ is $2\pi$. The Fourier series coefficients $F(r,k)$ in \eqref{eq:sfgrey}, for non-negative $k$, {\it i.e.}, the coefficients that determine the PMs in \eqref{eq:pmsf}, are given by
\begin{equation}
F(r,k)=R(r)\left(\delta(k-1)+\delta(k-2)\right),\label{eq:ex3}
\end{equation}
where $\delta(\cdot)$ denotes the Dirac delta function. This expression is easily obtained from the Fourier series synthesis formula $f(r,\theta)=1/2\pi\sum_{k=-\infty}^{+\infty} F(r,k)e^{-jk\theta}$ \citep[see, {\it e.g.},][]{oppss}. From \eqref{eq:pmsf} and \eqref{eq:ex3}, we obtain the PMs of the image $f(r,\theta)$:
\[
\mu_k=\delta(k-1)\int_{0}^{\infty}r^2R(r)\,dr+\delta(k-2)\int_{0}^{\infty}r^3R(r)\,dr\,.
\]
It is now clear that we can specify a function $R(r)$ such that only one PM is nonzero. For example, with
\begin{equation}
R(r)=H(r)-2H(r-1)+H(r-\sqrt[3]{2})\,,\label{eq:ex2}
\end{equation}
where $H(\cdot)$ denotes the Heaviside step function, we obtain:
\begin{align}
\mu_0&=0\nonumber\\
\mu_1&=0\nonumber\\
\mu_2&=\frac{1-\sqrt[3]{2}}{2}\nonumber\\
\mu_3&=0\nonumber\\
\mu_4&=0\nonumber\\
\mu_5&=0\nonumber\\
\cdots&=0\,.\label{eq:solution}
\end{align}
This shows that the PMA algorithm fails to process the image specified by \eqref{eq:ex1} and \eqref{eq:ex2}, since, from \eqref{eq:solution}, besides wrongly assuming the image is $2$-fold rotationally symmetric, PMA would fruitlessly search for a pair of nonzero PMs. This grey-level image $f(r,\theta)$ is shown in Fig.~\ref{fig:exception}.

We distinguish two cases where PMA applied to grey-level images fails, caused by the failure of two different parts of the algorithm. Namely, PMA fails when the image $f(r,\theta)$
\begin{itemize}
\item has a single nonzero PM, caused by
\begin{itemize}
\item nonzero Fourier series coefficients $F(r,k)$ in \eqref{eq:sfgrey} for more than one value of $k$ but nonzero moments $\mu_k$ in \eqref{eq:pmsf} for a single value of $k$;
\item nonzero $F(r,k)$ in \eqref{eq:sfgrey} for a single $k\geq 1$ (the case $k=0$ corresponds to a radial image, not normalizable in what respects to orientation);
\end{itemize}
\item has zero PMs for all orders $k$ not multiple of a given $\gamma>1$ without being $\gamma$-fold rotationally symmetric, caused by
\begin{itemize}
\item at least one Fourier series coefficient $F(r,k)$ in \eqref{eq:sfgrey} is nonzero for $k$ not multiple of $\gamma$ but the nonzero moments $\mu_k$ in \eqref{eq:pmsf} occur only for $k$ multiple of $\gamma$.
\end{itemize}
\end{itemize}
In the case of a single nonzero PM, the algorithm fails to find co-prime pairs, whereas in the case of nonzero PMs only for $k$ multiple of a given $\gamma>1$, the failure lies on the incorrect detection of a $\gamma$-fold symmetry. Grey-level images of these classes appear to be somewhat particular, as the example in Fig.~\ref{fig:exception} illustrates, thus we did not face any failure when processing real images of trademark logos, see examples in Figs.~\ref{fig:trade}-\ref{fig:trade3}.

\bibliographystyle{spbasic}      
\bibliography{rot}   

\end{document}